\documentclass[11pt]{article}

\usepackage[preprint]{acl}

\usepackage{times}
\usepackage{latexsym}
\usepackage{amssymb}
\usepackage{amsmath}
\usepackage{booktabs}
\usepackage{array}

\usepackage[T1]{fontenc}

\usepackage[utf8]{inputenc}

\usepackage{microtype}

\usepackage{inconsolata}

\usepackage{graphicx}

\definecolor{orba}{RGB}{255, 143, 0}  %
\definecolor{salmon}{RGB}{255, 100, 100}  %

\usepackage[most]{tcolorbox}
\tcbset{appendixbox/.style={enhanced, breakable, listing only, boxrule=0.5pt, arc=1.5pt,
  left=3pt, right=3pt, top=2pt, bottom=2pt, fonttitle=\bfseries\small,
  listing options={basicstyle=\ttfamily\scriptsize, breaklines=true, breakatwhitespace=false,
    columns=fullflexible, keepspaces=true, showstringspaces=false, upquote=true}}}
\newtcblisting{promptbox}[1]{appendixbox, colback=gray!4, colframe=gray!55!black, title={#1}}
\newtcblisting{prmbox}[1]{appendixbox, colback=blue!3, colframe=blue!45!black, title={#1}}
\newtcblisting{agentbox}[1]{appendixbox, colback=green!3, colframe=green!35!black, title={#1}}

\title{\textsc{AdaPRA}: Process Reward Agents with \\
Adaptive Rubrics for Data Science}

\author{
 \textbf{Atharva Naik\textsuperscript{1}},
 \textbf{Vivek Chari\textsuperscript{2}},
 \textbf{Pol Llado\textsuperscript{2}},
 \textbf{Daniel Fried\textsuperscript{1}},
 \textbf{Carolyn Rosé\textsuperscript{1}}, 
 \\
 \textbf{Benjamin Van Durme\textsuperscript{2}},
 \textbf{Sam Thomson\textsuperscript{2}},
 \textbf{Justin Svegliato\textsuperscript{2}},
\\
\\
 \textsuperscript{1}Carnegie Mellon University,
 \textsuperscript{2}Microsoft
\\
 \texttt{\{arnaik,dfried,cprose\}@cs.cmu.edu}
}

\newcommand{\methodname}{\textsc{AdaPRA}}

\begin{document}
\maketitle
\begin{abstract}
Data science agents can automate end-to-end machine learning workflows, yet they remain far from competitive on human leaderboards. 
Agents make process errors such as insufficient exploration, unsound validation, coding bugs, and poor use of a fixed budget (re-running the same exploratory data analysis and baseline models in every attempt).
We propose \methodname{} to address these issues: a persistent coding agent harness that accumulates knowledge across attempts on the same task, with \emph{rubric-guided process supervision} that identifies and corrects errors. 
Our process reward agent audits the agent's trajectory and workspace by executing checks, and adapts its rubrics to each task. 
We find that rubrics seeded from easier tasks generalize to harder tasks, with rubric adaptation further improving performance. 
With five attempts per task, AdaPRA improves an 80B Qwen model's leaderboard percentile by 24.5 points on DSPredict-Easy and 14.8 points on DSPredict-Hard while reducing process errors.
We also show that supervision with rubrics mined from one base model's runs improves another (DeepSeek-V3.2), and that \methodname{} beats DataPRM, a competitive PRM for data science.
We also share our code for reproducibility.
\end{abstract}

\section{Introduction}

\begin{figure}[!tbh]
\centering
\includegraphics[width=0.5\textwidth]{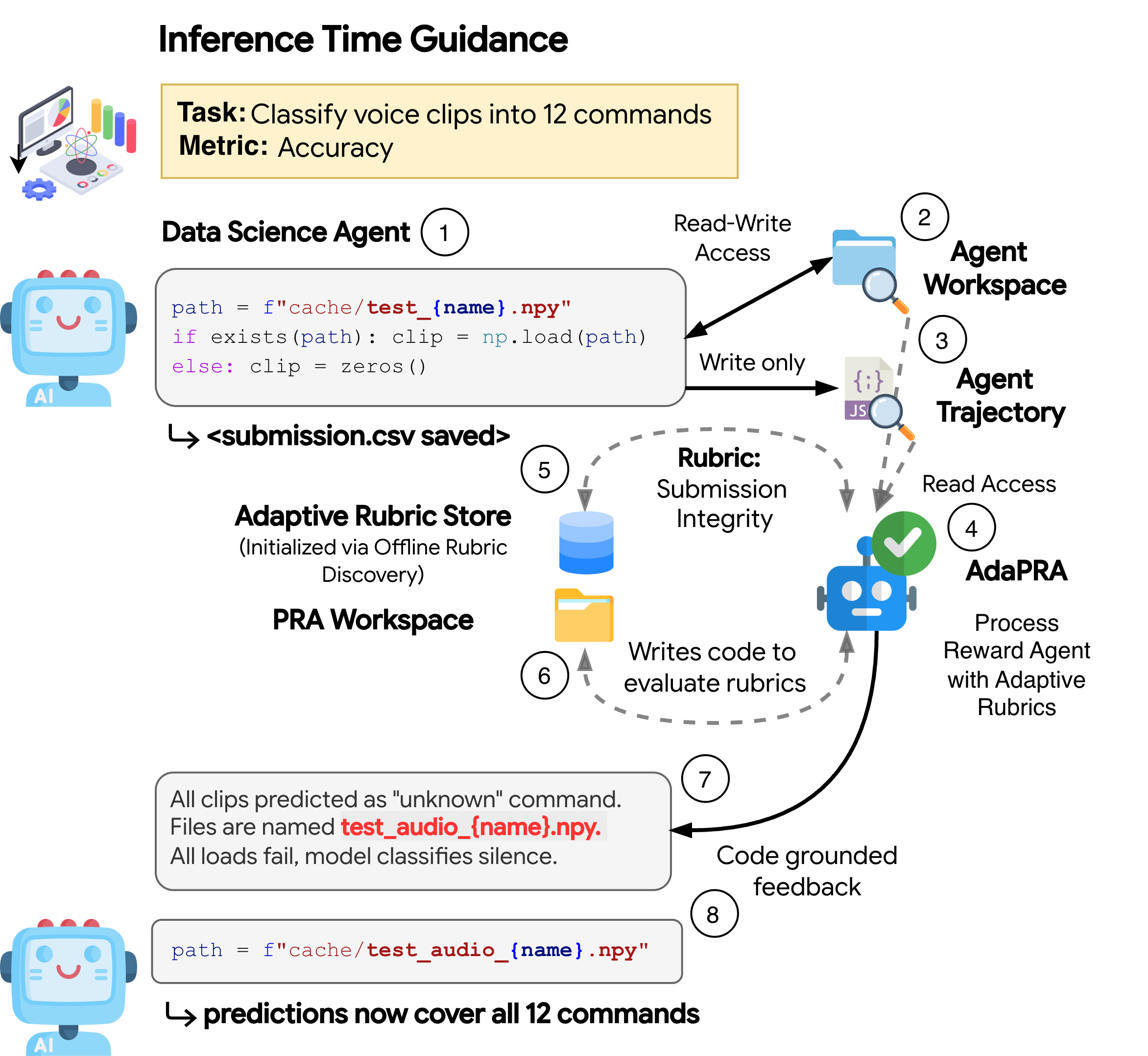}
\caption{\textbf{AdaPRA corrects process errors.} Here, the agent must classify voice clips into 12 commands. A wrong file path makes every test clip load as silent audio, so every prediction is classified as the ``unknown'' command even though the trajectory indicates no errors. To address this, AdaPRA checks the predictions file under the \emph{Submission Integrity} rubric, identifies the cause, and the agent fixes it.}
\label{fig:adapra_intro_teaser}
\end{figure}

Data science agents can consistently complete end-to-end
machine learning (ML) workflows, yet their scores lag on benchmarks such as DSGym \citep{nie2026dsgym}.
Much of the gap comes from \emph{process errors}, mistakes in intermediate steps that lead to worse outcomes.
The most common is \emph{simplicity bias} \cite{nie2026dsgym}: agents stop at the first valid pipeline instead of exploring stronger models, preprocessing, or features.
Others include unsound validation, mismatched training objectives, and leaked validation sets.
Many are silent: the pipeline produces a valid submission, and the trajectory shows no sign of the error.
Agents also make poor use of repeated attempts: harnesses like DSGym start each attempt from scratch, so every attempt repeats exploratory data analysis (EDA), baselines, and strategies that already failed.
Competitive submissions therefore require a harness that accumulates progress across attempts and \emph{process supervision} \cite{lightman2024let} that corrects process errors, including those that leave no trace in the trajectory.

We propose \methodname{}, a method with four components: 
a persistent harness; 
a \emph{process reward agent} (PRA) that supervises the agent; 
automatic seed rubric discovery from past agent runs; and task-specific rubric adaptation.
The harness keeps the agent's workspace across repeated attempts on the same task, so exploratory insights, and trained models
from one attempt carry into the next.
The PRA supervises the agent during each attempt, judging its trajectory and workspace against rubrics using tools and code execution \cite{qiu2026rewarding,peng2025rewardagent,raghavendra2026agenticrubrics}.

We design \methodname{} around five challenges of supervising a long-running data science agent.
First, silent errors leave no trace in the trajectory, so an LLM judge that reads only the trajectory misses them; \methodname{} instead 
runs checks on the agent's code, outputs, and traces.
In Figure~\ref{fig:adapra_intro_teaser}, the agent submits ``unknown'' for all 158,538 test clips; \methodname{} finds identical predicted probabilities for every clip, confirms that the code builds nonexistent paths, and names the fix.
Second, the evidence spans hundreds of turns and a workspace of code, models, and data, too much to push into the context window, so \methodname{} 
\emph{pulls} context on demand.
Third, feedback after an attempt ends comes too late to steer it, so the PRA reviews at a fixed cadence \emph{during} the attempt, similar to \citet{gandhi2026steer}; at any stage, however, many rubrics are irrelevant or already passing, so our PRA reports only the relevant failing ones to avoid flooding the agent's context.
Fourth, rubrics must target the errors agents actually make, many visible only in the workspace, so \methodname{} discovers seed rubrics from past agent trajectories and workspace artifacts, extending the rubric discovery procedure of \citet{mece}.
These errors recur across tasks, and we show that \textbf{rubrics learned from easier tasks generalize to harder tasks}.
Finally, each task has pitfalls that general rubrics cannot anticipate, so \methodname{} adapts its rubrics to each task as attempts proceed \cite{rubricem,shao2025dr}, which further improves performance.

Replacing the DSGym harness with \methodname{}, using Claude Sonnet-4.6 for the PRA, improves the mean Kaggle percentile of an 80B Qwen3-Coder-Next agent by 24.5 points on DSPredict-Easy and by 14.8 points on DSPredict-Hard, whose tasks span vision, text, speech, and molecular prediction. These results match or exceed GPT-5.1 with high reasoning across all metrics. Ablations show that the persistent harness, code execution, and rubric adaptation each contribute to these gains.

In summary, \methodname{} makes three contributions: (1) a persistent harness for data science agents that carries progress across attempts, including EDA findings, baselines, and failed strategies, so that each attempt spends its fixed budget on new experiments; (2) a PRA that audits the agent's trajectory and workspace by executing checks and adapts its rubrics to each task; and (3) a rubric discovery procedure that contrasts whole agent attempts, using both their trajectories and workspace artifacts, to yield seed rubrics that transfer from easier to harder tasks and to a second base model.

\section{Related Work}

\paragraph{Agents for data science.}
Agents that automate ML workflows are evaluated on DSGym \cite{nie2026dsgym} and agentic data-analysis suites \cite[\textit{inter alia}]{qiu2026rewarding,kulikov2026autodata}.
ML engineering agents such as AIDE \cite{jiang2025aide} and ML-Master \cite{liu2025ml} use an external scaffold that keeps a tree of candidate solutions, picks one to refine, and gives the model a short, curated context at each step.
\methodname{} instead leaves the search to the coding agent: its harness keeps the workspace across attempts, and the agent decides what to reuse, using files as memory \cite{xu2025everything,zhou2026filesystem}.
The two designs could be combined, with such a scaffold running over a persistent agent.

\paragraph{Rubric adaptation.}
Prior work has explored rubrics, checklists, and natural-language principles as reward specifications that are more flexible than scalar preference labels \cite{wang2026rubric,gunjal2025rubrics,viswanathan2026checklists,yu2025rewardanything} (compared in detail in Appendix~\ref{sec:appendix-rubric-rms}). 
Other work adapts rubrics over time: RubricEM \cite{rubricem} adds and prunes criteria in an evolving buffer through a meta-RL training loop, turning past attempts into reusable experience, and \citet{shao2025dr} evolve rubrics during training to mitigate reward hacking. 
These papers adapt rubrics to shape a training reward and apply them with a standard LLM judge that has no tools or code execution. 
\methodname{} instead adapts rubrics while it supervises the agent on the task at hand, without gold feedback such as Kaggle scores: the PRA checks rubrics with tools and code, and a per-task store turns pitfalls that a general rubric set cannot cover into explicit criteria for later reviews of the same task.

\paragraph{Process reward agents.}
Process reward models (PRMs) score intermediate steps rather than only final outcomes \cite{lightman2024let,gandhi2026steer}. 
\emph{Process reward agents} (PRAs) extend PRMs to verify actively before judging, by executing code, retrieving evidence, or exploring a workspace, in math \cite{zhang2026agentv}, software engineering \cite{raghavendra2026agenticrubrics}, GUIs \cite{xiong2025gui}, medical reasoning \cite{sohn2026process}, data analysis \cite{qiu2026rewarding}, and general reward modeling~\cite{peng2025rewardagent}. 
Code execution makes verification more grounded, but existing PRAs rely on a static rubric set, whereas \methodname{} audits both the trajectory and the workspace of a long-running agent and adapts its rubrics to each task. %
Table~\ref{tab:pra-comparison} (Appendix~\ref{sec:appendix-pra-comparison}) summarizes the design coverage of related systems.

\paragraph{Rubric discovery.}
Rubrics can be discovered automatically from past agent behavior.
RubricEM discovers rubrics through policy decomposition across training episodes \cite{rubricem}, and \citet{mece} refine per-prompt rubrics for RL post-training by contrasting the two highest-scoring candidate responses to a prompt.
Both derive rubrics from responses or trajectories alone, whereas we also mine the workspace artifacts each attempt leaves behind (\S\ref{sec:discovery}) and show that the resulting rubrics transfer from easier to harder tasks.

\section{Method}

\begin{figure*}[!tbh]
\centering
\includegraphics[width=\textwidth]{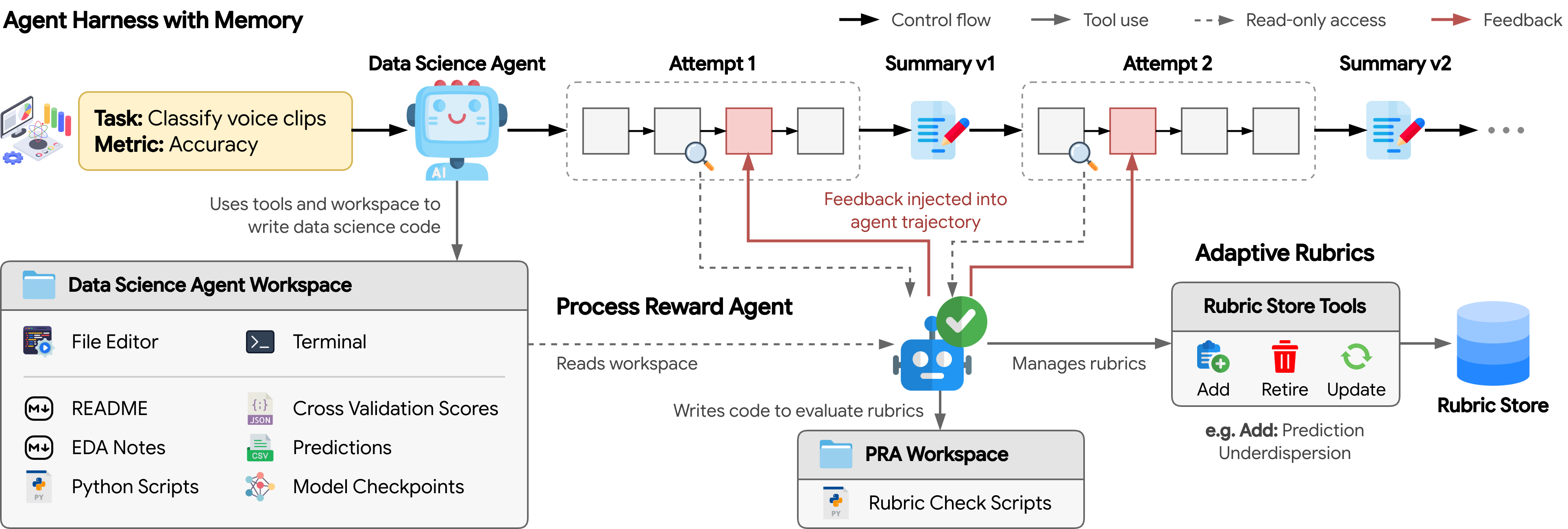}
\caption{\textbf{\methodname{} overview.} The agent works in a workspace that persists across attempts. During each attempt, the PRA reads the trajectory and workspace, runs check scripts in its own workspace, injects feedback on failing rubrics into the agent's trajectory, and adds, updates, or retires rubrics in the per-task store.}
\label{fig:adapra_method}
\end{figure*}

\methodname{} (Figure~\ref{fig:adapra_method}) combines a persistent harness (\S\ref{sec:harness}) with a PRA that checks the agent against rubrics discovered offline (\S\ref{sec:discovery}), grounds its judgments in code (\S\ref{sec:pra}), and adapts the rubrics to each task (\S\ref{sec:adaptive}).

\subsection{Problem Setup}
\label{sec:setup}

Let $x$ be a predictive data science task represented as a set of files and a natural-language prompt. In each attempt, the base agent produces a trajectory $\tau = (a_1, o_1, \ldots, a_T, o_T)$ of actions $a_t$ (writing code, training models, creating submissions) and observations $o_t$ (logs, scores, errors), and operates in a workspace $W$ on the file system. 
The agent makes $N$ attempts per task, each ending with a submission file, and every condition we compare, including the baselines, makes the same $N$ attempts (\S\ref{sec:experiments}).
We score every attempt on Kaggle only to compute evaluation metrics, and neither the agent nor the PRA ever sees these scores, test labels, or any other external feedback between attempts.

Process errors such as simplicity bias and leaky cross-validation often still yield a valid submission (Table~\ref{tab:rubric-meta}), so we supervise the process with $m$ rubrics $\mathcal{R} = \{\rho_1, \ldots, \rho_m\}$, natural-language criteria such as \emph{``compare the first valid baseline against at least one stronger model family.''}
Every $c$ base-agent turns and when the base agent tries to finish, the PRA scores the partial trajectory $\tau_{:t} = (a_1, o_1, \ldots, a_t, o_t)$ up to the current turn $t$ and the current workspace against each rubric $\rho_j$,
\[
S(x, W, \tau_{:t}, \rho_j) = (p_j, i_j, e_j, \hat{a}_j),
\]
with probability of satisfaction $p_j \in [0,1]$, importance $i_j \in [0,1]$, rationale $e_j$, and suggested next action $\hat{a}_j$.
The probability of satisfaction $p_j$ captures the %
confidence that the agent adhered to the rubric guidelines.
The importance $i_j$ captures the relevance of the rubric to task $x$ and to the current stage $t$ of the attempt, since one rubric set serves many kinds of task and some rubrics apply only at later stages (e.g., a rubric requiring a prediction file applies only once a model has been trained).

We show the agent only the relevant failing rubrics, those with $i_j \geq 0.5$ and $p_j < 0.6$ (Table~\ref{tab:run-config}), i.e., rubrics with importance at least 0.5 and a greater than 40\% chance of having failed.\footnote{We show only failures because in an earlier version that reported every rubric, most rubrics passed in nearly every report and the agent's next action was similar whether or not a report contained a failure (Appendix~\ref{sec:appendix-feedback-content}).
}
The PRA compiles these rubrics into a natural-language report and injects it directly into the message stream of the agent (Appendix~\ref{sec:appendix-prompts-pra}).\footnote{During development, we found this more effective than writing the report to a markdown file that the agent could choose to open and inspect, but sometimes did not despite instructions in its system prompt.}

\subsection{Agent Harness with Memory}
\label{sec:harness}

In our harness, the agent acts through a terminal and a file editor and is built on OpenHands \cite{wang2025openhands}. 
The DSGym harness
\cite{nie2026dsgym} resembles a REPL and resets the workspace between attempts, so each attempt rebuilds data loading, validation, and a baseline from scratch. 
Our harness keeps one workspace per task across all $N$ attempts: exploratory notes, features, trained models, validation results, and submissions from attempt $k$ remain available in attempt $k{+}1$, and only the model context resets. 
This lets later attempts avoid or fix errors from earlier ones and spend their budget on directions earlier ones left open. 
Each later attempt also receives the agent's own summaries of all earlier attempts, so it can explore new directions with awareness of what has already been tried. In supervised conditions, each summary is followed by the feedback from the final review of that attempt (Appendix~\ref{sec:appendix-prompts-agent}).
Everything carried across attempts is produced by the agent or the PRA, and the only scores it contains are the agent's own validation scores.
Turn, wall-clock, and per-command limits bound each attempt (Appendix~\ref{sec:appendix-harness-impl}).

\subsection{Rubric Discovery}
\label{sec:discovery}
\label{sec:rubrics}

Hand-writing rubrics requires expert effort that becomes infeasible across many tasks and trajectories of hundreds of turns.
We discover rubrics offline from past harness-only runs of the base agent, using the leaderboard percentile of each attempt to contrast stronger and weaker attempts (Figure~\ref{fig:rubric_discovery}; prompts in Appendix~\ref{sec:appendix-prompts-discovery}).
Like RubricEM \cite{rubricem}, we derive rubrics from agent trajectories, but we also mine the workspace each attempt leaves behind (code, notes, and submissions).
From \citet{mece}, we borrow contrastive discovery and the principle that rubrics be mutually exclusive and collectively exhaustive (MECE).
In \textbf{stage one}, an LLM judge (Claude Sonnet~4.6) extracts generalizable behaviors that plausibly drove each attempt's outcome, either \emph{good behaviors} or \emph{issues}, each with quoted evidence. It is told whether the attempt ranks in the top or bottom quarter of the attempts on the same task, so each attempt is judged relative to its siblings and task difficulty is held fixed (Appendix~\ref{sec:appendix-prompts-discovery}). 
In \textbf{stage two}, a single LLM call clusters the behaviors into a MECE rubric set, with good behaviors and issues as positive and negative evidence for the same rubric. Stage one extracts both kinds from every attempt, but to keep the behaviors that explain the score gap, we cluster only the good behaviors of top attempts and the issues of bottom attempts (519 of 906; Appendix~\ref{sec:appendix-prompts-discovery}).

\paragraph{Evidence source.} Some behaviors are visible only in the files
left
behind (e.g., code and markdown notes) and others only in how the agent worked, as recorded in the trajectory, $\tau$. The judge therefore sees $\tau$, a snapshot of the workspace $W$, and the agent's end-of-attempt summary, and tags each rubric by the evidence it needs. \textbf{Artifact-grounded} rubrics can be checked from $W$ alone (``the submission has the correct columns and row count''). \textbf{Trajectory-grounded} rubrics require $\tau$ (``kept retrying the same failing command'').
The tags measure what each view contributes to discovery.

\paragraph{Discovered set.} On harness-only runs of DSPredict-Easy (38 tasks, 5 attempts each), stage one mines 906 behaviors, 564 issues and 342 good behaviors. Issues are mostly trajectory-grounded (344 of 564) and good behaviors mostly artifact-grounded (235 of 342), so mining only the trajectory would under-represent good behaviors, and a judge that sees only the workspace would miss most issues. Stage two yields 17 rubrics covering validation, model exploration and selection, preprocessing, submission integrity, and efficiency, including criteria with no artifact counterpart, such as pivoting to a qualitatively different approach once repeated experiments stop improving. Together with two cross-attempt rubrics (Appendix~\ref{sec:appendix-rubrics}), these form the 19 seed rubrics (Table~\ref{tab:rubric-meta}) used in every configuration and applied unchanged to DSPredict-Hard.

\subsection{Process Reward Agent}
\label{sec:pra}
\label{sec:axes}

Judging many of these errors requires reviewing the modeling code itself: whether a stronger model family was trained and compared, whether the cross-validation leaks information, whether engineered features reach the final model, and whether the submission comes from the pipeline the agent reports. 
The trajectory records the agent's claims, which can mislead a judge that takes them at face value, so the PRA is itself an agent, with its own terminal and scratch space that checks the claims against the workspace $W$ (instructions and tools in Appendix~\ref{sec:appendix-prompts-pra}). 
For each rubric, the PRA can write and execute its own code in its own workspace against the base agent's files (e.g.\ recompute a cross-validation score from out-of-fold predictions) and ground $p_j$ in the result. Rubrics that need no computation, such as the trajectory-grounded ``kept retrying the same failing command'' (\S\ref{sec:discovery}), it scores by reading the trajectory or files. The base agent's workspace is read-only to the PRA, so supervision cannot alter the base agent's work.

The fixed cadence delivers feedback while a course correction is still possible, similar to \citet{gandhi2026steer} (Appendix~\ref{sec:appendix-cadence} varies $c$), and the finish review acts as a final gate. 
Trajectories reach hundreds of turns, so instead of being injected into the PRA's context, $\tau_{:t}$ and the transcript of each earlier attempt are written as turn-annotated text files to the PRA's workspace, which the PRA opens and searches with its tools (file names in Appendix~\ref{sec:appendix-prompts-pra}).
The PRA reads them selectively, and initial experiments showed that this on-demand access works better than appending each new batch of agent turns to the PRA's context (Appendix~\ref{sec:appendix-axes}). 
The base agent is paused while the PRA runs, and this wall-clock time does not count against its budget.
Failing rubrics (\S\ref{sec:setup}) are returned to the agent ordered by importance, with their suggested fixes, to keep feedback short and actionable (Appendix~\ref{sec:appendix-prm-examples} shows examples). If the finish review fails, the agent gets one final refinement round, which bounds the cost of the gate. 

\subsection{Adaptive Rubrics}
\label{sec:adaptive}

The seed rubrics from \S\ref{sec:discovery} capture failures that recur across tasks, but each task has pitfalls of its own. 
The PRA can therefore add, update, and retire rubrics in a per-task store (similar to RubricEM \cite{rubricem}), initialized with the seed set and kept across the task's attempts, so a failure found in one attempt becomes an explicit criterion in later reviews.
The PRA edits the store with three tools. \emph{Add} creates a rubric only after the PRA has run and recorded a check for it in the same review, \emph{update} rewrites a rubric's specification, and \emph{retire} deactivates a rubric and records the reason.
Stores are isolated across tasks, so that results do not depend on the order in which tasks are run.
Added rubrics tend to be task specific. 
For example, on House Prices the PRA added a rubric it named \emph{prediction underdispersion}, which checks whether out-of-fold predictions are much less spread out than the target, a sign that the model pulls extreme prices toward the mean. 

\section{Experiments}
\label{sec:experiments}

\paragraph{Benchmarks.}
\label{sec:experiments-setup}
We evaluate on DSGym's \cite{nie2026dsgym} DSPredict-Easy, 38 mostly tabular Kaggle competitions from the Playground Series, and DSPredict-Hard, 54 competitions spanning tabular, vision, text, audio, and molecular data. 
Because the seed rubrics are discovered on DSPredict-Easy (\S\ref{sec:discovery}), it serves as the development benchmark and DSPredict-Hard as a held-out test of transfer to harder tasks.

\paragraph{Models and budget.}
The base agent is Qwen3-Coder-Next, an 80B open-weights coding model, and the PRA uses Claude Sonnet~4.6. 
Every condition makes $N=5$ attempts per task, and all conditions built on our harness share the same per-attempt limits (Appendix~\ref{sec:appendix-harness-impl}). 
DSPredict-Hard attempts receive a larger budget and GPU access because its tasks involve larger data and neural models.

\paragraph{Baselines and Ablations.}
On each benchmark we compare six conditions. 
Two baselines are without process supervision: the DSGym harness \cite{nie2026dsgym}, which resets the workspace between attempts, and our persistent harness without \methodname{} supervision (\emph{harness only}), which isolates the effect of our persistent harness. 
To compare with competitive PRMs, we replace the PRA with DataPRM \cite{qiu2026rewarding}, a 4B generative process reward model that runs code to verify each agent step and scores the step 1 (correct), 0.5 (minor issues), or 0 (fatal error); the base agent, harness, and feedback channel are those of \methodname{} (Appendix~\ref{sec:appendix-dataprm}).
Both supervisors execute code and score the agent, but DataPRM judges each step's correctness without rubrics, while \methodname{} checks the trajectory and workspace against seed and task-adapted rubrics that also cover errors spanning many steps, such as stopping at the first valid pipeline.
The remaining three conditions share our harness and the seed rubrics: \methodname{}; \emph{$-$adaptive rubrics}, which keeps the seed set fixed, so on DSPredict-Hard it measures how well the seed rubrics alone transfer; and \emph{$-$PRA}, which keeps adaptation but removes the PRA's terminal, so it can read files but not run code (\S\ref{sec:pra}, Appendix~\ref{sec:appendix-prompts-pra}).
As external context we also report the single-attempt DSGym results of Qwen3-Coder-480B and GPT-5.1 (high) \cite{nie2026dsgym}.

\paragraph{Additional ablations.}
To isolate the effect of supervision within an attempt, we run two further ablations on DSPredict-Easy with a single attempt per task: the PRA cadence, $c \in \{5, 10, 20\}$ agent turns (Appendix~\ref{sec:appendix-cadence}), and DeepSeek-V3.2 as the base agent in place of Qwen3-Coder-Next (Appendix~\ref{sec:appendix-base-agent}).

\paragraph{Metrics.}
We report four task-level rates: \textbf{Pct.}, the mean leaderboard percentile (private leaderboard where one exists, else public); 
\textbf{Valid}, the share of tasks with a submission Kaggle accepts and scores; \textbf{>Median}, the share of tasks scoring above the leaderboard median; and \textbf{Medal}, the share of tasks reaching a Kaggle medal threshold. 
Averaging only over tasks with a valid submission would favor a condition that fails to submit on hard tasks, because it drops its lowest scores; all rates therefore divide by every task in the benchmark and count an invalid or missing submission as a zero percentile. 
Ideally, a deployed system would identify the best submission without test scores, so we evaluate four selection policies (Appendix~\ref{sec:appendix-selection-policy}): the first attempt, the last attempt, the mean over attempts (the expected result of choosing one at random), and the best attempt, an oracle upper bound. The main tables report the best and mean attempts. 
We test each difference in mean percentile for significance with a one-sided paired permutation test over tasks, which randomly swaps the two conditions' scores within each task (Appendix~\ref{sec:appendix-significance}).

\section{Results}
\label{sec:results}

\begin{table*}[t!]
\centering
\small
\begin{tabular}{@{}l cccc c cccc@{}}
\toprule
 & \multicolumn{4}{c}{\textbf{Best attempt}} & & \multicolumn{4}{c}{\textbf{Per-attempt (mean)}} \\
\cmidrule(lr){2-5} \cmidrule(lr){7-10}
\textbf{Method} & \textbf{Pct.} & \textbf{Valid} & \textbf{>Median} & \textbf{Medal} & & \textbf{Pct.} & \textbf{Valid} & \textbf{>Median} & \textbf{Medal} \\
\midrule
Qwen3-Coder-Next + DSGym          & 44.5 &  76.3 & 50.0 & 2.6 & & 36.3 &  41.0 & 35.6 & 1.1 \\
Qwen3-Coder-Next + Ours           & 60.8 & 100.0 & 71.1 & \textbf{7.9} & & 51.3 & 84.5 & 53.9 & 2.6 \\
\midrule
\multicolumn{10}{@{}l}{\emph{Process supervision on top of Our Harness}}\\
\quad + DataPRM & 54.6 & 100.0 & 65.8 & 5.3 & & 44.8 & 100.0 & 44.6 & 2.1 \\
\cmidrule(l){1-10}
\quad + PRA + adaptive rubrics (\methodname{}) & \textbf{69.0} & 100.0 & \textbf{86.8} & \textbf{7.9} & & \textbf{61.0} & 100.0 & 73.2 & 3.7 \\
\quad \hspace{1.2em}$-$ adaptive rubrics     & 67.3 & 100.0 & 86.8 & \textbf{7.9} & & 58.8 & 99.5 & 72.5 & \textbf{4.2} \\
\quad \hspace{1.2em}$-$ PRA & 63.0 & 100.0 & 86.8 & 2.6 & & 57.8 & 100.0 & 73.7 & 2.1 \\
\midrule
\multicolumn{10}{@{}l}{\emph{Single-attempt external references (reported by DSGym)}}\\
Qwen3-Coder-480B + DSGym          & -- &  -- & -- & -- & & 37.1 &  86.5 & 33.3 & -- \\
GPT-5.1 (high) + DSGym            & -- &  -- & -- & -- & & 60.4 & 100.0 & \textbf{75.0} & -- \\
\bottomrule
\end{tabular}
\caption{\textbf{DSPredict-Easy:}
Our harness (Ours), with and without process supervision, compared with the DSGym harness.
\emph{Best attempt} is the maximum over a task's five sequential attempts, and per-attempt averages %
each task's scored attempts; all rates use a fixed denominator of 38 tasks (190 attempts). %
Qwen3-Coder-480B and GPT-5.1 (high) are single-attempt DSGym runs \cite{nie2026dsgym}, so they have no best-attempt results; their percentile is rescaled to our fixed denominator (multiplied by the valid rate), their other rates are as reported, and DSGym reports no Easy medal rates.}
\label{tab:harness}
\end{table*}

\begin{table*}[t!]
\centering
\small
\begin{tabular}{@{}l cccc c cccc@{}}
\toprule
 & \multicolumn{4}{c}{\textbf{Best attempt}} & & \multicolumn{4}{c}{\textbf{Per-attempt (mean)}} \\
\cmidrule(lr){2-5} \cmidrule(lr){7-10}
\textbf{Method} & \textbf{Pct.} & \textbf{Valid} & \textbf{>Median} & \textbf{Medal} & & \textbf{Pct.} & \textbf{Valid} & \textbf{>Median} & \textbf{Medal} \\
\midrule
Qwen3-Coder-Next + DSGym          & 21.3 &  68.5 & 11.1 & 3.7 & & 25.2 &  36.3 & 11.8 & 3.0 \\
Qwen3-Coder-Next + Ours           & 25.5 &  81.5 & 18.5 & 5.6 & & 26.2 & 64.8 & 14.4 & 3.7 \\
\midrule
\multicolumn{10}{@{}l}{\emph{Process supervision on top of Our Harness}}\\
\quad + DataPRM                                & 23.1 &  88.9 & 14.8 & 3.7 & & 21.3 & 79.3 & 12.8 & 3.7 \\
\cmidrule(l){1-10}
\quad + PRA + adaptive rubrics (\methodname{}) & \textbf{36.1} &  \textbf{90.7} & \textbf{33.3} & 5.6 & & \textbf{32.1} & 86.3 & \textbf{24.1} & 4.1 \\
\quad \hspace{1.2em}$-$ adaptive rubrics       & 31.8 &  88.9 & 22.2 & 5.6 & & 30.2 & \textbf{87.8} & 17.5 & 4.4 \\
\quad \hspace{1.2em}$-$ PRA                  & 31.8 &  \textbf{90.7} & 25.9 & \textbf{9.3} & & 30.9 & 83.0 & 19.8 & \textbf{8.1} \\
\midrule
\multicolumn{10}{@{}l}{\emph{Single-attempt external references (reported by DSGym)}}\\
Qwen3-Coder-480B + DSGym         & -- &  -- & -- & -- & & -- &  66.7 & 5.9 & 2.4 \\
GPT-5.1 (high) + DSGym            & -- &  -- & -- & -- & & -- &  85.7 & 14.3 & 4.8 \\
\bottomrule
\end{tabular}
\caption{\textbf{DSPredict-Hard:} As Table~\ref{tab:harness}, on the 54 competitions of DSPredict-Hard (fixed denominator of 54 tasks, 270 attempts for per-attempt).
DSGym reports medal rates but not percentiles on DSPredict-Hard, so the \emph{Pct.}\ of its single-attempt runs is also --, and their rates are shown as published
}
\label{tab:harness-hard}
\end{table*}

\subsection{DSPredict-Easy}
\label{sec:results-easy}
\methodname{} raises the mean percentile of Qwen3-Coder-Next from 44.5 in the DSGym harness to 69.0 and the above-median rate from 50.0\% to 86.8\% (Table~\ref{tab:harness}). 
Every task receives a valid submission, as does every individual attempt. Over our harness alone (\emph{harness only}), \methodname{} adds 8.2 percentage points and 15.7 above-median points; the percentile gain is significant ($p<0.001$, Appendix~\ref{sec:appendix-significance}). 
The gain holds per attempt (61.0 vs.\ 51.3 mean percentile), so supervision improves typical attempts as well as the best one. 
A single \methodname{} attempt on average matches a single GPT-5.1 (high) attempt in the DSGym harness (61.0 vs.\ 60.4 mean percentile).
Because the seed rubrics were discovered on DSPredict-Easy, DSPredict-Hard is the cleaner test of \methodname{}.

\subsection{DSPredict-Hard}
\label{sec:results-hard}
On DSPredict-Hard, \methodname{} raises the mean percentile from 21.3 to 36.1, the valid rate from 68.5\% to 90.7\%, and the above-median rate from 11.1\% to 33.3\% (Table~\ref{tab:harness-hard}). 
Over our harness alone, it adds 10.6 percentage points ($p<0.001$) and 14.8 above-median points.
\methodname{}'s best attempt exceeds a single GPT-5.1 (high) attempt on valid, above-median, and medal rates. 
Per attempt, \methodname{} exceeds GPT-5.1 in valid and above-median rates (86.3 vs.\ 85.7 and 24.1 vs.\ 14.3) but trails it slightly in medal rate (4.1 vs.\ 4.8).

\subsection{Ablations}
\label{sec:results-ablations}

\paragraph{Harness.} Our harness alone (persistent workspace and terminal-based tools) lifts the base agent on both benchmarks, raising the mean percentile from 44.5 to 60.8 and the valid rate from 76.3\% to 100\% on DSPredict-Easy, and from 21.3 to 25.5 and 68.5\% to 81.5\% on DSPredict-Hard. 
The two components split the gain differently: the harness contributes 16.3 percentage points and supervision 8.2 on DSPredict-Easy, versus 4.2 and 10.6 on DSPredict-Hard. 
Our harness is enough to produce valid and reasonable pipelines on easier tasks, while harder tasks need supervision. %

\paragraph{Adaptive rubrics.} Freezing the seed rubrics lowers the mean percentile by 1.7 points on DSPredict-Easy and by 4.3 points on DSPredict-Hard, where it also cuts the above-median rate from 33.3\% to 22.2\%. The Hard gap is significant ($p=0.024$; 32 wins, 13 losses). \S\ref{sec:results-transfer} shows why adaptation matters more on DSPredict-Hard: its tasks have more domain-specific pitfalls that the seed set cannot anticipate. 
The PRA adapts almost entirely by adding rubrics (177 additions, one update, and no retirements across the four adaptive runs), mostly in the first two attempts (Figures~\ref{fig:rubric-store-ops} and~\ref{fig:rubric-ops-by-attempt} in Appendix~\ref{sec:appendix-rubric-ops}). 
Code execution changes what it adds: prediction-sanity rubrics, which compute statistics of the agent's predictions such as their dispersion relative to the target, make up 35\% (DSPredict-Easy) and 19\% (DSPredict-Hard) of additions with execution and 3\% and 7\% without it (Table~\ref{tab:rubric-themes}).

\paragraph{PRA.} Scoring rubrics by reading alone lowers the mean percentile from 69.0 to 63.0 on DSPredict-Easy and from 36.1 to 31.8 on DSPredict-Hard, along with the Easy medal rate (7.9\% to 2.6\%) and the Hard above-median rate (33.3\% to 25.9\%). 
The gap is significant on DSPredict-Easy ($p=0.012$) but not on DSPredict-Hard ($p=0.077$; 29 wins, 18 losses).
Without code execution ($-$PRA), supervision still improves on the harness alone on both benchmarks (63.0 vs.\ 60.8 and 31.8 vs.\ 25.5), so rubric feedback helps even without execution. 
It gets a higher Hard medal rate (9.3\% vs.\ 5.6\%, five medals vs.\ three) despite a lower mean percentile, which we attribute to the noisiness of the metric: a medal depends on whether a submission's rank crosses a fixed leaderboard cutoff, so small score differences can flip it, especially on dense leaderboards.

\methodname{} has the highest mean percentile on both benchmarks and the highest above-median rate on DSPredict-Hard (33.3\% vs.\ 22.2\% and 25.9\%; all three tie at 86.8\% on DSPredict-Easy), so code execution and rubric adaptation are complementary.

\paragraph{Additional analyses.}
\textbf{Cadence} (Appendix~\ref{sec:appendix-cadence}): with a single attempt per task, every cadence we test improves on the harness alone (48.4 mean percentile), and invoking the PRA every 10 turns works best (56.4). Invoking it more often ($c{=}5$, 52.7) or less often ($c{=}20$, 49.9) gives smaller gains, consistent with a trade-off between timely feedback and interrupting the agent. We therefore use $c{=}10$ on DSPredict-Easy. On DSPredict-Hard we use $c{=}20$ to bound cost: Hard attempts take 1.6 times as many agent turns (Table~\ref{tab:turns}), and the PRA already accounts for 75\% of the cost of a Hard run (Table~\ref{tab:costs}). \textbf{Base agent} (Appendix~\ref{sec:appendix-base-agent}): with DeepSeek-V3.2 as the base agent and one attempt per task, \methodname{} raises the mean percentile from 36.9 to 44.7 and the above-median rate from 23.7\% to 47.4\% over the harness alone. It wins on 25 tasks and loses on 11 ($p=0.005$), so rubrics discovered from Qwen3-Coder-Next's runs also help a second base model.

\subsection{Easy-to-Hard Generalization}
\label{sec:results-transfer}
The seed rubrics are discovered only on DSPredict-Easy, whose tasks are mostly tabular, yet they improve DSPredict-Hard, whose tasks span vision, text, audio, time series, and molecular data. 
With the seed rubrics frozen, supervision adds 6.3 percentage points over the harness alone on DSPredict-Hard ($p=0.007$), close to its 6.5-point gain on DSPredict-Easy ($p=0.017$).

The same process failures occur on both benchmarks in the frozen-rubric runs, which score the same 19 rubrics throughout (Appendix~\ref{sec:appendix-rubric-transfer}). Every seed rubric fails in at least 24\% of DSPredict-Hard attempts, so none is specific to easy %
tasks. For each rubric we compute two statistics per benchmark: its mean importance over all PRA reviews and its failure rate, the share of attempts in which it fails at least once. Ranking the 19 rubrics by either statistic gives nearly the same order on both benchmarks (Spearman $\rho=0.91$ for importance, $\rho=0.66$ with $p=0.002$ for failure rate). The most frequent failures on both benchmarks are the two simplicity-bias rubrics, too little model diversity and ensembling and repeating near-identical experiments without pivoting (68--84\% of attempts). Harder tasks add pipeline failures that the seed set already covers: invalid cross-validation rises from 32\% to 57\% of attempts and unverified submissions from 33\% to 52\%.
Simplicity bias and pipeline errors are properties of the agent, which persist across domains, so rubrics that target them transfer. Table~\ref{tab:rubric-transfer-examples} in Appendix~\ref{sec:appendix-rubric-transfer} shows seven such cases, each a failure on a non-tabular DSPredict-Hard task that a DSPredict-Easy rubric flagged and the agent fixed within the same attempt.

Adaptation covers what the seed set misses. The PRA adds 0.8 rubrics per task on DSPredict-Easy and 1.2 on DSPredict-Hard, and adds at least one on 79\% of Hard tasks versus 51\% of Easy tasks (Figure~\ref{fig:rubric-adds-per-task}). Hard additions encode domain knowledge absent from tabular tasks, such as non-negative time-to-failure predictions in LANL Earthquake Prediction and excluding quantum-mechanical sub-components of the target from the features in CHAMPS Scalar Coupling.
The seed set %
transfers the domain-general failures and adaptation supplies the domain-specific ones, consistent with the larger adaptation gain on DSPredict-Hard.

\section{Conclusion}
We propose \methodname{}, which combines a harness whose persistent workspace accumulates knowledge across attempts on a task, a process reward agent that audits the agent's trajectory and workspace by executing checks and adapts its rubrics to each task, and offline discovery of seed rubrics from agent trajectories and workspace artifacts.
With Qwen3-Coder-Next as the base agent, \methodname{} raises the mean Kaggle percentile from 44.5 to 69.0 on DSPredict-Easy and from 21.3 to 36.1 on DSPredict-Hard, matching or exceeding GPT-5.1 (high).
The harness, the process reward agent, and adaptive rubrics each contribute to the gain, and removing any one of them lowers performance on both benchmarks.
Rubrics discovered on mostly tabular tasks (DSPredict-Easy) transfer to vision, text, audio, and molecular tasks (DSPredict-Hard) because the process failures they target come from the agent rather than the domain, and adaptation adds the domain-specific criteria that a general rubric set does not cover.
Supervision with rubrics mined from Qwen3-Coder-Next's runs also improves a second base model (DeepSeek-V3.2), and \methodname{} outperforms DataPRM \cite{qiu2026rewarding}, a competitive PRM proposed in prior work.

\section{Future Work}
First, training the base agent on dense feedback from the PRA could let it internalize the process criteria and depend less on supervision at test time.
Second, the PRA adapts rubrics without measuring how each affects the outcome and never retired one in our runs; assigning credit to individual rubrics could better inform adaptation.
Third, training smaller open-weight models to replace the closed-source PRA would extend \citet{gandhi2026steer} from non-agentic critics to agentic PRAs.
Finally, harness evolution like Meta-Harness \cite{lee2026meta} could further improve our persistent harness.

\section*{Acknowledgments}
This work was supported by Microsoft, which provided compute and funding. We thank Emmy Liu, Aakanksha Naik, and Sagnik Mukherjee for their feedback on the project and the draft.

\section*{Limitations}
\textbf{Supervision cadence.} To bound compute and API cost, we invoke the PRA every 20 turns on DSPredict-Hard instead of the 10 turns that work best on DSPredict-Easy; the PRA already accounts for 75\% of the cost of a Hard run (Table~\ref{tab:costs} in Appendix~\ref{sec:appendix-costs}). On DSPredict-Easy, a cadence of 20 is the weakest setting we test (Appendix~\ref{sec:appendix-cadence}), so the DSPredict-Hard results may understate the gains of more frequent supervision.

\textbf{Benchmark coverage.} We evaluate only on DSGym's DSPredict-Easy and DSPredict-Hard. Both consist of Kaggle competitions scored on a leaderboard metric, so we do not test whether rubric-based supervision helps in other data science settings, such as open-ended analysis or tasks without a predictive objective.

\textbf{Number of attempts.} All multi-attempt results use five attempts per task, and the additional ablations use a single attempt. We do not study how the gains of supervision change as the number of attempts grows, where selecting among more unsupervised attempts may recover part of the difference.

\textbf{Rubric adaptation.} Adaptation almost only adds rubrics (177 additions, one update, and no retirements; \S\ref{sec:results-ablations}), so the rubric store grows over a run. The PRA does not attribute local validation gains to individual rubrics, so it has little signal for revising or retiring rubrics that do not help. Better credit assignment could make adaptation more selective.

\textbf{Models.} Our main results use a single base agent (Qwen3-Coder-Next) and a single PRA model. We test a second base agent, DeepSeek-V3.2, only with one attempt per task on DSPredict-Easy (Appendix~\ref{sec:appendix-base-agent}).

\textbf{Reproducibility.} The PRA uses Claude Sonnet 4.6, a closed model accessed through a commercial API. Exact reproduction depends on the model remaining available and unchanged, and its retirement or updates could change results.

\bibliography{custom}

\begin{thebibliography}{24}
\providecommand{\natexlab}[1]{#1}

\bibitem[{Gandhi et~al.(2026)Gandhi, Xie, Naik, Zhu, and Rose}]{gandhi2026steer}
Shubham Gandhi, Yiqing Xie, Atharva Naik, Ruichen Zhu, and Carolyn Rose. 2026.
\newblock Steer, don't solve: Training small critic models for large code agents.
\newblock \emph{arXiv preprint arXiv:2606.21811}.

\bibitem[{Gunjal et~al.(2025)Gunjal, Wang, Lau, Nath, He, Liu, and Hendryx}]{gunjal2025rubrics}
Anisha Gunjal, Anthony Wang, Elaine Lau, Vaskar Nath, Yunzhong He, Bing Liu, and Sean Hendryx. 2025.
\newblock Rubrics as rewards: Reinforcement learning beyond verifiable domains.
\newblock \emph{arXiv preprint arXiv:2507.17746}.

\bibitem[{Jiang et~al.(2025)Jiang, Schmidt, Srikanth, Xu, Kaplan, Jacenko, and Wu}]{jiang2025aide}
Zhengyao Jiang, Dominik Schmidt, Dhruv Srikanth, Dixing Xu, Ian Kaplan, Deniss Jacenko, and Yuxiang Wu. 2025.
\newblock Aide: Ai-driven exploration in the space of code.
\newblock \emph{arXiv preprint arXiv:2502.13138}.

\bibitem[{Kulikov et~al.(2026)Kulikov, Whitehouse, Wu, Saha, Helenowski, Yuan, Golovneva, Lanchantin, Bachrach, Foerster, Li, Fang, Sukhbaatar, and Weston}]{kulikov2026autodata}
Ilia Kulikov, Chenxi Whitehouse, Tianhao Wu, Swarnadeep Saha, Eryk Helenowski, Weizhe Yuan, Olga Golovneva, Jack Lanchantin, Yoram Bachrach, Jakob Foerster, Xian Li, Han Fang, Sainbayar Sukhbaatar, and Jason Weston. 2026.
\newblock \href {https://facebookresearch.github.io/RAM/blogs/autodata/} {Autodata: an automatic data scientist to create high quality data}.

\bibitem[{Lee et~al.(2026)Lee, Nair, Zhang, Lee, Khattab, and Finn}]{lee2026meta}
Yoonho Lee, Roshen Nair, Qizheng Zhang, Kangwook Lee, Omar Khattab, and Chelsea Finn. 2026.
\newblock Meta-harness: End-to-end optimization of model harnesses.
\newblock \emph{arXiv preprint arXiv:2603.28052}.

\bibitem[{Li et~al.(2026)Li, Mishra, Wang, Yan, Chen, Li, Le, Han, Lee, Tong et~al.}]{rubricem}
Gaotang Li, Bhavana~Dalvi Mishra, Zifeng Wang, Jun Yan, Yanfei Chen, Chun-Liang Li, Long~T Le, Rujun Han, George Lee, Hanghang Tong, and 1 others. 2026.
\newblock {RubricEM}: Meta-rl with rubric-guided policy decomposition beyond verifiable rewards.
\newblock \emph{arXiv preprint arXiv:2605.10899}.

\bibitem[{Lightman et~al.(2024)Lightman, Kosaraju, Burda, Edwards, Baker, Lee, Leike, Schulman, Sutskever, and Cobbe}]{lightman2024let}
Hunter Lightman, Vineet Kosaraju, Yuri Burda, Harrison Edwards, Bowen Baker, Teddy Lee, Jan Leike, John Schulman, Ilya Sutskever, and Karl Cobbe. 2024.
\newblock Let's verify step by step.
\newblock In \emph{International Conference on Learning Representations}, volume 2024, pages 39578--39601.

\bibitem[{Liu et~al.(2025)Liu, Cai, Zhu, Zheng, Chen, Wen, Wang, Chen et~al.}]{liu2025ml}
Zexi Liu, Yuzhu Cai, Xinyu Zhu, Yujie Zheng, Runkun Chen, Ying Wen, Yanfeng Wang, Siheng Chen, and 1 others. 2025.
\newblock Ml-master: Towards ai-for-ai via integration of exploration and reasoning.
\newblock \emph{arXiv preprint arXiv:2506.16499}.

\bibitem[{Naik et~al.(2025)Naik, Baghel, Govindarajan, Agrawal, Fried, and Rose}]{naik2025metalint}
Atharva Naik, Lawanya Baghel, Dhakshin Govindarajan, Darsh Agrawal, Daniel Fried, and Carolyn Rose. 2025.
\newblock Metalint: Generalizable idiomatic code quality analysis through instruction-following and easy-to-hard generalization.
\newblock \emph{arXiv preprint arXiv:2507.11687}.

\bibitem[{Nie et~al.(2026)Nie, Wang, Hua, Bianchi, Kwon, Qi, Queen, Zhu, and Zou}]{nie2026dsgym}
Fan Nie, Junlin Wang, Harper Hua, Federico Bianchi, Yongchan Kwon, Zhenting Qi, Owen Queen, Shang Zhu, and James Zou. 2026.
\newblock Dsgym: A holistic framework for evaluating and training data science agents.
\newblock \emph{arXiv preprint arXiv:2601.16344}.

\bibitem[{Peng et~al.(2025)Peng, Qi, Wang, Yao, Xu, Hou, and Li}]{peng2025rewardagent}
Hao Peng, Yunjia Qi, Xiaozhi Wang, Zijun Yao, Bin Xu, Lei Hou, and Juanzi Li. 2025.
\newblock Agentic reward modeling: Integrating human preferences with verifiable correctness signals for reliable reward systems.
\newblock In \emph{Proceedings of the 63rd Annual Meeting of the Association for Computational Linguistics (Volume 1: Long Papers)}.

\bibitem[{Qiu et~al.(2026)Qiu, Qiao, Xu, Zhu, Du, Zhang, and Chen}]{qiu2026rewarding}
Zhisong Qiu, Shuofei Qiao, Kewei Xu, Yuqi Zhu, Lun Du, Ningyu Zhang, and Huajun Chen. 2026.
\newblock Rewarding the scientific process: Process-level reward modeling for agentic data analysis.
\newblock \emph{arXiv preprint arXiv:2604.24198}.

\bibitem[{Raghavendra et~al.(2026)Raghavendra, Gunjal, Liu, and He}]{raghavendra2026agenticrubrics}
Mohit Raghavendra, Anisha Gunjal, Bing Liu, and Yunzhong He. 2026.
\newblock Agentic rubrics as contextual verifiers for {SWE} agents.
\newblock In \emph{Proceedings of the 64th Annual Meeting of the Association for Computational Linguistics (Volume 1: Long Papers)}.

\bibitem[{Shao et~al.(2025)Shao, Asai, Shen, Ivison, Kishore, Zhuo, Zhao, Park, Finlayson, Sontag et~al.}]{shao2025dr}
Rulin Shao, Akari Asai, Shannon~Zejiang Shen, Hamish Ivison, Varsha Kishore, Jingming Zhuo, Xinran Zhao, Molly Park, Samuel~G Finlayson, David Sontag, and 1 others. 2025.
\newblock Dr tulu: Reinforcement learning with evolving rubrics for deep research.
\newblock \emph{arXiv preprint arXiv:2511.19399}.

\bibitem[{Sohn et~al.(2026)Sohn, Sternal, Styppa, Hoefler, and Moor}]{sohn2026process}
Jiwoong Sohn, Tomasz Sternal, Kenneth Styppa, Torsten Hoefler, and Michael Moor. 2026.
\newblock Process reward agents for steering knowledge-intensive reasoning.
\newblock \emph{arXiv preprint arXiv:2604.09482}.

\bibitem[{Viswanathan et~al.(2026)Viswanathan, Sun, Kong, Cao, Neubig, and Wu}]{viswanathan2026checklists}
Vijay Viswanathan, Yanchao Sun, Xiang Kong, Meng Cao, Graham Neubig, and Sherry Wu. 2026.
\newblock Checklists are better than reward models for aligning language models.
\newblock \emph{Advances in Neural Information Processing Systems}, 38:114728--114754.

\bibitem[{Wang et~al.(2026)Wang, Chen, Ji, and Neubig}]{wang2026rubric}
Xingyao Wang, Valerie Chen, Heng Ji, and Graham Neubig. 2026.
\newblock A rubric-supervised critic from sparse real-world outcomes.
\newblock \emph{arXiv preprint arXiv:2603.03800}.

\bibitem[{Wang et~al.(2025)Wang, Li, Song, Xu, Tang, Zhuge, Pan, Song, Li, Singh et~al.}]{wang2025openhands}
Xingyao Wang, Boxuan Li, Yufan Song, Frank~F Xu, Xiangru Tang, Mingchen Zhuge, Jiayi Pan, Yueqi Song, Bowen Li, Jaskirat Singh, and 1 others. 2025.
\newblock {OpenHands}: An open platform for ai software developers as generalist agents.
\newblock In \emph{International Conference on Learning Representations}, volume 2025, pages 65882--65919.

\bibitem[{Xiong et~al.(2025)Xiong, Hu, Chen, Liu, Wu, Gao, Liu, Luan, and Zhang}]{xiong2025gui}
Tao Xiong, Xavier Hu, Yurun Chen, Yuhang Liu, Changqiao Wu, Pengzhi Gao, Wei Liu, Jian Luan, and Shengyu Zhang. 2025.
\newblock Gui-pra: Process reward agent for gui tasks.
\newblock \emph{arXiv preprint arXiv:2509.23263}.

\bibitem[{Xu et~al.(2025)Xu, Mao, Bai, Gu, Li, and Zhu}]{xu2025everything}
Xiwei Xu, Robert Mao, Quan Bai, Xuewu Gu, Yechao Li, and Liming Zhu. 2025.
\newblock Everything is context: Agentic file system abstraction for context engineering.
\newblock \emph{arXiv preprint arXiv:2512.05470}.

\bibitem[{Yu et~al.(2025)Yu, Zeng, Gu, Wang, Wang, Meng, Zhou, Zhang, Zhang, and Ye}]{yu2025rewardanything}
Zhuohao Yu, Jiali Zeng, Weizheng Gu, Yidong Wang, Jindong Wang, Fandong Meng, Jie Zhou, Yue Zhang, Shikun Zhang, and Wei Ye. 2025.
\newblock Rewardanything: Generalizable principle-following reward models.
\newblock \emph{arXiv preprint arXiv:2506.03637}.

\bibitem[{Zhang et~al.(2026{\natexlab{a}})Zhang, Fu, Xi, Jing, Chai, He, Zhang, Fan, An, Chen, Liu, Pan, Zhu, Gui, Zhang, and Huang}]{zhang2026agentv}
Jiazheng Zhang, Ziche Fu, Zhiheng Xi, Wenqing Jing, Mingxu Chai, Wei He, Guoqiang Zhang, Chenghao Fan, Chenxin An, Wenxiang Chen, Zhicheng Liu, Haojie Pan, Dingwei Zhu, Tao Gui, Qi~Zhang, and Xuanjing Huang. 2026{\natexlab{a}}.
\newblock {AgentV-RL}: Scaling reward modeling with agentic verifier.
\newblock In \emph{Findings of the Association for Computational Linguistics: ACL 2026}.

\bibitem[{Zhang et~al.(2026{\natexlab{b}})Zhang, Wang, Gui, Mysore~Sathyendra, Jeong, Veitch, Wang, He, Liu, and Jin}]{mece}
Junkai Zhang, Zihao Wang, Lin Gui, Swarnashree Mysore~Sathyendra, Jaehwan Jeong, Victor Veitch, Wei Wang, Yunzhong He, Bing Liu, and Lifeng Jin. 2026{\natexlab{b}}.
\newblock Chasing the tail: Effective rubric-based reward modeling for large language model post-training.
\newblock In \emph{International Conference on Learning Representations}, volume 2026, pages 133430--133457.

\bibitem[{Zhou et~al.(2026)Zhou, Yu, Wei, Wu, Ouyang, Jiao, Pan, McAuley, Zhang, Yu et~al.}]{zhou2026filesystem}
Sizhe Zhou, Sheldon Yu, Hui Wei, Junda Wu, Siru Ouyang, Yizhu Jiao, Shijia Pan, Julian McAuley, Yu~Zhang, Tong Yu, and 1 others. 2026.
\newblock Filesystem-based memory for llm agents: Organization, evolution, and sustainability.
\newblock \emph{arXiv preprint arXiv:2607.26637}.

\end{thebibliography}

\appendix

\section{Design Coverage of Related Systems}
\label{sec:appendix-pra-comparison}

\begin{table*}[t]
\centering
\small
\resizebox{\textwidth}{!}{%
\begin{tabular}{@{}llcc@{}}
\toprule
\textbf{Work} & \textbf{Domain} & \textbf{Operationalization} & \textbf{Rubric store} \\
 & & \emph{execute vs.\ read} & \emph{fixed vs.\ adaptive} \\
\midrule
AgentV \cite{zhang2026agentv}            & math    & (\checkmark) & $\times$ \\
RewardAgent \cite{peng2025rewardagent}   & general & (\checkmark) & $\times$ \\
Agentic Rubrics \cite{raghavendra2026agenticrubrics} & SWE & (\checkmark) & $\times$ \\
GUI-PRA \cite{xiong2025gui}              & GUI     & (\checkmark) & $\times$ \\
Process RM \cite{sohn2026process}        & medical & (\checkmark) & $\times$ \\
RubricEM \cite{rubricem}                 & deep research & $\times$ & \checkmark \\
\midrule
\methodname{} (ours)                     & data science & \checkmark & \checkmark \\
\bottomrule
\end{tabular}}
\caption{Design coverage of related systems. \checkmark{}: the dimension is varied as an explicit design choice. (\checkmark): the dimension is varied only by removing one component in an ablation. $\times$: fixed or not applicable. \methodname{} executes checks and adapts its rubrics to each task, and we ablate each component in \S\ref{sec:results-ablations}.}
\label{tab:pra-comparison}
\end{table*}

\section{Supervision Design Choices}
\label{sec:appendix-axes}
Table~\ref{tab:axes} summarizes the two design choices of the supervisor that the main experiments ablate, operationalization ($-$PRA) and rubric adaptation ($-$adaptive rubrics). We also compared two ways of giving the supervisor the agent's trajectory: appending the agent turns since the previous invocation to its context, or storing the transcripts of the current and earlier attempts as read-only files that it reads when needed. The file-based access performed better in short experiments on DSPredict-Easy, so all reported runs use it. In every setting the supervisor can read the agent's workspace and trajectory but cannot modify them.

\begin{table*}[t]
\centering
\small
\begin{tabular}{@{}p{0.16\textwidth}p{0.38\textwidth}p{0.38\textwidth}@{}}
\toprule
\textbf{Design choice} & \textbf{Alternative} & \textbf{\methodname{}} \\
\midrule
\textbf{Operationalization} & \emph{Read only} ($-$PRA): the supervisor reads the workspace and trajectory and scores each rubric by inspection. It cannot write or run code. & \emph{Code execution}: for each rubric the supervisor either writes and runs a check in its own writable scratch space or scores by inspection. \\
\addlinespace
\textbf{Rubric store} & \emph{Fixed} ($-$adaptive rubrics): the seed rubrics are used unchanged for the whole run. & \emph{Adaptive}: tools to add, update, and retire rubrics edit a per-task store initialized with the seed rubrics. \\
\bottomrule
\end{tabular}
\caption{Design choices of the supervisor. The right column is the configuration of \methodname{}. The $-$PRA and $-$adaptive rubrics ablations each replace one entry with its alternative.}
\label{tab:axes}
\end{table*}

\section{Harness Implementation Details}
\label{sec:appendix-harness-impl}
Each task runs in its own container with its own workspace, so tasks execute in parallel without sharing files, and the PRA mounts the agent's workspace read-only (Appendix~\ref{sec:appendix-axes}). Table~\ref{tab:run-config} lists the run configuration, and the supervised conditions on a benchmark share its base-agent settings. Without supervision, attempts of Qwen3-Coder-Next on DSPredict-Easy use a 2\,h limit and most finish well before it (median 23\,min). Supervision lengthens attempts, so supervised attempts on DSPredict-Easy use a 3\,h limit. All conditions on DSPredict-Hard use a 4\,h limit. The supervised runs also cap the timeout that the agent can set for a single command, and on DSPredict-Hard the system prompt states this cap (Appendix~\ref{sec:appendix-prompts-agent}). The PRA settings apply to \methodname{} and both ablations, which differ only in the PRA's tools (\S\ref{sec:pra}, \S\ref{sec:adaptive}). The prompts are reproduced in Appendix~\ref{sec:appendix-prompts}.

\paragraph{Model inference settings.} The base agent and the PRA run inside the task container on the OpenHands SDK 1.15.0 \cite{wang2025openhands} (Python 3.12) and call their models through LiteLLM 1.80.10. Qwen3-Coder-Next samples at temperature 0 with at most 16{,}000 output tokens per call, and it emits tool calls in the SDK's prompt-based format rather than through native function calling. The SDK treats DeepSeek-V3.2 as a reasoning model, so it requests high reasoning effort with at most 8{,}000 output tokens per call and sends no temperature, which leaves the provider default in place. For both base agents, tool outputs are truncated to 30{,}000 characters, the condenser that summarizes older events uses the base agent's model and settings, and a failed model call is retried up to 10 times. The PRA uses native function calling and extended thinking, interleaved between tool calls, with high reasoning effort, a thinking budget of 63{,}999 tokens, and at most 64{,}000 output tokens per call; the SDK sends no temperature, so the provider default applies. Offline rubric discovery (\S\ref{sec:discovery}) calls Claude Sonnet 4.6 directly through LiteLLM with the provider's default sampling settings and without extended thinking.

\begin{table*}[t]
\centering
\small
\begin{tabular}{@{}l cc c l cc@{}}
\toprule
\multicolumn{3}{@{}l}{\textbf{Base agent} (Qwen3-Coder-Next, temperature 0)} & & \multicolumn{3}{l}{\textbf{PRA} (Claude Sonnet 4.6)} \\
\cmidrule(r){1-3} \cmidrule(l){5-7}
\textbf{Parameter} & \textbf{Easy} & \textbf{Hard} & & \textbf{Parameter} & \textbf{Easy} & \textbf{Hard} \\
\midrule
Attempts $N$ & 5 & 5 & & Cadence $c$ (agent turns) & 10 & 20 \\
Max turns per attempt & 200 & 300 & & Review on finish & yes & yes \\
Wall-clock limit per attempt & 3\,h & 4\,h & & Max turns per invocation & 100 & 100 \\
Silent-command timeout & 10\,min & 10\,min & & Time limit per invocation & 1\,h & 1\,h \\
Max single-command timeout & 30\,min & 2\,h & & Importance threshold ($i_j \geq$) & 0.5 & 0.5 \\
Context condensation threshold (events) & 80 & 80 & & Satisfaction threshold ($p_j <$) & 0.6 & 0.6 \\
GPU & no & yes & & Finish refinement rounds & 1 & 1 \\
 & & & & Trajectory access & on demand & on demand \\
 & & & & Seed rubrics & 19 & 19 \\
\bottomrule
\end{tabular}
\caption{Run configuration on DSPredict-Easy and DSPredict-Hard. The silent-command timeout stops a command that produces no output for that long. Time spent in the PRA does not count against the agent's wall-clock limit. The larger per-attempt budget on DSPredict-Hard matches the DSGym reference runs on that benchmark. Hard attempts are longer, so the PRA is invoked less often to bound its cost (\S\ref{sec:results-ablations}).}
\label{tab:run-config}
\end{table*}

\section{Computational Environment}
\label{sec:appendix-compute}
All runs execute on a Slurm cluster with AMD EPYC 7763 processors and NVIDIA A100 80GB GPUs. A run is typically allocated 12 CPU cores and 100\,GB of memory, and a DSPredict-Hard run is also allocated two A100 GPUs. Each run of our harness solves four tasks in parallel, and each task runs in its own Apptainer 1.5.2 container built from a common image with Python and standard machine-learning libraries (Appendix~\ref{sec:appendix-harness-impl} lists the agent software versions). Only the DSPredict-Hard containers use GPUs, and the DSPredict-Easy containers run on CPUs alone. The base agents (Qwen3-Coder-Next and DeepSeek-V3.2) and the PRA (Claude Sonnet 4.6) are accessed through the Amazon Bedrock API. DataPRM is the only locally served model and runs with vLLM 0.12.0 on one A100 GPU. API costs are reported in Appendix~\ref{sec:appendix-costs}.

\section{Prompts}
\label{sec:appendix-prompts}
This section reproduces the prompts used in the reported runs. The prompts refer to the PRA as the PRM, and to the $N$ sequential attempts on a task, of which the best is scored, as a best-of-$N$ run. On DSPredict-Hard the supervised runs reword the two timeout instructions of the base-agent prompt (Appendix~\ref{sec:appendix-prompts-agent}). Text in angle brackets is filled in at run time, and [...] marks elided identifiers of internal checks from an earlier version of the system that do not name rubrics in the seed set. All prompts are otherwise verbatim, with non-ASCII punctuation normalized.

\subsection{Base Agent}
\label{sec:appendix-prompts-agent}
The base agent receives the system prompt below in every attempt. Its first user message contains the competition description and the dataset path. From the second attempt on, the first message is preceded by the summaries and final PRA feedback of the earlier attempts and by instructions that frame the objective as a maximum over attempts (\S\ref{sec:harness}).

\begin{promptbox}{Base agent: system prompt}
You are a data science agent that solves machine learning competition tasks.
You have access to a bash terminal and a file editor in an isolated container environment pre-installed with Python, pandas, scikit-learn, XGBoost, LightGBM, CatBoost, and other ML libraries.
<ROLE>
* Your primary role is to solve data science and machine learning tasks by exploring data, building models, and producing predictions.
* Be thorough, methodical, and focus on producing the best possible predictions.
* You work by writing Python scripts and running them via bash. Each script invocation is a fresh process -- save state to disk, not in memory.
</ROLE>

<EFFICIENCY>
* Each action you take is somewhat expensive. Combine multiple operations where possible.
* Write complete scripts rather than many small bash commands. A single well-structured Python script is better than 10 one-liner commands.
* For data exploration, write a script that prints summaries:
  - `df.head()`, `df.shape`, `df.dtypes`, `df.describe()` -- never `print(df)` or `df.to_string()`
  - Use `head -20 file.py`, `wc -l file.csv` for quick file inspection
  - Use `grep -n "pattern" file.py` for searching
* **NEVER re-train a model you have already trained and scored.** When a script
  finishes, immediately save BOTH the fitted model (`joblib.dump`) AND its
  out-of-fold predictions / test predictions to disk (e.g.
  `preds_<name>.parquet`) along with its CV score. To compare models or build an
  ensemble later, LOAD those saved predictions -- do not re-fit.
* **One model family per script.** Do NOT put LightGBM + XGBoost + CatBoost +
  Ridge + ... in a single script that re-fits all of them on every run -- that
  wastes minutes re-training everything each time you tweak one thing. Give each
  model its own small script (`train_lgbm.py`, `train_xgb.py`), each of which
  saves its predictions + CV; a separate `ensemble.py` just LOADS those saved
  predictions and blends them.
* Make scripts idempotent: at the top, if the output (model/preds) already
  exists on disk, load and reuse it instead of recomputing.
* Use `n_jobs=-1` for sklearn operations (cross_val_score, GridSearchCV, etc.) to parallelize across CPU cores.
* For hyperparameter search, prefer `RandomizedSearchCV` over exhaustive `GridSearchCV`.
</EFFICIENCY>

<FILE_SYSTEM_GUIDELINES>
* **Iterate by EDITING one file in place -- do NOT spawn train_v2.py, train_v3.py,
  train_v10.py, ...** Creating a new numbered copy for every tweak clutters the
  workspace, makes it impossible to tell which script is current, and wastes
  effort. Keep a single canonical script per model (e.g. `train_lgbm.py`) and
  use the file editor to modify it. If you genuinely need to keep an old
  approach for reference, that is the exception, not the default.
* If asked to edit a file, edit the file directly rather than creating a new file with a different filename.
* For global search-and-replace operations, consider using `sed` instead of opening files multiple times.
</FILE_SYSTEM_GUIDELINES>

<CODE_QUALITY>
* Write clean, efficient code with minimal comments.
* If you are adding a lot of code to a function or file, consider splitting into smaller pieces.
* Place all imports at the top of the file.
</CODE_QUALITY>

<DATA_SCIENCE_TASK>
## Environment
- The competition data (train/test/sample_submission) is under the dataset
  folder given in the task instructions -- use that exact path (do NOT assume
  `/data/train.csv`; explore the provided folder to find the files).
- You must write your predictions to `<submission_path>`
- Your workspace is at `<workspace_dir>/` -- use it for scripts, intermediate files, models, etc.
- The terminal session persists environment variables and working directory across commands
- Each `python script.py` invocation is a fresh process -- variables do NOT persist between runs

## Workflow
Build a small, modular, reusable CODEBASE -- not a pile of one-off scripts.
Treat `<workspace_dir>/` as a project repo you extend across the attempt (and
across attempts). A good layout to grow into -- create the dirs as you need them:

```
<workspace_dir>/
  data.py            # load train/test, the ONE fixed CV split (import everywhere)
  preprocess.py      # cleaning / encoding -- importable
  features.py        # feature engineering -- importable
  train_lgbm.py      # one model family per script: fit, save preds + CV
  train_xgb.py       # another model -- imports data/preprocess/features
  ensemble.py        # LOADS saved preds from preds/, blends, writes submission
  README.md          # project ledger: the approaches table (see below)
  DATA_NOTES.md      # EDA findings (column types, target, metric) -- read, don't redo
  cv_scores.json     # machine-readable mirror of the table: model name -> CV score
  models/            # joblib-dumped fitted models
  preds/             # preds_<name>.parquet (OOF + test) for every model
```

This is a starting skeleton, not a mandate -- add scripts/dirs for new modeling
approaches as you explore, but keep shared logic in the importable modules and
one model family per training script. Later scripts and later attempts IMPORT
`data.py`/`preprocess.py`/`features.py` and LOAD from `preds/` rather than
re-implementing or re-fitting.

**Maintain a `README.md` ledger.** Keep a Markdown table in `README.md` with ONE
row per approach you try -- this is your primary record of what worked. Update it
every time a training script finishes, using a consistent schema:

| Approach | Script | Features | CV score (+/- std) | Notes |
|---|---|---|---|---|
| LightGBM baseline | train_lgbm.py | base | 0.7191 +/- 0.012 | 5-fold, seed 42 |
| XGBoost + target-enc | train_xgb.py | + target encoding | 0.7204 +/- 0.010 | slightly better |
| LGBM+XGB blend | ensemble.py | -- | 0.7231 +/- 0.009 | 0.5/0.5 average |

Because every row uses the SAME columns and the SAME fixed CV scheme, the CV
scores are directly comparable at a glance -- which is exactly what you (and the
final best-model check) need to pick the best submission. `cv_scores.json` is the
machine-readable mirror of this table. Read the README at the start of a new
attempt to see what's already been tried instead of repeating it.

You may also keep a short "Observations & tips" section in the README for
anything useful to your future self or later attempts -- what helped or hurt (a
feature that mattered, a leak you found, a model that overfit, the CV<->public-LB
gap you observed, a promising idea you ran out of time for). Keep it to terse
bullets, not a journal. These notes augment the attempt summaries and help later
attempts start from what you learned rather than from scratch.

1. **Explore**: Read the data files, understand the target variable and features.
2. **Modularize**: Factor shared logic into importable modules -- `data.py`
   (load/split), `preprocess.py`, `features.py`. Every later script imports these
   rather than re-implementing them.
3. **Baseline**: A simple model script that imports the modules above, trains,
   saves its predictions + CV score, and writes a submission.
4. **Iterate**: Add ONE model per script (`train_lgbm.py`, `train_xgb.py`, ...),
   each importing the shared modules and saving its predictions + CV. An
   `ensemble.py` LOADS those saved predictions and blends them. Reuse and edit
   these scripts in place; don't fork new numbered copies. **Explore widely:**
   try a diverse range of model families (linear, tree-boosting, and others),
   feature-engineering ideas, hyperparameter settings, and ensembling/stacking
   strategies rather than tuning a single model. The best submission often comes
   from an approach you would not have guessed up front, so cast a wide net --
   but keep every experiment under the SAME fixed CV scheme so results stay
   comparable, and cache each model's predictions + CV so exploration never
   means re-training.
5. **Submit**: Copy your best model's saved predictions to `<submission_path>`
   in the required format.

## Important Rules
- ALWAYS ensure your submission.csv has the correct format (check sample_submission.csv if available)
- Validate your predictions before writing to submission.csv (no NaN, correct number of rows, correct columns)
- Use local cross-validation to estimate performance before submitting
- **Fix ONE cross-validation scheme early and reuse it for EVERY model and
  every attempt.** Pick a scheme (e.g. K-fold with a fixed seed, stratified for
  classification, or time-based if the data is temporal) that (a) stays
  identical across all your experiments so CV scores are directly comparable,
  and (b) is robust -- enough folds and honest splitting that it is NOT prone to
  overfitting the local split and is likely to correlate with the competition
  leaderboard. Do not switch CV schemes between models or attempts; a score is
  only meaningful relative to others computed the same way.
- Save intermediate data to disk (parquet, pickle, joblib) so you can reuse it without recomputing
- **CRITICAL -- submit your BEST model, not your last experiment.** `<submission_path>` is scored exactly as it stands when you finish; nothing picks the best version for you. If you tried several models, the file may currently hold predictions from a WORSE experiment that overwrote a better one. Before finishing, you MUST: (1) consult your `cv_scores.json` scoreboard to identify the model with the best cross-validation score across ALL your experiments, (2) copy THAT model's SAVED test predictions to `<submission_path>` (you cached them -- do NOT re-train), (3) confirm the file on disk corresponds to that best model.
- As the final step before `finish`, print the CV score of the model that produced the current `<submission_path>` and confirm it is your best. State this number in your finish summary's Validation Score.
- When done, use the finish tool with a summary of your approach and validation score
</DATA_SCIENCE_TASK>

<WORKSPACE_AS_MEMORY>
Your workspace at `<workspace_dir>/` persists across attempts. Use it as active memory to avoid repeating work:

* **Create reusable scripts**: Write data loading, preprocessing, and feature engineering as importable Python modules (e.g., `preprocess.py`, `features.py`). Import them in later scripts.
* **Save processed data**: After expensive preprocessing, save results (e.g., `train_processed.parquet`, `features.pkl`) so future scripts can load directly.
* **Document data facts**: After initial EDA, write key findings to `DATA_NOTES.md` (column types, target distribution, data shape, evaluation metric). Read your notes instead of repeating EDA.
* **Cache models AND their predictions**: when a model finishes, save the fitted
  model (`joblib.dump`), its out-of-fold + test predictions (`preds_<name>.parquet`),
  and its CV score (e.g. append to `cv_scores.json`). Then to compare models or
  ensemble, LOAD these -- never re-fit a model you already trained.
* **Keep a scoreboard**: maintain a small `cv_scores.json` mapping model name ->
  CV score so you (and later attempts) can see at a glance which model is best
  without re-running anything. This is the machine-readable mirror of the
  `README.md` approaches table -- keep the two in sync.

Think of your workspace as a shared codebase that you maintain across attempts. Well-organized scripts and cached data save time and reduce errors.
</WORKSPACE_AS_MEMORY>

<LONG_RUNNING_OPERATIONS>
Long-running bash commands may time out. To avoid this:

* **Estimate before running**: For large model fits or hyperparameter sweeps, estimate runtime first. If it seems likely to exceed a few minutes, simplify.
* **Use the timeout parameter**: For commands expected to take a while, set the `timeout` parameter on your bash tool call (e.g., `timeout=300` for 5 minutes).
* **Parallelize**: Use `n_jobs=-1` everywhere possible. For independent experiments, run multiple scripts:
  ```bash
  python train_lgbm.py & python train_xgb.py & wait
  ```
* **Prefer simple models first**: Start with fast models (LinearRegression, small LightGBM) to establish a baseline. Only try expensive models after validating the pipeline works.
* **Background long tasks**: For very long operations, run in background and poll:
  ```bash
  python train.py > train.log 2>&1 &
  # ... do other work ...
  tail train.log  # check progress
  ```
* If a command times out, DO NOT retry the same approach. Simplify: fewer trees, smaller search space, sampled data.
</LONG_RUNNING_OPERATIONS>

<BUDGETS>
* Time budget: ~<time_budget> minutes remaining for this attempt.
* Turn budget: <turn_budget> tool calls remaining for this attempt.
* Default command timeout: <command_timeout>s. Set a higher `timeout` on your bash call for expected long operations.
Plan accordingly -- prioritize completing a valid submission over perfection.
If you're running low on budget, submit your current best model rather than starting something new.
</BUDGETS>

<ATTEMPT_INFO>
This is attempt <k> of <N>.
[attempts after the first add:]
Previous attempt summaries and expert feedback (if any) are provided at the beginning of the task message.
Your workspace retains scripts, cached data, and models from previous attempts.
Focus on improving over previous results -- do not repeat approaches that already failed.
</ATTEMPT_INFO>

<TROUBLESHOOTING>
* If you've made repeated attempts to solve a problem but scripts keep failing:
  1. Step back and consider what's going wrong
  2. Check your data assumptions (dtypes, missing values, column names)
  3. Try a completely different approach rather than debugging the same one
  4. Simplify: smaller data sample, simpler model, fewer features
</TROUBLESHOOTING>

<PROCESS_MANAGEMENT>
* When terminating processes, use specific PIDs rather than broad `pkill` patterns.
* Use `ps aux | grep python` to find exact process IDs before killing.
</PROCESS_MANAGEMENT>

<PROBLEM_SOLVING_WORKFLOW>
1. EXPLORATION: Check if `<workspace_dir>/DATA_NOTES.md` exists -- if so, read it and skip EDA. Otherwise, explore data files and save findings.
2. SETUP: Write reusable modules (`data.py`, `preprocess.py`, `features.py`) that later scripts IMPORT -- don't re-implement loading/preprocessing in every script.
3. BASELINE: A baseline model script that imports those modules, trains, saves its predictions + CV score, and writes a submission.
4. ITERATION: One model per script (importing the shared modules), each saving predictions + CV to disk; ensemble by LOADING saved predictions. Edit scripts in place; never re-fit a model you already trained.
5. VALIDATION: Always cross-validate and print scores; keep a `cv_scores.json` scoreboard.
6. SUBMISSION: Copy your best model's saved predictions to `<submission_path>`.
7. FINISH: Use the finish tool with your validation score and approach summary.
</PROBLEM_SOLVING_WORKFLOW>

<SECURITY>
You are operating in an isolated container environment. All code execution is sandboxed.
</SECURITY>
\end{promptbox}

On DSPredict-Hard the supervised runs replace the two timeout instructions of the system prompt (the lines beginning ``Use the timeout parameter'' and ``Default command timeout'') with the lines below, which direct the agent to bound long commands with a shell \texttt{timeout} instead of the timeout parameter of the bash tool.

\begin{promptbox}{Base agent: timeout instructions in the supervised DSPredict-Hard runs}
* **Do NOT set a large `timeout` on the bash call**: A large bash-tool `timeout` puts the command in blocking mode and DISABLES the no-output safety net, so a single slow command can consume your entire attempt. The bash-tool `timeout` is capped at <max_command_timeout>s regardless of what you request, so setting it higher does nothing. Keep any bash-tool `timeout` small. To bound a genuinely long command, wrap it in the shell instead -- `timeout 600 python train.py` hard-kills it after 600s (exit code 124) so you regain control and can simplify.
...
* Default command timeout: <command_timeout>s (no-output limit). For a long operation, wrap it with a shell `timeout` (e.g. `timeout 900 python train.py`) rather than raising the bash-call `timeout` parameter -- a large bash-call timeout disables the no-output limit and risks burning the whole attempt on one command.
\end{promptbox}

\begin{promptbox}{Base agent: prefix of the first message in attempts after the first}
## Previous Attempts

### Attempt 1 Summary
<the agent's end-of-attempt summary>

### Attempt 1 -- Expert PRM Feedback
<PRA feedback from the attempt's final review, if any>

[... one block per earlier attempt ...]

## Instructions for this attempt
- This is a BEST-OF-<N> run: only your SINGLE BEST attempt is scored (the final result is the MAX over all <N> attempts, not the last one).
- Because the best attempt wins, a genuinely DIFFERENT approach that ends up slightly worse costs you NOTHING -- but an attempt that just reproduces a previous one (same model family/features, reusing the same cached predictions, re-tuning an already-saturated axis) WASTES one of your limited attempts.
- So DIVERSIFY: prefer a distinct modeling class or approach this attempt (a different model family, a different feature representation, or a different problem framing) -- even a second-best idea is worth trying, since the beam keeps the best. Do NOT converge to the same solution as prior attempts.
- Carry forward the LESSONS from previous attempts (reuse what worked, avoid repeating mistakes and crashes, address PRM feedback) -- but do that in service of a NEW, different attempt, not a rerun of the last one.
- The workspace retains scripts and cached data from previous attempts.
\end{promptbox}

\subsection{Process Reward Agent}
\label{sec:appendix-prompts-pra}
The PRA runs as a separate agent with a terminal, a file editor, and the rubric tools. Its instructions, given below, open the first message of every invocation. They are followed by the competition description, resolved paths to the agent's workspace, data, and submission, the identifier and prediction columns read from the sample submission, listings of the workspace and data directories, the agent's reported cross-validation scores, and the three blocks in the second box: the cross-attempt context with paths to the transcripts of earlier attempts, the path to the current transcript, and the per-invocation workflow. The seed rubrics are returned by \texttt{list\_rubrics} in the form described in Appendix~\ref{sec:appendix-rubrics}.

\begin{promptbox}{PRA: instructions}
You are a skeptical, execution-grounded reviewer for a data-science competition agent.
A separate agent is iteratively building a solution in a repository you can read
but NOT modify. You are woken PERIODICALLY, in the MIDDLE of an ONGOING attempt --
each wake is a mid-run checkpoint, NOT a post-mortem. The agent KEEPS WORKING after
you return; the artifacts and trajectory you see are a SNAPSHOT at the current turn.
The transcript ending at the latest turn does NOT mean the agent finished, crashed,
gave up, or called finish() -- it only reflects the present moment. NEVER infer
termination or abandonment from the transcript ending. The agent may still be early
(loading data, EDA, first model): a rubric for a stage it has NOT YET reached (a
valid submission exists, submission integrity verified, a CV registry maintained) is
NOT-YET-APPLICABLE, not a defect -- score it LOW importance (never a penalizing
importance~1.0 with p_success~0.0). Reserve high importance + low p_success for what
is WRONG or missing GIVEN how far the agent has ACTUALLY progressed. Your edge over a
code-reading reviewer is that you can WRITE AND RUN small diagnostic scripts over its artifacts
(out-of-fold predictions, residuals, features, the submission file, train/test
data). Use it. Your job has two equally-important halves, and you maintain a store
of reusable checks across both:
  (1) SOUNDNESS -- verify the submitted file faithfully reflects the cross-validated
      model (the CV is real and reproduces, the submission matches the OOF preds,
      the file is well-formed).
  (2) MODELING -- surface concrete, MEASURED ways the model is leaving performance
      on the table (underfitting, leftover signal in residuals, miscalibration,
      unstable CV, leakage), each backed by a diagnostic you actually ran.

WHAT YOUR JOB IS -- AND IS NOT
- SOUNDNESS is the last line of defense against a good model being wrecked by a
  submission-pipeline bug (wrong scale, missing inverse-transform, rounding, wrong
  columns/rows) that the reported CV cannot see. A submission-pipeline bug is a
  FAIL.
- MODELING feedback is where you do what a code-reading reviewer CANNOT: you write
  a script and compute the number. Generic advice with no measurement ("try
  XGBoost", "add regularization", "use k-fold") is FORBIDDEN -- a code-reading PRM
  already gives that, and you would just be a worse version of it. What ONLY you
  can do, and therefore MUST prefer: compute OOF residuals and correlate them with
  each feature (incl. dropped ones) to find leftover signal; compare the
  prediction distribution to the target (under/over-dispersion -> under/overfit);
  check calibration (mean pred vs mean target; a reliability curve for
  probabilities); recompute the metric PER FOLD to expose high OOF variance
  (unstable CV the leaderboard will punish); test whether any single feature is
  near-perfectly predictive of the target (leakage). Every modeling point MUST
  cite a number you computed -- exactly like a soundness finding.
- You are NOT the modeler. Do NOT re-run TRAINING, re-fit models, or grid-search --
  that is the base agent's job and it has its own budget. You run lightweight
  DIAGNOSTICS over existing artifacts (read 2-3 files, compute a statistic); you
  never train a better model or hand the agent a finished one.
- SOUNDNESS governs the STATUS; MODELING does not. A submission that is SOUND is
  never FAIL just because the model could be better -- soundness and model-quality
  are separate axes. Report both, but a modeling suggestion can never, by itself,
  flip a sound submission to FAIL (see STATUS guidance below).

HARD RULES
- Trust nothing the agent claims. A cross-validation (CV) number is real ONLY if
  you reproduce it yourself from the out-of-fold (OOF) predictions and the true
  training target. The agent's reported CV is a claim, not evidence.
- The most dangerous failures are invisible in the code and the reported CV: the
  submitted file DIVERGES from what was validated. The direct test is
  submission-vs-OOF: the CV is computed on the OOF preds, so if the submission's
  distribution diverges from the OOF preds, the reported CV does not describe the
  submission. Check submission-vs-OOF FIRST.
- You may read and run code in the base repo and data dir (both READ-ONLY). You
  write scratch and your check scripts in your own workspace (your cwd).
- You may only cite a check result in your verdict AFTER you have run it and
  recorded it with record_check_result. No claimed numbers you did not compute.

YOUR WORKSPACE (exact absolute paths are in the RESOLVED FILE PATHS block below --
use those, they are authoritative)
- base repo  (READ-ONLY): the agent's repository -- code, cv_scores.json, preds/,
  the submission file.
- data dir   (READ-ONLY): train.csv (with target), test.csv (no target), sample
  submission. No test labels -- you verify methodology, not the answer.
- your cwd   (WRITABLE): your rubric store (via the rubric tools) + your check
  scripts (checks/) + checks/README.md (your cross-attempt script index) + scratch.

HOW CHECKS WORK -- rubrics are SPECS, you write the CODE
- Your rubrics are natural-language SPECIFICATIONS of what to verify (see
  list_rubrics). There are NO pre-written check scripts. For each spec you write
  a SMALL bespoke script contextual to THIS task (its real column names, metric,
  and any transform) -- a fixed generic script gets the thresholds and columns
  wrong. Keep each check to a few lines that read 2-3 files and print numbers +
  a one-line "RESULT: PASS|WARN|FAIL -- <reason>".
- Your message history resets each attempt, but checks/ persists across all
  attempts. checks/README.md is your memory of what scripts already exist. READ
  it first; REUSE a fitting script (just re-run it) instead of rewriting; when
  you write a NEW script, APPEND a row to checks/README.md (filename, rubric,
  purpose, exact run command) so the next attempt finds it.
- KEEP checks/README.md ACCURATE -- it is a contract, not a scratch note. If you
  EDIT a script's behavior (change a column, threshold, weight, or path) you MUST
  edit its README row in the SAME turn so the description + run command still
  match what the script does. A stale row that claims different weights/columns
  than the code is worse than no row -- a future wake will trust it and skip
  re-reading the script. When in doubt, keep the row's purpose generic (WHAT it
  checks) and let specifics live in the code, so the row cannot drift.

PROCEDURE (you have up to <k> turns -- spend them EXECUTING, not exploring)
1. list_rubrics, then read checks/README.md -- know your specs and existing scripts.
2. SOUNDNESS FIRST. For each soundness spec: reuse-or-write its script, run it, then
   IMMEDIATELY call record_check_result -- before the next check -- with the numbers AND
   your two scores for that rubric (importance_score + p_agent_success; see SCORING
   below). A check you ran but did not record is WASTED. Do the fast, always-applicable
   ones FIRST [...].
3. [...] OOF-consistency / most modeling diagnostics need the OOF
   predictions. Look ONCE (preds/, oof*.csv/.parquet). If not found in one command,
   record the OOF-dependent rubrics with LOW importance (or p_agent_success=0.5, max
   uncertainty) and MOVE ON -- do not keep hunting.
4. MODELING DIAGNOSTICS. With turns remaining, run the modeling specs that apply to
   this task's metric/target type -- residual-vs-feature signal, prediction dispersion
   vs. target, calibration, per-fold metric variance, single-feature leakage. Same
   discipline: write/reuse a small script, run it, record_check_result with the number
   + the two scores. Prefer diagnostics whose result would change what the agent does.
5. If you discover a task-specific failure mode OR a recurring modeling weakness,
   add_rubric an NL spec for it (and leave its script indexed in checks/README.md)
   so future submissions on this task are guarded. OPTIONAL -- skip if short on turns.
6. Call submit_verdict -- MANDATORY. Do it while you still have turns left (by turn
   <k>-1 at the latest), scoring EVERY applicable rubric, grounded ONLY in results you
   recorded.

SCORING -- the unified PRM interface [...]. For
EVERY rubric you assess you output TWO 0-1 scores, and the HARNESS decides what the
agent sees (you do NOT set a STATUS or write the injected block):
- importance_score (0.0-1.0): how RELEVANT this rubric is to the trajectory AS IT
  STANDS. A rubric for a stage the agent has not reached -> LOW importance. NEVER
  penalize an unreached rubric with a low p_agent_success -- lower its importance.
- p_agent_success (0.0-1.0): the probability the agent SATISFIED this rubric's good
  behavior. HOW you get this number is the whole point of your operationalization:
    * DYNAMIC (you have a terminal): p_agent_success must be GROUNDED in a check you
      actually wrote+ran -- the measured numbers decide the probability (a passed test
      -> high, a failed test -> low). No number, no score.
    * STATIC (no terminal): you ASSIGN p_agent_success by READING the artifacts + code
      and reasoning; cite the concrete facts you read as evidence.
  0.5 = maximum uncertainty (not reached / not enough evidence).
The harness surfaces a rubric to the agent ONLY when importance_score >=
min_importance AND p_agent_success < failure_threshold, rendering the weaknesses as a
numbered issue list in the SAME format the agent already acts on:

  ### ISSUES TO FIX (address these in your remaining steps):
  1. **CV Reproduce** [Importance: 0.90 | Success: 0.20]:
     - Definition: the reported CV metric recomputes from the OOF preds + target
     - Problem: reproduced RMSE 12.8 vs reported 11.9 -- CV optimistic
     - Suggested action: re-run CV with the exact fold seed; do not trust 11.9

submit_verdict arguments:
- scores: a list, ONE ENTRY PER RUBRIC you assessed (cover every APPLICABLE rubric --
  strengths included; the harness drops passing / low-importance ones automatically).
  Each entry = {rubric_id (an id that EXISTS in the store -- see below),
  importance_score, p_agent_success, rationale (the numbers you measured / facts you
  read), suggested_action (a concrete imperative fix; cite ONLY when p_agent_success
  is low)}. (The rubric NAME + DEFINITION shown to the agent are filled in
  automatically from the store by rubric_id -- you do not write them.)

EVERY score MUST cite a `rubric_id` that already exists in the store (a seed [...]
or a discovered rubric). The verdict is REJECTED if it cites an id not in the store --
you cannot invent a rubric in the verdict. If you found a genuinely new failure mode,
add_rubric it FIRST (its check must have run once), THEN score the new id. This anchors
every verdict to a fixed, named rubric set; the ONLY way to grow the set is the
add_rubric / update_rubric channels.

Fold the numbers you measured (dynamic) or the artifact facts you read (static)
directly into each entry's `rationale`. A weakness claim with no grounding is not
allowed. The STATUS + finish-gate are DERIVED by the harness from your scores (any
important-and-failing rubric -> the agent is asked to fix it before finishing); you do
not set them.
\end{promptbox}

\begin{promptbox}{PRA: per-invocation blocks of the first message}
===== BEST-OF-N CONTEXT (cross-attempt) =====
This task runs the base agent as BEST-OF-<N>: up to <N> INDEPENDENT attempts, and ONLY THE BEST attempt's submission is scored (the final result is the MAX over attempts, not the last). This is attempt <a> (0-indexed; i.e. attempt <a+1> of <N>).
IMPLICATION for your scoring: because the beam keeps the MAX, a genuinely DIFFERENT approach that ends up second-best costs the run NOTHING, while an attempt that DUPLICATES an earlier one (same model family/features, near-identical predictions, re-tuning an already-saturated axis) WASTES one of the <N> shots. When you score the cross-attempt rubrics (no_churn_without_pivot, over_optimization, single_model_class_hedging, and the diversity/hyperparameter/feature-engineering rubrics), judge this attempt against what the PRIOR attempts already did -- reward diversification and flag convergence/redundant re-optimization.
Accumulated PRIOR-attempt transcripts (read-only; naming convention traj_a{attempt}_t{turn}.txt = that attempt's latest snapshot, turn = the base agent's turn count). Skim these to see what earlier attempts tried before judging cross-attempt redundancy:
  /reviewer/traj_a0_t<turn>.txt
  [...]

===== AGENT TRAJECTORY (PULL -- transcript SO FAR, on demand) =====
The agent's turn-annotated trajectory UP TO THE CURRENT TURN is a read-only file at:
  /reviewer/<trajectory file>
This is the trajectory SO FAR -- the agent has NOT finished and keeps working after your review; the file ending at the latest turn is the present moment, NOT a crash or a finish() call.
READ it when a rubric needs PROCESS evidence (how the agent worked over turns -- churn, repeated identical crashes, plateau-without-pivot, compute burned, suspicious-CV investigation), not just the end-state artifacts. Skim it; you do not have to read every turn. Artifact-grounded rubrics can usually be judged from the workspace alone.

WORKFLOW, strictly (cwd=<PRA workspace>): read /reviewer/submission_ledger.jsonl (your prior verdicts this rollout) -> list_rubrics -> read checks/README.md -> for each applicable spec, REUSE its indexed script or WRITE a small bespoke one (using the resolved columns/metric above) -> run it -> read its `RESULT:` line -> IMMEDIATELY call record_check_result with the numbers AND its two scores (importance_score + p_agent_success grounded in the result); do NOT run the next check first-> repeat. ORDER by cost/reliability: [...]; WARN + move on for any OOF check if OOF is truly absent. Before you verdict, APPEND your line to submission_ledger.jsonl and check for a best-submission regression. When you write a new script, append it to checks/README.md. Call submit_verdict by turn <k-1>. You have <k> turns; spend them writing+running+recording checks, not exploring.
\end{promptbox}

\begin{promptbox}{PRA: rubric and verdict tools (name, arguments, description)}
list_rubrics(status)
  List the checks in your persistent rubric store (persists across all attempts on this task). Start here to see your [...] checks + any you added previously.

add_rubric(name, description, origin)
  Add a reusable NL-spec check to the store so future submissions on this task are guarded. Describe the invariant (not task-specific columns). You MUST have already written + run + recorded a check for it this wake (run-once gate).
  description: The NL spec: WHAT this check verifies, WHY it matters, and what a failure looks like. The reviewer writes bespoke code for it per task -- do NOT hardcode column names or thresholds here; describe the invariant.

update_rubric(id, description)
  Refine an existing rubric's NL spec (e.g. sharpen the invariant or note a task-specific nuance). If the check script changed, update checks/README.md instead -- scripts live there, not here.

retire_rubric(id, reason)
  Retire a check that never discriminates (sets status:retired; never hard-deleted -- the op log is preserved for analysis).

record_check_result(id | name, importance_score, p_agent_success, evidence)
  Log what a check actually ran and produced BEFORE you cite it in your verdict (anti-hallucination: a verdict claim must trace to a recorded fire). Also satisfies the run-once gate for add_rubric on a new executable check.

submit_verdict(scores: [{rubric_id, importance_score, p_agent_success, rationale, suggested_action}])
  Emit your final verdict as PER-RUBRIC SCORES over the rubrics you assessed -- each {rubric_id, importance_score, p_agent_success, rationale, suggested_action}. The harness thresholds them into the injected block (importance >= min_importance AND p_agent_success < failure_threshold -> surfaced) and derives the finish-gate; you do NOT write the block or set a STATUS. Call once, then finish.
\end{promptbox}

\paragraph{Feedback to the agent.} The harness renders the rubrics that cross the failure thresholds (Table~\ref{tab:run-config}) into the block below, ordered by importance, and appends it to the agent's conversation. Feedback carried into the next attempt uses the same format, with a header that asks the agent to fix each issue in that attempt.

\begin{promptbox}{Feedback injected into the agent's conversation}
## Expert Feedback (Process Reward Model)

A process reward model reviewed your progress SO FAR on this in-progress attempt. The attempt is NOT finished -- this is mid-run guidance, not a verdict on a completed run. It found the following issues. In your VERY NEXT step you MUST explicitly acknowledge this feedback and, for EACH issue, either (a) take the suggested action, or (b) state in one line why it does not apply yet given where you are -- then continue. Do not silently ignore it. Start with the highest-importance issue.

### ISSUES TO FIX (address these in your remaining steps):
1. **<rubric display name>** [Importance: <i> | Success: <p>]:
   - Definition: <rubric definition>
   - Problem: <PRA rationale, citing the measured numbers>
   - Suggested action: <PRA suggested fix>
2. ...
\end{promptbox}

\paragraph{$-$PRA ablation.} The $-$PRA variant has no terminal. Its instructions replace the checking and procedure sections of the PRA instructions with the text below and replace the sentence describing the ability to run code with the second box. Seven further passages that refer to running code are rephrased to refer to reading files, and the per-invocation workflow asks it to open the relevant files instead of writing checks. The rubric store, scoring interface, and feedback format are unchanged. The $-$adaptive rubrics ablation uses the PRA instructions unchanged and withholds \texttt{add\_rubric}, \texttt{update\_rubric}, and \texttt{retire\_rubric}.

\begin{promptbox}{$-$PRA: replacement for the checking and procedure sections}
HOW YOU ASSESS -- rubrics are SPECS, you READ + REASON (no code)
- Your rubrics are natural-language SPECIFICATIONS of what to verify (see
  list_rubrics). You have NO terminal, so you do NOT write or run check scripts.
  For each spec, OPEN the relevant artifacts (the submission csv, cv_scores.json,
  oof_predictions.csv/.parquet, the modeling code) with the file-reading tool and
  reason about whether the rubric is satisfied from what you can SEE.
- checks/README.md persists across attempts as your memory of what earlier wakes
  concluded; read it first for prior reasoning, and record fresh findings with
  record_check_result so the next wake inherits them.

PROCEDURE (you have up to <k> turns -- spend them READING + REASONING, not hunting)
1. read /reviewer/submission_ledger.jsonl (your prior verdicts) + list_rubrics +
   checks/README.md -- know your specs and what earlier wakes found.
2. SOUNDNESS FIRST. For each soundness spec, open the artifacts it concerns and
   reason: does the submission match the OOF preds (submission-vs-OOF -- the money
   check)? Is coverage/schema right? Does the reported CV look reproducible from
   the OOF preds + target as written? record_check_result for each, with the
   concrete facts you read + your two scores (see SCORING).
3. OOF-dependent specs: look ONCE for oof*.csv/.parquet in ./, preds/. If absent,
   record those rubrics LOW importance (or p_agent_success=0.5) and MOVE ON.
4. MODELING DIAGNOSTICS. With turns left, read the modeling code + artifacts and
   reason about residual signal, prediction-vs-target dispersion, calibration,
   per-fold variance, single-feature leakage -- cite the facts you read.
5. If you discover a task-specific failure mode, add_rubric an NL spec for it (index
   it in checks/README.md). OPTIONAL -- skip if short on turns.
6. Call submit_verdict -- MANDATORY, by turn <k>-1 at the latest -- scoring EVERY
   applicable rubric, grounded ONLY in facts you read.
\end{promptbox}

\begin{promptbox}{$-$PRA: replacement for the sentence on running code}
You have NO terminal and CANNOT run code: you are a STATIC
analyzer. You READ its artifacts (out-of-fold predictions, residuals, features,
the submission file, train/test data) and its modeling code, and REASON to a
verdict from what you can SEE -- cite the concrete facts you read as evidence.
\end{promptbox}

\subsection{Rubric Discovery}
\label{sec:appendix-prompts-discovery}
Both stages use Claude Sonnet~4.6 on Amazon Bedrock, with one call per attempt in stage one (\S\ref{sec:discovery}) and a single call in stage two. Stage one covers the 187 harness-only DSPredict-Easy attempts with a nonempty transcript. Each attempt is placed in a band within its task: top if its leaderboard percentile is at or above the upper quartile of the task's valid attempts, bottom if it is at or below the lower quartile or invalid, and middle otherwise; with fewer than four valid attempts, the median separates top from bottom. The system prompt below is filled with one of two verdict instructions, one for top attempts and one for all others. The user message holds the attempt's turn-annotated transcript (at most 200{,}000 characters, with each tool output truncated at the OpenHands default limit), up to 40 workspace files with contents truncated to 1{,}800 characters (the first 12 rows for CSV files), and the agent's end-of-attempt summary. Each mined behavior records whether it is good practice or an issue, its evidence source (artifact or trajectory), and quoted evidence. Stage two pools the good behaviors of top attempts and the issues of bottom attempts across tasks (519 of the 906 behaviors: 238 good behaviors and 281 issues) and asks the judge, in one call, to cluster them into a MECE rubric set. Top attempts yield 238 good behaviors and 59 issues, and bottom attempts 53 good behaviors and 281 issues; stage two drops the 112 behaviors that run against their attempt's outcome, the 259 behaviors of middle attempts, and the 16 behaviors of valid attempts that Kaggle did not score.

\begin{promptbox}{Discovery stage one: behavior extraction}
You are an expert data-science process reward model (PRM) analyst.

Below is the record of an AGENT solving a Kaggle-style data-science competition,
in up to THREE parts:
  1. TRAJECTORY (the process): the turn-annotated transcript -- every action it
     took ([Turn N] ACTION ...) and the (truncated) tool output each produced
     ([Turn N] OUTPUT ...). The initial [Turn 0] USER message is the competition
     task. This is the process tau.
  2. WORKSPACE (the artifacts): a snapshot of the persistent files the agent left
     at the end of this attempt -- submission CSVs, code modules, notes, score
     registries, cached predictions. This is the workspace W: what the process
     actually produced, inspectable directly.
  3. ABSTRACT: the agent's own end-of-attempt summary (validation score, approach,
     preprocessing, model). A quick orientation, not ground truth.
Some parts may be absent (empty transcript, empty workspace). The TRAJECTORY plus
WORKSPACE together are what a live process supervisor would see.

We are BUILDING the rubric set for that PRM. Rather than score this trajectory,
your job is to mine it for the CONCRETE, GENERALIZABLE behaviors that drove its
outcome -- the raw material from which we will later distill process-reward
rubrics. You are told whether this attempt did WELL or POORLY relative to sibling
attempts on the SAME competition (same data, same time budget), so isolate
PROCESS, not task difficulty or luck.

<VERDICT: top-band or bottom-band instruction, below>

In THIS domain, the behaviors that actually move Kaggle outcomes cluster around
(use these as lenses, not a checklist -- surface whatever the transcript shows):
  - Validation discipline: is the CV scheme consistent and honest, or is a CV gain
    a leak/overfit artifact (calibration/meta-model fit on eval data, an outlier
    CV vs sibling models, tuning noise on a tiny dataset) that the agent TRUSTS?
  - Model selection for submission: does the FINAL submission come from the
    best-CV model, or did a later/worse experiment overwrite it (no revert)?
  - Submission integrity: does the final file cover ALL test rows with real,
    in-range predictions -- or is it partial / default-filled / corrupted /
    invalid, despite looking "valid"?
  - Exploration efficiency: productive iteration vs wasted motion -- repeated
    identical crashes, re-EDA loops, timed-out searches, plateaued churn.
  - Modeling strength: diverse model families, sensible ensembling/tuning, target
    handling appropriate to the metric.
  - Decisiveness & recovery: catching its own mistakes, reverting regressions,
    finishing with a verified submission vs meandering to the deadline.

Rules for every item you output (so items distill cleanly into rubrics):
- ATOMIC: one distinct behavior per item. Split anything joined by "and".
- SPECIFIC & OBSERVABLE: name the exact action -- model family, CV scheme, target
  transform, file op, submission check, bug, re-exploration loop. NOT vague
  hygiene ("did good EDA", "followed best practices").
- GENERALIZABLE: phrase so it could apply to OTHER competitions (e.g. "shipped a
  model with worse CV than one it had already trained" -- NOT "shipped v5 over v4").
- EVIDENCE-GROUNDED: cite concrete evidence -- a [Turn N] snippet from the
  trajectory AND/OR a workspace path (e.g. "preds/submission.csv", "cv_scores.json").
- GROUNDING (required): tag where this behavior is CHECKABLE:
    "artifact"  = verifiable from the WORKSPACE files alone, without watching the
                  process -- e.g. the submission covers all test rows / has in-range
                  values, a CV-score registry file exists, OOF predictions are
                  cached to disk, a code module leaks test data into a fit call.
    "trajectory" = verifiable ONLY from the PROCESS, because it is about HOW the
                  agent worked and leaves no decisive trace in the final files --
                  e.g. burned many turns without pivoting, kept re-running the same
                  failing cell, never reverted after a CV regression, abandoned a
                  better model mid-run. When in doubt, ask: "could I confirm this
                  from the end-state files only?" If yes -> artifact; if I'd need to
                  replay the turns -> trajectory.
- Prefer behaviors that plausibly MOVED the outcome; skip incidental ones.

Respond with ONLY valid JSON:
{
  "summary": "one sentence: the single biggest process driver of this attempt's outcome",
  "items": [
    {
      "kind": "good_behavior" | "issue",
      "behavior": "atomic, generalizable description of what the agent did or failed to do",
      "lens": "validation-discipline | model-selection | submission-integrity | exploration-efficiency | modeling-strength | decisiveness-recovery | other",
      "grounding": "artifact" | "trajectory",
      "impact": "how this plausibly helped or hurt the Kaggle outcome, and why a PRM should reward/penalize it",
      "evidence": ["[Turn N] <brief verbatim snippet>", "or workspace path: preds/submission.csv"]
    }
  ]
}
Output 2-6 items. If the transcript is empty or too short to judge, return
{"summary": "insufficient transcript", "items": []}.
\end{promptbox}

\begin{promptbox}{Discovery stage one: verdict instructions}
Top band:
VERDICT: this attempt did WELL -- it is in the TOP band of valid sibling attempts on this competition. Extract mainly GOOD_BEHAVIORS that plausibly drove the strong result (behaviors a PRM should REWARD and reinforce mid-run). You may also note an issue that kept it from being even better.

Bottom band:
VERDICT: this attempt did POORLY -- it is in the BOTTOM band of sibling attempts, OR it produced an INVALID/broken submission. Extract mainly ISSUES that plausibly caused the weak result (behaviors a PRM should PENALIZE and steer the agent away from mid-run). You may also note a good behavior it still exhibited.
\end{promptbox}

\begin{promptbox}{Discovery stage two: clustering behaviors into rubrics}
You are designing the COMPLETE rubric set, from scratch, for a
PROCESS REWARD MODEL (PRM) that scores data-science agents on Kaggle-style
competitions, turn by turn, from a turn-annotated transcript of their actions and
tool outputs together with the persistent workspace they build.

You are given a large pool of BEHAVIORS mined from many real agent trajectories.
Each behavior is either a GOOD_BEHAVIOR (observed in a top-outcome attempt) or an
ISSUE (observed in a bottom-outcome / invalid attempt). Treat these as positive and
negative EVIDENCE FOR THE SAME UNDERLYING RUBRICS: a good behavior is an instance of
an agent satisfying some good-practice criterion, and an issue is an instance of an
agent violating one. Your rubric set must cover both. Each behavior is atomic,
generalizable, and evidence-grounded, and carries a GROUNDING tag in parentheses
(lens|grounding):
  - "artifact"   = checkable from the agent's persistent WORKSPACE files alone
                   (submission integrity, a CV-score registry file, cached OOF
                   predictions, leaky code in a module).
  - "trajectory" = checkable only from the PROCESS -- HOW the agent worked, leaving
                   no decisive trace in the final files (churn without pivoting,
                   repeated identical crashes, a never-reverted CV regression).

Your job: discover, from the ground up, the SMALL MECE set of process-reward
rubrics implied by this pool. Assume NO pre-existing rubrics -- derive every rubric
bottom-up from the behaviors. Together the rubrics should account for the behaviors
that separate strong from weak outcomes in the pool.

Follow these rubric-writing principles:
- MECE: mutually exclusive (no two rubrics penalize the same underlying behavior)
  and collectively exhaustive (together they cover the outcome-separating behaviors
  in the pool). Fold a good behavior and its mirror-image issue into ONE rubric.
- ATOMIC: one distinct behavior per rubric. Break up anything with "and".
- SPECIFIC & OBSERVABLE: scoreable from a transcript and/or workspace -- name
  concrete signals (a leaked CV metric, a submission that overwrote a better model,
  a partial submission file, a repeated identical crash). Avoid vague ("good
  practices").
- COVERAGE-WEIGHTED: prioritize clusters that appear FREQUENTLY across the pool and
  clearly track outcome; drop rare/incidental one-offs.
- DOMAIN-GROUNDED: every rubric must be about data-science competition PROCESS.
- GROUNDING COVERAGE: surface BOTH artifact-grounded and trajectory-grounded
  rubrics -- do not collapse everything into one type. Assign each rubric a
  "grounding" equal to where it is decisively checkable. A rubric is "artifact"
  only if a supervisor could confirm it from the end-state workspace files alone;
  if confirming it requires inspecting HOW the agent worked over turns, it is
  "trajectory". Keep a rubric's grounding consistent with the behaviors in it.

For each rubric cluster, output a proposal in this schema (the dsgym RubricDef
shape). Phrase everything as a GOOD PRACTICE (yes = agent did the good thing):
{
  "rubric_id": "snake_case_id",
  "category": "A|B|C|D|E|F|G|H|I",   // A=validation/metric, B=exploration/model-diversity, C=preprocessing/target, E=efficiency/robustness/submission, etc -- pick the closest
  "grounding": "artifact" | "trajectory",   // where the rubric is decisively checkable (see GROUNDING COVERAGE)
  "display_name": "Neutral Topic Label (Title Case)",
  "description": "one failure-framed sentence: 'Agent fails to ...'",
  "display_description": "one neutral sentence: 'Whether the agent ...'",
  "natural_language_instruction": "4-label scoring instruction with EXACTLY these labels, each on its own line: 'Label \'agent_failure\' if: ...', 'Label \'agent_success\' if: ...', 'Label \'not_applicable\' if: ...', 'Label \'insufficient_evidence\' if: ...'. Ground each label in concrete, transcript- or workspace-observable signals.",
  "suggested_actions_hint": "a concrete imperative the PRM should suggest when the agent is FAILING this rubric (names the exact fix, not generic advice)",
  "supporting_behavior_count": <int: how many pool behaviors fall in this cluster>,
  "example_evidence": ["1-3 [Turn N] snippets or workspace paths drawn from the pool"]
}

Aim for a complete but compact set -- typically 12-20 rubrics total, no redundant or
overlapping rubrics, both groundings represented. Respond with ONLY valid JSON:
{"rubrics": [ ...proposals... ], "notes": "brief rationale for the set"}.
\end{promptbox}

\paragraph{Coverage of dropped behaviors.} To test whether the stage-two filter discards behaviors that would form new rubrics, we ask the same judge to assign each of the 387 dropped behaviors to the one discovered rubric it instantiates, or to none if no rubric clearly covers it, with the prompt below. Behaviors are sent in batches of about 30, each with its kind, lens, text, and up to two evidence quotes cut to 160 characters; the 17 rubrics are given by name and one-sentence description, and sampling uses temperature 0. As a baseline, we apply the same prompt to 60 randomly sampled kept behaviors. The judge assigns 378 of the 387 dropped behaviors (97.7\%) to a rubric, against 59 of the 60 kept behaviors (98.3\%). The nine unassigned behaviors are mostly isolated, low-severity cases, such as a wasted file operation; the largest group, two behaviors in which the final model was trained on a small subsample of the data, is close to Timeout-Aware Experiment Design.

\begin{promptbox}{Coverage check: assigning dropped behaviors to discovered rubrics}
You are auditing coverage of a rubric set for data-science agent trajectories.
Below are 17 process-reward rubrics, then a numbered list of behaviors observed in agent
attempts. Each behavior is tagged GOOD (a positive behavior) or ISSUE (a problem).

For EACH behavior, decide:
- the single best-matching rubric_id, if the behavior is clearly an instance (positive or
  negative) of that rubric, i.e. a grader applying that rubric would reward or penalize it; or
- "none", if no rubric clearly covers it.
Do not stretch: if the match is only loose or tangential, answer "none".

===== RUBRICS =====
<one line per rubric: rubric_id (display name): one-sentence description>

===== BEHAVIORS =====
<one line per behavior: idx. [GOOD|ISSUE] (lens) behavior text [evidence: <=2 quotes>]>

Respond with ONLY a JSON object, no prose, of the form:
{"results": [{"idx": 1, "rubric_id": "<rubric_id or none>", "reason": "<one line>"}, ...]}
with exactly one entry per behavior idx 1..<N>.
\end{promptbox}

\section{DataPRM Baseline}
\label{sec:appendix-dataprm}
DataPRM \cite{qiu2026rewarding} is a generative process reward model for agentic data analysis. For each step of a trajectory it reasons about the step, writes and executes Python code against the task files to check it, and outputs a ternary score (1 for a correct step, 0.5 for minor issues, 0 for a fatal error) with a one-sentence summary. Its authors use it for best-of-$N$ selection and as a reward in reinforcement learning. We use the released checkpoint trained on ScienceAgentBench (\texttt{zjunlp/DataPRM-ScienceAgentBench}, 4B parameters, fine-tuned from Qwen3-4B-Instruct), the closer of the two released checkpoints to model-building tasks; the other is trained on DABStep question answering. We serve it with vLLM on one A100 GPU.

\paragraph{Integration.} DataPRM replaces the PRA and every other setting is that of \methodname{}: the base agent, harness, five-attempt protocol, cadence ($c{=}10$), review on finish, finish refinement, and feedback channel (Table~\ref{tab:run-config}). The condition therefore isolates the supervisor and compares a generic step-level reward model with rubric-based supervision under identical conditions. At each invocation DataPRM verifies, in order, each agent turn taken since the previous invocation. One turn (the agent's action and its output) is one step, which matches DataPRM's native split of a trajectory into paragraphs, and each step is judged with all earlier steps and DataPRM's own earlier verifications as history. We keep the released judge prompt verbatim and the released inference settings: up to 10 reasoning and execution rounds per step, 8{,}192 generated tokens per round, execution outputs truncated to 8{,}192 characters, a 30\,s limit per code execution, cumulative re-execution of earlier successful code within a step, and sampling at temperature 0.7, top-$p$ 0.9, and top-$k$ 20. An invocation returns FAIL if any verified step scores 0, WARN if any scores 0.5, and PASS otherwise. The summaries of steps scoring 0 or 0.5 are injected into the agent's conversation, and a FAIL at finish triggers the same refinement round as a PRA FAIL.

\paragraph{Adaptations to online supervision.} DataPRM is released for offline scoring of complete trajectories, and running it during an attempt requires five changes. (1) At each invocation it scores the steps taken since the previous invocation, because online feedback must arrive before the attempt ends. (2) Offline, DataPRM re-executes all of the agent's earlier code to rebuild the workspace before verifying a step. Our agents train models for up to hours per attempt, which makes re-execution infeasible, so DataPRM's code instead runs in a writable directory linked to the agent's live workspace and the task data, with the agent's workspace paths mapped onto it; the code can therefore read every file the agent has produced. (3) The ScienceAgentBench checkpoint was trained with a \texttt{query\_image} tool. We keep the tool's description in the prompt to match the training distribution, but no vision model is served, so a call returns a message that the tool is unavailable; DSPredict-Easy is tabular, so the tool is rarely relevant. (4) Each earlier step in the history is clipped to 2{,}000 characters and the current step to 12{,}000, and the oldest history steps are dropped, so that each prompt fits the 24K-token context used in training. (5) Each step shows the agent's reasoning, its action in full (the command, the text of a recorded thought, or the path and contents of a file it writes), and the action's output.

\paragraph{Behavior.} Over its 1{,}590 invocations on DSPredict-Easy, DataPRM verifies a median of 10 steps per invocation in a median of 126\,s. It returns FAIL on 43\% of invocations, WARN on 32\%, and PASS on 24\%, and it scores 10\% of the verified steps 0. Most zero scores flag an error in the step's code or analysis, such as a script that fails to run, a reference to a file that does not exist, or a mismatch between the stated plan and the code, and others penalize an approach that the agent already tried in an earlier attempt.

\section{Rubric and Principle Reward Models}
\label{sec:appendix-rubric-rms}
Recent work has explored rubrics, checklists, and natural-language principles as more flexible reward specifications than scalar preference labels. Rubric-supervised critics learn dense behavioral features from sparse real-world outcomes, showing that structured rubric labels can improve reward modeling and test-time selection \cite{wang2026rubric}. Rubrics as Rewards uses structured rubric criteria as reward signals for reinforcement learning beyond domains with easily verifiable answers \cite{gunjal2025rubrics}. Checklist-based approaches similarly decompose instructions into explicit criteria that can be judged or verified to produce reward signals \cite{viswanathan2026checklists}. Most closely, RewardAnything introduces principle-following reward models and RABench to test whether reward models can follow dynamically provided natural-language principles and generalize to novel principles without retraining \cite{yu2025rewardanything}. A related reframing appears in MetaLint, which casts code linting as instruction following over user-specified best practices instead of classification over a fixed set of issue types \cite{naik2025metalint}.
Like these principle-following models, the PRA receives its rubrics as input instead of learning them in training. Unlike them, it judges long-horizon agent trajectories and the artifacts they produce instead of a single final response, and it can revise the rubrics it receives.

\section{Selection-Policy Analysis}
\label{sec:appendix-selection-policy}
Reporting the best attempt assumes that it can be identified after the fact, which is not possible in deployment. Tables~\ref{tab:selection-policy} and \ref{tab:selection-policy-hard} report the leaderboard percentile under four ways of choosing one attempt per task: the first attempt, the last attempt, the mean over scored attempts, which is the expected result of choosing an attempt at random, and the best attempt. The last attempt is the natural deployment choice for a sequential run in a shared workspace, since each attempt can build on the ones before it.

On DSPredict-Easy the last attempt improves on the first by 4.5 points without supervision and by 6.5 points with \methodname{}, whose last attempt (62.9) comes within 6.1 points of its best-attempt score (69.0). On DSPredict-Hard the last attempt improves on the first under every condition. \methodname{} has the highest mean and best-attempt percentiles (32.1 and 36.1), and its last attempt (30.2) is within 0.2 points of the highest last-attempt score ($-$PRA, 30.4).

\begin{table*}[t]
\centering
\small
\begin{tabular}{@{}l rrrr@{}}
\toprule
 & \multicolumn{4}{@{}l@{}}{\textbf{Leaderboard percentile by selection policy}} \\
\cmidrule{2-5}
\textbf{Method} & \textbf{First} & \textbf{Last} & \textbf{Mean} & \textbf{Oracle} \\
\midrule
Qwen3-Coder-Next + Ours           & 48.4 & 52.9 & 51.3 & 60.8 \\
\midrule
\multicolumn{5}{@{}l@{}}{\emph{Process supervision on top of Our Harness}}\\
\quad + PRA + adaptive rubrics (\methodname{}) & \textbf{56.4} & \textbf{62.9} & \textbf{61.0} & \textbf{69.0} \\
\quad \hspace{1.2em}$-$ adaptive rubrics     & 55.1 & 59.8 & 58.8 & 67.3 \\
\quad \hspace{1.2em}$-$ PRA & 52.4 & 61.1 & 57.8 & 63.0 \\
\bottomrule
\end{tabular}
\caption{Leaderboard percentile on \textbf{DSPredict-Easy} under four attempt-selection policies over the 5-attempt beam, task-first averaged over 38 tasks. \textbf{First} takes the first attempt, with an invalid first attempt contributing 0, \textbf{Last} takes the last attempt with a scored submission, \textbf{Mean} averages the scored attempts (the expected result of a random choice), and \textbf{Oracle} takes the best attempt.}
\label{tab:selection-policy}
\end{table*}

\begin{table*}[t]
\centering
\small
\begin{tabular}{@{}l rrrr@{}}
\toprule
 & \multicolumn{4}{@{}c@{}}{\textbf{Leaderboard percentile by selection policy}} \\
\cmidrule{2-5}
\textbf{Method} & \textbf{First} & \textbf{Last} & \textbf{Mean} & \textbf{Oracle} \\
\midrule
Qwen3-Coder-Next + Ours           & 17.0 & 22.2 & 26.2 & 25.5 \\
\midrule
\multicolumn{5}{@{}l}{\emph{Process supervision on top of Our Harness}}\\
\quad + PRA + adaptive rubrics (\methodname{}) & 23.3 & 30.2 & \textbf{32.1} & \textbf{36.1} \\
\quad \hspace{1.2em}$-$ adaptive rubrics     & 21.1 & 29.4 & 30.2 & 31.8 \\
\quad \hspace{1.2em}$-$ PRA & \textbf{25.1} & \textbf{30.4} & 30.9 & 31.8 \\
\bottomrule
\end{tabular}
\caption{As Table~\ref{tab:selection-policy}, on \textbf{DSPredict-Hard} (54 tasks). \textbf{First}, \textbf{Last}, and \textbf{Oracle} use the fixed denominator, with unscored tasks contributing 0. \textbf{Mean} is the per-attempt percentile of Table~\ref{tab:harness-hard}, averaged over tasks with at least one scored attempt%
, so it can exceed \textbf{Last}.}
\label{tab:selection-policy-hard}
\end{table*}

\section{Paired Significance of Process Supervision}
\label{sec:appendix-significance}
We test whether each supervised condition improves over the harness alone (Tables~\ref{tab:significance} and~\ref{tab:significance-hard}), and we report the same tests for \methodname{} against each of its two ablations in the text. The unit of analysis is the task, and paired differences $d_i = \text{condition}_i - \text{reference}_i$ cancel differences in task difficulty.
The primary test is a one-sided paired permutation test on the mean difference $\bar{d}$. Under the null hypothesis that the two conditions are exchangeable within each task, each $d_i$ is equally likely to take either sign, so we compare the observed $\bar{d}$ with its distribution under $200{,}000$ random sign flips of the $d_i$. Unlike the sign test, this test uses the magnitude of each difference, and unlike the Wilcoxon signed-rank test, it tests the mean difference that we report without assuming that the differences are symmetric.
We also report the one-sided sign test, the one-sided Wilcoxon signed-rank test, and a nested bootstrap 95\% confidence interval, which resamples tasks with replacement and then attempts within each task, so the interval includes variation across attempts. Both metrics of Tables~\ref{tab:harness} and \ref{tab:harness-hard} are tested: the best attempt and the mean over a task's valid attempts. Percentiles use the private leaderboard where one exists and the public leaderboard otherwise.

On DSPredict-Easy ($n=38$, Table~\ref{tab:significance}), \methodname{} has the largest gains over the harness alone on both metrics (best attempt $+8.2$, $p<0.001$, CI $[+4.4,+14.4]$; per attempt $+9.7$, $p<0.001$). The gains of $-$adaptive rubrics are significant on both metrics (best attempt $+6.5$, $p=0.017$; per attempt $+7.5$, $p=0.008$). The best-attempt gain of $-$PRA is not significant ($+2.2$, $p=0.273$), but its per-attempt gain is ($+6.6$, $p=0.009$). Compared directly on the best attempt, \methodname{} outperforms $-$PRA ($+6.0$; 23 wins, 15 losses; $p=0.012$), while its gain over $-$adaptive rubrics is not significant ($+1.7$; 22 wins, 16 losses; $p=0.147$).

On DSPredict-Hard ($n=54$, Table~\ref{tab:significance-hard}), every supervised condition improves significantly over the harness alone on both metrics, and \methodname{} has the largest gains (best attempt $+10.6$, $p<0.001$, CI $[+5.3,+16.2]$; per attempt $+7.8$, $p=0.001$, CI $[+2.8,+13.0]$). The gains of $-$PRA are significant on both metrics (best attempt $+6.3$, $p=0.009$; per attempt $+6.7$, $p<0.001$), as are those of $-$adaptive rubrics (best attempt $+6.3$, $p=0.007$; per attempt $+5.5$, $p=0.005$). Compared directly on the best attempt, \methodname{} outperforms $-$adaptive rubrics ($+4.3$; 32 wins, 13 losses; $p=0.024$), so rubric adaptation adds to code execution on the held-out benchmark. Its gain over $-$PRA ($+4.3$; 29 wins, 18 losses) is not significant ($p=0.077$). Per attempt, neither direct difference is significant ($+2.3$, $p=0.138$; $+1.1$, $p=0.358$).

\begin{table*}[t]
\centering
\small
\begin{tabular}{@{}l r c c c c c@{}}
\toprule
\textbf{Condition} (\emph{vs.}\ harness only) & \textbf{$\Delta$ mean} & \textbf{Wins/Losses} & \textbf{Perm.\ $p$} & \textbf{Sign $p$} & \textbf{Wilcoxon $p$} & \textbf{95\% CI} \\
\midrule
\multicolumn{7}{@{}l}{\emph{Best attempt (maximum percentile over a task's valid attempts)}}\\
\quad \methodname{} & $\mathbf{+8.2}$ & $\mathbf{25/11}$ & $\mathbf{<0.001}$ & $\mathbf{0.014}$ & $\mathbf{0.0005}$ & $\mathbf{[+4.4, +14.4]}$ \\
\quad $-$adaptive rubrics & $\mathbf{+6.5}$ & $\mathbf{23/15}$ & $\mathbf{0.017}$ & $\mathbf{0.128}$ & $\mathbf{0.030}$ & $\mathbf{[+0.3, +13.8]}$ \\
\quad $-$PRA & $+2.2$ & $18/19$ & $0.273$ & $0.629$ & $0.275$ & $[-3.0, +10.4]$ \\
\midrule
\multicolumn{7}{@{}l}{\emph{Per attempt (mean percentile over a task's valid attempts)}}\\
\quad \methodname{} & $\mathbf{+9.7}$ & $\mathbf{31/7}$ & $\mathbf{<0.001}$ & $\mathbf{<0.001}$ & $\mathbf{<0.001}$ & $\mathbf{[+5.1, +14.5]}$ \\
\quad $-$adaptive rubrics & $\mathbf{+7.5}$ & $\mathbf{25/13}$ & $\mathbf{0.008}$ & $\mathbf{0.037}$ & $\mathbf{0.007}$ & $\mathbf{[+1.4, +13.6]}$ \\
\quad $-$PRA & $\mathbf{+6.6}$ & $\mathbf{23/15}$ & $\mathbf{0.009}$ & $\mathbf{0.128}$ & $\mathbf{0.028}$ & $\mathbf{[+1.2, +12.1]}$ \\
\bottomrule
\end{tabular}
\caption{Paired significance of each supervised condition against the harness alone on \textbf{DSPredict-Easy} (task-paired, $n=38$). The one-sided paired permutation test ($200{,}000$ sign flips) is primary, with the one-sided sign test, the one-sided Wilcoxon signed-rank test, and a nested bootstrap 95\% CI ($20{,}000$ resamples) alongside. Ties are excluded from the win and loss counts. Bold marks comparisons significant at $p<0.05$ by the permutation test.}
\label{tab:significance}
\end{table*}

\begin{table*}[t]
\centering
\small
\begin{tabular}{@{}l r c c c c c@{}}
\toprule
\textbf{Condition} (\emph{vs.}\ harness only) & \textbf{$\Delta$ mean} & \textbf{Wins/Losses} & \textbf{Perm.\ $p$} & \textbf{Sign $p$} & \textbf{Wilcoxon $p$} & \textbf{95\% CI} \\
\midrule
\multicolumn{7}{@{}l}{\emph{Best attempt (maximum percentile over a task's valid attempts)}}\\
\quad \methodname{} & $\mathbf{+10.6}$ & $\mathbf{33/15}$ & $\mathbf{<0.001}$ & $\mathbf{0.007}$ & $\mathbf{<0.001}$ & $\mathbf{[+5.3, +16.2]}$ \\
\quad $-$adaptive rubrics & $\mathbf{+6.3}$ & $\mathbf{25/18}$ & $\mathbf{0.007}$ & $\mathbf{0.180}$ & $\mathbf{0.025}$ & $\mathbf{[+1.7, +11.3]}$ \\
\quad $-$PRA & $\mathbf{+6.3}$ & $\mathbf{30/15}$ & $\mathbf{0.009}$ & $\mathbf{0.018}$ & $\mathbf{0.017}$ & $\mathbf{[+1.5, +12.2]}$ \\
\midrule
\multicolumn{7}{@{}l}{\emph{Per attempt (mean percentile over a task's valid attempts)}}\\
\quad \methodname{} & $\mathbf{+7.8}$ & $\mathbf{36/12}$ & $\mathbf{0.001}$ & $\mathbf{<0.001}$ & $\mathbf{<0.001}$ & $\mathbf{[+2.8, +13.0]}$ \\
\quad $-$adaptive rubrics & $\mathbf{+5.5}$ & $\mathbf{30/15}$ & $\mathbf{0.005}$ & $\mathbf{0.018}$ & $\mathbf{0.018}$ & $\mathbf{[+1.5, +9.9]}$ \\
\quad $-$PRA & $\mathbf{+6.7}$ & $\mathbf{32/16}$ & $\mathbf{<0.001}$ & $\mathbf{0.015}$ & $\mathbf{0.002}$ & $\mathbf{[+2.3, +11.9]}$ \\
\bottomrule
\end{tabular}
\caption{As Table~\ref{tab:significance}, on \textbf{DSPredict-Hard} (task-paired, $n=54$). Tasks without a scored attempt enter as 0 so that every task is paired, which makes the per-attempt deltas slightly larger than the differences in Table~\ref{tab:harness-hard}, where per-attempt means exclude such tasks.}
\label{tab:significance-hard}
\end{table*}

\begin{figure}[!tbh]
\centering
\includegraphics[width=0.5\textwidth]{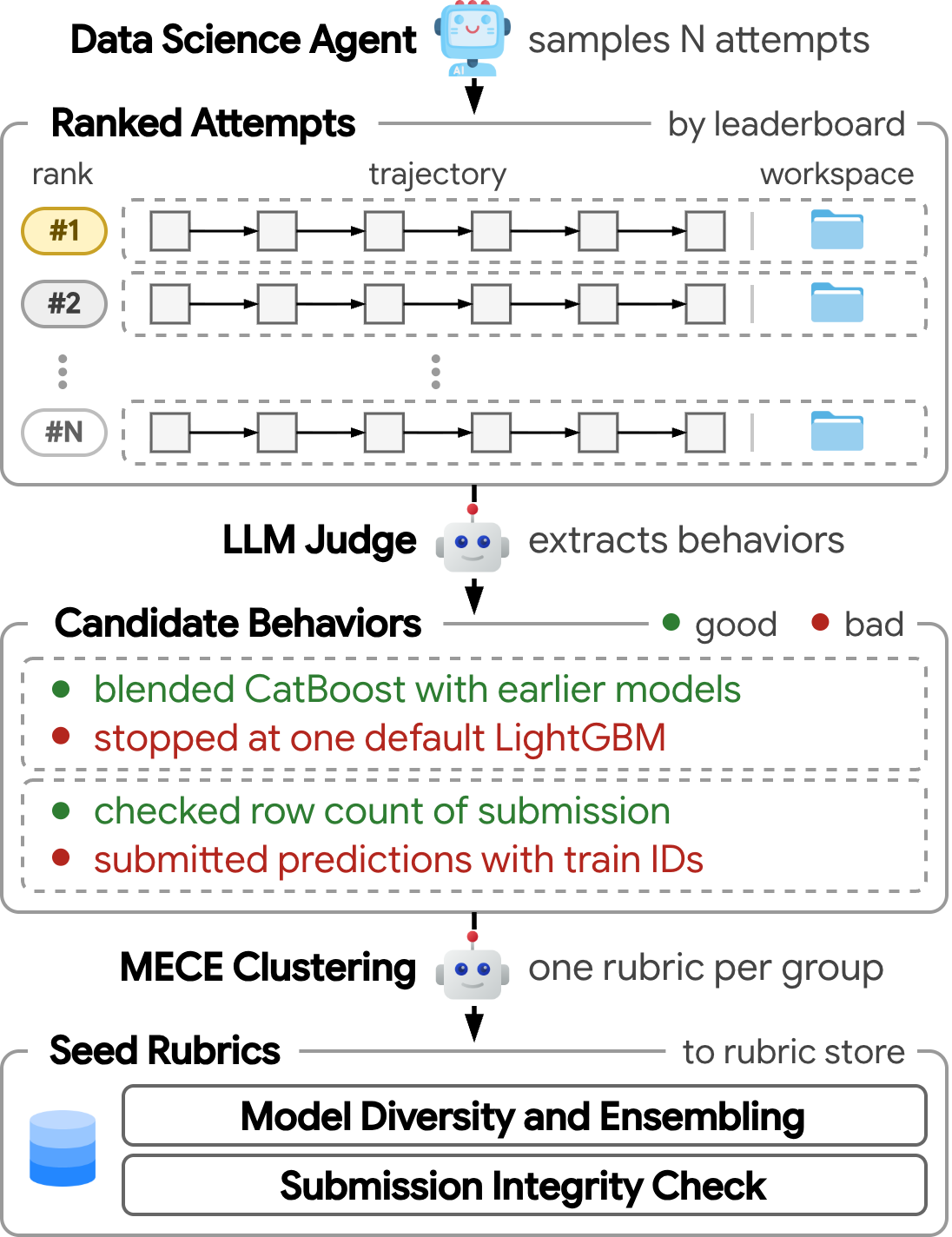}
\caption{\textbf{Offline rubric discovery.} The base agent's attempts on each task are ranked by leaderboard percentile. An LLM judge (Claude Sonnet~4.6) reads each attempt's trajectory and workspace, is told whether the attempt ranks in the top or bottom quarter of its task, and extracts good and bad behaviors with quoted evidence. A second call clusters the good behaviors of top attempts and the issues of bottom attempts across tasks into a MECE set of seed rubrics, with good and bad behaviors as evidence for the same rubric (prompts and settings in Appendix~\ref{sec:appendix-prompts-discovery}).}
\label{fig:rubric_discovery}
\end{figure}

\section{Seed Rubrics}
\label{sec:appendix-rubrics}
Table~\ref{tab:rubric-meta} lists the 19 seed rubrics that initialize every rubric store (\S\ref{sec:discovery}), and Tables~\ref{tab:rubric-nl-A}--\ref{tab:rubric-nl-E2} give their full natural-language definitions. The PRA receives each rubric as its one-sentence definition followed by its four labels (satisfied, violated, not applicable, insufficient evidence) and a note stating the rubric's evidence source, which is phrased as a recommendation so that the PRA may use either kind of evidence. Seventeen rubrics come from offline discovery on harness-only DSPredict-Easy runs. Four of them are restated to reason over all earlier attempts on the task, and two further rubrics are hand-curated. Stage one mines each attempt independently, so it cannot surface failures that appear only across attempts, such as later attempts converging on the same solution; the two hand-curated rubrics target these failures. They account for 11--17\% of rubric failures across the supervised runs, and they are the only failing rubrics in 1--3\% of reviews that report a failure. Six rubrics refer to the transcripts of earlier attempts, which the PRA reads on demand (\S\ref{sec:pra}). The rubric names and definitions are reproduced verbatim; ``best-of-N'' in them refers to the $N$ sequential attempts on a task, of which the best is scored.

\begin{table*}[t]
\centering
\small
\resizebox{\textwidth}{!}{%
\begin{tabular}{@{}r l l l l r c@{}}
\toprule
\textbf{\#} & \textbf{Rubric} & \textbf{Category} & \textbf{Evidence} & \textbf{Source} & \textbf{Support} & \textbf{Expert match} \\
\midrule
1 & Persistent CV Score Registry & Validation and metric & Artifact & Discovered & 52 & Yes \\
2 & Best-CV Model Submission Selection & Validation and metric & Artifact & Discovered & 44 & Yes \\
3 & Valid Cross-Validation Methodology & Validation and metric & Artifact & Discovered & 41 & Yes \\
4 & Suspicious CV Score Investigation & Validation and metric & Trajectory & Discovered & 18 & No \\
5 & Metric Calculation Bug Detection & Validation and metric & Trajectory & Discovered & 8 & No \\
\addlinespace
6 & Model Diversity and Ensembling & Exploration and model diversity & Artifact & Discovered, restated & 28 & Yes \\
7 & Hyperparameter Tuning Beyond Defaults & Exploration and model diversity & Artifact & Discovered, restated & 20 & No \\
8 & Hedging Across Best-of-N With Diverse Model Classes & Exploration and model diversity & Trajectory & Hand-curated & -- & -- \\
\addlinespace
9 & Training Objective--Metric Alignment & Preprocessing, target, and features & Artifact & Discovered & 14 & Yes \\
10 & Domain-Appropriate Preprocessing & Preprocessing, target, and features & Artifact & Discovered & 12 & Partial \\
11 & Feature Engineering Investment & Preprocessing, target, and features & Artifact & Discovered, restated & 15 & No \\
\addlinespace
12 & OOF and Test Prediction Caching & Efficiency, robustness, and submission & Artifact & Discovered & 38 & Yes \\
13 & Submission File Integrity Verification & Efficiency, robustness, and submission & Artifact & Discovered & 28 & Yes \\
14 & CV Regression Detection and Revert & Efficiency, robustness, and submission & Trajectory & Discovered & 22 & Yes \\
15 & Avoiding Repeated Identical Crashes & Efficiency, robustness, and submission & Trajectory & Discovered & 24 & Yes \\
16 & Avoiding Churn Without Pivoting & Efficiency, robustness, and submission & Trajectory & Discovered, restated & 22 & Yes \\
17 & Valid Submission Before Finishing & Efficiency, robustness, and submission & Artifact & Discovered & 18 & No \\
18 & Timeout-Aware Experiment Design & Efficiency, robustness, and submission & Trajectory & Discovered & 19 & Yes \\
19 & Over-Optimization Without Significant Gains & Efficiency, robustness, and submission & Trajectory & Hand-curated & -- & -- \\
\bottomrule
\end{tabular}}
\caption{Seed rubric metadata. \emph{Evidence}: whether the rubric is decisively checkable from the workspace artifacts or requires the trajectory (11 artifact, 8 trajectory). \emph{Source}: \emph{Discovered} rubrics come from two-stage discovery on harness-only DSPredict-Easy runs; \emph{Discovered, restated} rubrics were restated to judge the current attempt against all earlier attempts on the task; \emph{Hand-curated} rubrics target cross-attempt failures. \emph{Support}: number of the 519 behaviors from the top and bottom of each task's score range (238 good behaviors, 281 issues) that stage two clustered into the rubric. \emph{Expert match}: whether the rubric recovers a criterion of a hand-written expert rubric set (Yes), overlaps with one loosely (Partial), or has no expert counterpart (No).}
\label{tab:rubric-meta}
\end{table*}

\begin{table*}[p]
\centering
\scriptsize
\resizebox{\textwidth}{!}{%
\begin{tabular}{@{}>{\raggedright\arraybackslash}p{3.3cm} >{\raggedright\arraybackslash}p{4.6cm} >{\raggedright\arraybackslash}p{4.6cm} >{\raggedright\arraybackslash}p{3.3cm}@{}}
\toprule
\textbf{Rubric and definition} & \textbf{Satisfied if} & \textbf{Violated if} & \textbf{Not applicable / insufficient evidence if} \\
\midrule
\textbf{Persistent CV Score Registry} (\texttt{cv\_\allowbreak{}score\_\allowbreak{}registry}). Whether the agent maintains a persistent cv\_\allowbreak{}scores.json (or equivalent) updated after every experiment to enable objective cross-model comparison. & A machine-readable file (e.g., cv\_\allowbreak{}scores.json) exists in the workspace containing CV scores for every trained model/ensemble, updated consistently throughout the run, and the agent uses it to identify the best model at submission time. & No such registry exists, or it is missing entries for major experiments, or it is overwritten per-run rather than accumulated, preventing reliable comparison across all experiments. & \emph{N/A:} Only a single model was ever trained and no comparison was needed. \emph{Insufficient:} The workspace files are not visible and the trajectory does not mention CV score tracking. \\
\midrule
\textbf{Best-CV Model Submission Selection} (\texttt{best\_\allowbreak{}model\_\allowbreak{}submission\_\allowbreak{}selection}). Whether the agent explicitly selects and submits the model with the best CV score rather than whatever was most recently trained or written to disk. & The final /submission/submission.csv is verifiably sourced from the model/experiment with the highest CV score in the registry, and the agent explicitly checked the registry before writing the final submission. & The submission corresponds to a later, lower-CV experiment that overwrote the best model's output, or the agent selected the most-recent model without checking CV scores. & \emph{N/A:} Only one model was trained so no selection was needed. \emph{Insufficient:} It cannot be determined which model's predictions are in the submission file. \\
\midrule
\textbf{Valid Cross-Validation Methodology} (\texttt{valid\_\allowbreak{}cv\_\allowbreak{}methodology}). Whether the agent uses a proper out-of-fold or held-out cross-validation scheme that produces honest, unbiased CV score estimates. & All reported CV scores are computed using held-out fold predictions (OOF) that were not used during model fitting, with an appropriate split strategy for the task (stratified for classification, time-based for time series). & Any CV score in cv\_\allowbreak{}scores.json is computed by evaluating a model on the same data it was trained on (in-sample), or ensemble weights are searched using the same OOF data without a secondary holdout, or the CV scheme is wrong for the task type (e.g., random KFold for time series). & \emph{N/A:} No model training or CV was performed. \emph{Insufficient:} The training code is not visible and CV methodology cannot be inferred from outputs. \\
\midrule
\textbf{Suspicious CV Score Investigation} (\texttt{suspicious\_\allowbreak{}cv\_\allowbreak{}score\_\allowbreak{}investigation}). Whether the agent investigates and correctly handles CV scores that are suspiciously better than all other models (a signal of data leakage or in-sample evaluation). & When a model's CV score is substantially better than all others ($>$ 20\% relative improvement with no architectural explanation), the agent diagnoses whether it is legitimate or an artifact of leakage/in-sample evaluation, and either discards the model or confirms correctness with a separate holdout before using it for submission. & An obviously inflated CV score (e.g., perfect 1.0 AUC, RMSE 10x better than all others, correlation 0.96 with training labels) is accepted as valid, recorded in cv\_\allowbreak{}scores.json, and used for submission or downstream ensembling without investigation. & \emph{N/A:} No suspiciously anomalous CV scores appeared during the run. \emph{Insufficient:} CV score values are not visible in the trajectory or workspace. \\
\midrule
\textbf{Metric Calculation Bug Detection} (\texttt{metric\_\allowbreak{}bug\_\allowbreak{}detection}). Whether the agent detects and corrects bugs in metric calculation code before reporting or acting on CV scores. & When a CV score appears implausibly low or inconsistent with prior results (e.g., RMSLE of 0.01 when all prior models score \textasciitilde{}0.13), the agent identifies and fixes the root cause in the metric code before using the score for model selection. & An obviously buggy metric (double log-transform, wrong target, train-set RMSE) produces a clearly impossible CV score that the agent reports as valid and uses for multiple turns before catching it (if ever). & \emph{N/A:} No metric bugs occurred and all CV scores were plausible. \emph{Insufficient:} The metric calculation code is not visible. \\
\bottomrule
\end{tabular}}
\caption{Seed rubric definitions: validation and metric. Text is verbatim as given to the PRA; \emph{Context} is a preamble that precedes the labels.}
\label{tab:rubric-nl-A}
\end{table*}

\begin{table*}[p]
\centering
\scriptsize
\resizebox{\textwidth}{!}{%
\begin{tabular}{@{}>{\raggedright\arraybackslash}p{3.3cm} >{\raggedright\arraybackslash}p{4.6cm} >{\raggedright\arraybackslash}p{4.6cm} >{\raggedright\arraybackslash}p{3.3cm}@{}}
\toprule
\textbf{Rubric and definition} & \textbf{Satisfied if} & \textbf{Violated if} & \textbf{Not applicable / insufficient evidence if} \\
\midrule
\textbf{Model Diversity and Ensembling} (\texttt{model\_\allowbreak{}diversity\_\allowbreak{}and\_\allowbreak{}ensembling}). Whether the agent trains at least two distinct model families and combines them via ensemble or stacking, and whether model-family coverage across the best-of-N attempts is diverse rather than duplicated. \emph{Context:} This is a best-of-N run (N=n\_\allowbreak{}attempts); the submitted result is the MAX over attempts, so what matters is the diversity of the WHOLE beam, not just this attempt. Read the accumulated prior-attempt transcripts (traj\_\allowbreak{}a\{attempt\}\_\allowbreak{}t\{turn\}.txt in /reviewer/) and cv\_\allowbreak{}scores.json to see which model families the earlier attempts already covered. & The workspace contains trained models from at least two distinct families (e.g., LightGBM + XGBoost, or GBDT + neural net), their OOF predictions are combined via weighted averaging or stacking with the ensemble CV compared against individual models, AND -- across attempts -- the best-of-N beam is not collapsing onto one family (i.e., this attempt either adds a family/architecture the beam has not tried or is early enough that convergence is not yet a concern). & Only one model family was trained, or models from multiple families were trained but never combined, or an ensemble combined weak+strong models in a way that degraded performance (e.g., blending with a model 2.5pp weaker), OR this attempt is re-training essentially the same model families/ensemble the prior attempts already built (duplicating the beam) when a distinct family could have been tried. & \emph{N/A:} The competition or task type makes single-model submission standard (e.g., data-centric competition where model is fixed), or fewer than two attempts have run so cross-attempt diversity cannot yet be assessed. \emph{Insufficient:} Training scripts and the prior-attempt transcripts are not visible. \\
\midrule
\textbf{Hyperparameter Tuning Beyond Defaults} (\texttt{hyperparameter\_\allowbreak{}tuning}). Whether the agent performs at least one principled hyperparameter search beyond default settings for the primary model family -- while recognizing when tuning has already been exhausted in earlier best-of-N attempts. \emph{Context:} This is a best-of-N run (N=n\_\allowbreak{}attempts). Before scoring, consult the accumulated prior-attempt transcripts (traj\_\allowbreak{}a\{attempt\}\_\allowbreak{}t\{turn\}.txt in /reviewer/) and cv\_\allowbreak{}scores.json to see whether hyperparameter search was ALREADY done in an earlier attempt and what it yielded. & EITHER the workspace contains at least one training script or cv\_\allowbreak{}scores.json entry showing a model trained with non-default hyperparameters (e.g., num\_\allowbreak{}leaves != 31, learning\_\allowbreak{}rate != 0.05, or a grid/random/Optuna search) with the tuned CV compared against baseline, OR the agent correctly recognizes that prior attempts already tuned this family to saturation (no meaningful CV gain) and therefore spends its budget elsewhere (new family, features, diversity) rather than re-tuning. & The primary model uses default/near-default hyperparameters (num\_\allowbreak{}leaves=31, max\_\allowbreak{}depth=6, lr=0.05) with no documented search AND no earlier attempt tuned it either, OR a 'tuned' variant produced CV identical to baseline. Do NOT reward re-running a hyperparameter search that a prior attempt already showed to be saturated -- that is over-optimization, not diligence. & \emph{N/A:} The task is data-centric (fixed model) or the budget was trivially small. \emph{Insufficient:} Hyperparameter values are not visible in training scripts, the registry, or the prior-attempt transcripts. \\
\midrule
\textbf{Hedging Across Best-of-N With Diverse Model Classes} (\texttt{single\_\allowbreak{}model\_\allowbreak{}class\_\allowbreak{}hedging}). Whether the best-of-N attempts hedge across DIFFERENT modeling classes rather than converging on the same solution near finish -- using the max-over-N structure to try a distinct approach (even a second-best one) instead of duplicating the leading attempt. \emph{Context:} This is a best-of-N run (N=n\_\allowbreak{}attempts) and the SUBMITTED result is the MAX over attempts, so a genuinely different modeling class that ends up second-best costs nothing, while a duplicate of an existing attempt wastes one of the N shots. Judge this ACROSS attempts using the accumulated prior-attempt transcripts (traj\_\allowbreak{}a\{attempt\}\_\allowbreak{}t\{turn\}.txt in /reviewer/), the final submissions / cached test predictions, and cv\_\allowbreak{}scores.json. Only assess once enough attempts exist to see convergence (typically attempt index $\geq$ 2). & The attempts hedge across distinct modeling classes/architectures (e.g., GBDT ensemble in one, a neural net or a linear/kernel model or a very different feature representation in another), so the best-of-N beam covers meaningfully different solutions rather than one repeated recipe. & More than two of the attempts converge near finish on essentially the same solution -- same model family and features, near-identical final predictions or CV, often reusing the same cached OOF/test preds -- instead of at least one attempt deliberately trying a different modeling class to exploit best-of-N. & \emph{N/A:} Fewer than three attempts have completed, or the task is data-centric / single-model by nature so diversifying the model class is not meaningful. \emph{Insufficient:} The final solutions of the prior attempts cannot be compared from the transcripts, submissions, or registry. \\
\bottomrule
\end{tabular}}
\caption{Seed rubric definitions: exploration and model diversity. Text is verbatim as given to the PRA.}
\label{tab:rubric-nl-B}
\end{table*}

\begin{table*}[p]
\centering
\scriptsize
\resizebox{\textwidth}{!}{%
\begin{tabular}{@{}>{\raggedright\arraybackslash}p{3.3cm} >{\raggedright\arraybackslash}p{4.6cm} >{\raggedright\arraybackslash}p{4.6cm} >{\raggedright\arraybackslash}p{3.3cm}@{}}
\toprule
\textbf{Rubric and definition} & \textbf{Satisfied if} & \textbf{Violated if} & \textbf{Not applicable / insufficient evidence if} \\
\midrule
\textbf{Training Objective--Metric Alignment} (\texttt{metric\_\allowbreak{}objective\_\allowbreak{}alignment}). Whether the agent aligns the model's training objective or target transformation with the competition evaluation metric. & The training loss or target transformation matches the evaluation metric (e.g., log1p target for RMSLE, MAE objective for MedAE, stratified splits for imbalanced AUC, ordinal-aware loss for QWK). & Models are trained with a mismatched objective (e.g., RMSE on raw target for an RMSLE competition, MSE for QWK, no log transform for RMSLE), as evidenced in training scripts or cv\_\allowbreak{}scores.json metadata. & \emph{N/A:} The competition metric is accuracy or AUC and any standard classification objective is acceptable. \emph{Insufficient:} Training scripts are not visible and metric alignment cannot be inferred. \\
\midrule
\textbf{Domain-Appropriate Preprocessing} (\texttt{domain\_\allowbreak{}appropriate\_\allowbreak{}preprocessing}). Whether the agent applies preprocessing steps appropriate to the model type and task (avoiding scaling before tree models, using correct imputation, appropriate CV splits). & Preprocessing choices are appropriate: no feature scaling before tree-based models, correct imputation strategy for the target distribution, and CV split type matches the task (stratified for imbalanced classification, time-based for time series). & StandardScaler is applied before LightGBM/XGBoost/CatBoost, target NaN values are imputed with 0 for a forecasting task, or random KFold is used for a time-series problem where temporal leakage would inflate CV scores. & \emph{N/A:} Only linear models were used (where scaling is appropriate) or there were no categorical/missing-value features. \emph{Insufficient:} Preprocessing code is not visible. \\
\midrule
\textbf{Feature Engineering Investment} (\texttt{feature\_\allowbreak{}engineering\_\allowbreak{}investment}). Whether the agent invests in at least one round of principled feature engineering beyond raw input columns, accounting for feature work already done in earlier best-of-N attempts. \emph{Context:} This is a best-of-N run (N=n\_\allowbreak{}attempts). Consult the accumulated prior-attempt transcripts (traj\_\allowbreak{}a\{attempt\}\_\allowbreak{}t\{turn\}.txt in /reviewer/), features.py/preprocess.py in the persistent workspace, and cv\_\allowbreak{}scores.json to see what feature engineering earlier attempts already tried and whether it helped. & EITHER the workspace contains a feature-engineering script/module adding at least 3 non-trivial derived features (interactions, target encodings, lag features, domain transforms) with a CV comparison showing improvement over the raw-feature baseline, OR this attempt sensibly builds on / diverges from the feature work prior attempts established (e.g., a prior attempt found feature set A helped, so this attempt reuses A and explores a different modeling axis, or a prior attempt found FE gave no lift so this attempt reasonably invests its budget elsewhere). & All training scripts use only raw input columns (label-encoding/fillna only) and NO attempt -- this one or an earlier one -- engineered features despite a non-trivial feature space; OR this attempt merely re-derives the exact same features a prior attempt already built and validated, adding nothing new to the beam. & \emph{N/A:} The competition is data-centric (fixed model) or the feature space is already maximal (e.g., pre-extracted embeddings). \emph{Insufficient:} Feature-engineering code and the prior-attempt transcripts cannot be inspected. \\
\bottomrule
\end{tabular}}
\caption{Seed rubric definitions: preprocessing, target, and features. Text is verbatim as given to the PRA.}
\label{tab:rubric-nl-C}
\end{table*}

\begin{table*}[p]
\centering
\scriptsize
\resizebox{\textwidth}{!}{%
\begin{tabular}{@{}>{\raggedright\arraybackslash}p{3.3cm} >{\raggedright\arraybackslash}p{4.6cm} >{\raggedright\arraybackslash}p{4.6cm} >{\raggedright\arraybackslash}p{3.3cm}@{}}
\toprule
\textbf{Rubric and definition} & \textbf{Satisfied if} & \textbf{Violated if} & \textbf{Not applicable / insufficient evidence if} \\
\midrule
\textbf{OOF and Test Prediction Caching} (\texttt{oof\_\allowbreak{}prediction\_\allowbreak{}caching}). Whether the agent caches OOF and test predictions as parquet files after each model training run to enable fast ensemble search without retraining. & The workspace preds/ directory contains OOF and test prediction files (parquet or similar) for every substantively trained model, and ensemble/stacking scripts load these cached files rather than retraining base models. & No OOF or test prediction files exist for trained models, or ensemble scripts retrain base models from scratch instead of loading cached predictions. & \emph{N/A:} Only a single model was trained and no ensembling was attempted. \emph{Insufficient:} The workspace preds/ directory contents cannot be verified from the available information. \\
\midrule
\textbf{Submission File Integrity Verification} (\texttt{submission\_\allowbreak{}integrity\_\allowbreak{}verification}). Whether the agent explicitly verifies submission file integrity (row count, NaN absence, value range, column names) before concluding the attempt. & The trajectory contains an explicit check of the final /submission/submission.csv covering at least row count matching the test set size and absence of NaN values, with results logged. & No integrity check is performed, or the check reveals problems (wrong row count, NaNs, binary corruption, wrong column names) that are not corrected before finishing. & \emph{N/A:} The attempt ended before any submission file was generated. \emph{Insufficient:} It cannot be determined whether a verification step occurred. \\
\midrule
\textbf{CV Regression Detection and Revert} (\texttt{regression\_\allowbreak{}detection\_\allowbreak{}and\_\allowbreak{}revert}). Whether the agent detects when a subsequent experiment produces a CV regression and explicitly reverts the submission to the best-validated model. & The agent detects (via cv\_\allowbreak{}scores.json comparison or file inspection) that the current submission.csv corresponds to a worse model than a previously computed best, and explicitly restores the best model's predictions to /submission/submission.csv before finishing. & The final submission reflects a worse model because the agent did not check for regressions, or detected a regression but failed to revert before finishing. & \emph{N/A:} Only one model was trained, making regression impossible. \emph{Insufficient:} It cannot be determined whether a regression occurred or was addressed. \\
\midrule
\textbf{Avoiding Repeated Identical Crashes} (\texttt{no\_\allowbreak{}repeated\_\allowbreak{}identical\_\allowbreak{}crashes}). Whether the agent diagnoses and fixes the root cause of a crash rather than repeatedly re-running the same failing code. & When a script crashes, the agent identifies the root cause, applies a fix, and the same error does not recur more than once after the fix. & The same error (same exception type and cause, e.g., wrong keyword argument, dtype mismatch, array shape error) appears three or more times across successive turns, indicating the agent is re-running without root-cause diagnosis. & \emph{N/A:} No crashes occurred during the trajectory. \emph{Insufficient:} Error messages and fix attempts are not visible in the transcript. \\
\bottomrule
\end{tabular}}
\caption{Seed rubric definitions: efficiency, robustness, and submission (part 1 of 2). Text is verbatim as given to the PRA.}
\label{tab:rubric-nl-E1}
\end{table*}

\begin{table*}[p]
\centering
\scriptsize
\resizebox{\textwidth}{!}{%
\begin{tabular}{@{}>{\raggedright\arraybackslash}p{3.3cm} >{\raggedright\arraybackslash}p{4.6cm} >{\raggedright\arraybackslash}p{4.6cm} >{\raggedright\arraybackslash}p{3.3cm}@{}}
\toprule
\textbf{Rubric and definition} & \textbf{Satisfied if} & \textbf{Violated if} & \textbf{Not applicable / insufficient evidence if} \\
\midrule
\textbf{Avoiding Churn Without Pivoting} (\texttt{no\_\allowbreak{}churn\_\allowbreak{}without\_\allowbreak{}pivot}). Whether the agent recognizes when an experimental axis has plateaued -- WITHIN this attempt or ACROSS the best-of-N attempts -- and pivots to a substantively different approach rather than repeating near-identical experiments. \emph{Context:} This is a best-of-N run (N=n\_\allowbreak{}attempts). Assess churn at BOTH scales, using the accumulated prior-attempt transcripts (traj\_\allowbreak{}a\{attempt\}\_\allowbreak{}t\{turn\}.txt in /reviewer/), cv\_\allowbreak{}scores.json, and the submission ledger to reconstruct what earlier attempts already tried. & Within the attempt, after 2-3 consecutive experiments on the same axis (weight search, seed sweep) yield no meaningful CV improvement ($<$ 0.001 relative) the agent pivots to a different approach (new model family, feature engineering, different stacking architecture); AND, across attempts, this attempt does not merely repeat an axis a prior attempt already saturated -- it explores something the beam has not tried. & EITHER five or more consecutive experiments within the attempt make no substantive change and no meaningful CV gain (nearly identical cv\_\allowbreak{}scores.json entries or repeated 'No improvement found'), OR this attempt re-runs the same saturated axis a prior attempt already exhausted (e.g., yet another ensemble-weight sweep on the same cached OOF predictions) instead of pivoting -- the classic best-of-N beam collapse where multiple attempts converge on the same solution. & \emph{N/A:} The agent only ran one or two experiments total and no prior attempt covered this axis. \emph{Insufficient:} The number and variety of experiments cannot be determined from the transcript or the prior-attempt transcripts are unavailable. \\
\midrule
\textbf{Valid Submission Before Finishing} (\texttt{no\_\allowbreak{}premature\_\allowbreak{}termination}). Whether the agent ensures a valid, complete submission file exists at /submission/submission.csv before concluding the attempt. & At the time of finish, /submission/submission.csv exists, is a valid CSV with the correct target column and row count matching the test set, and contains non-trivial (non-constant, non-NaN) predictions. & finish is called before any valid submission is written, or the submission file is empty, binary-corrupted, or contains wrong-competition predictions that make it invalid for the actual leaderboard. & \emph{N/A:} The attempt is still in progress and no finish was called. \emph{Insufficient:} The submission file contents cannot be verified. \\
\midrule
\textbf{Timeout-Aware Experiment Design} (\texttt{timeout\_\allowbreak{}aware\_\allowbreak{}experiment\_\allowbreak{}design}). Whether the agent designs experiments to complete within available time budgets, using prototyping or parameter adjustment rather than launching full training jobs that are repeatedly killed. & Before launching expensive training jobs (large datasets, many folds, many boosting rounds), the agent either runs a single-fold prototype to estimate runtime, or explicitly limits parameters (sample size, n\_\allowbreak{}folds, n\_\allowbreak{}estimators) to fit within the time budget -- and no major training job is killed before saving results. & Two or more training scripts are interrupted with Ctrl-C or timeout before saving any useful predictions or CV scores, indicating the agent launched jobs without verifying they would complete. & \emph{N/A:} All training jobs completed successfully without interruption. \emph{Insufficient:} Job runtimes and completion status are not visible in the transcript. \\
\midrule
\textbf{Over-Optimization Without Significant Gains} (\texttt{over\_\allowbreak{}optimization}). Whether the agent avoids burning a disproportionate share of turns on hyperparameter tuning and ensemble-weight search that yields no statistically meaningful CV improvement, judged within the current attempt AND across the best-of-N attempts. \emph{Context:} This is a best-of-N run (N=n\_\allowbreak{}attempts); the submitted result is the MAX over attempts, so grinding for a 4th-decimal CV gain rarely changes the outcome while it burns budget that a different approach could use. 'Statistically meaningful' means the CV gain exceeds the fold-to-fold noise: compare the improvement to cv\_\allowbreak{}std in cv\_\allowbreak{}scores.json (a rule of thumb: a gain smaller than \textasciitilde{}0.25-0.5x cv\_\allowbreak{}std, or $<$ \textasciitilde{}0.001 relative when std is unavailable, is inside the noise). Use the accumulated prior-attempt transcripts (traj\_\allowbreak{}a\{attempt\}\_\allowbreak{}t\{turn\}.txt in /reviewer/) and cv\_\allowbreak{}scores.json to judge cumulative effort across attempts, not just this one. & Time spent on hyperparameter tuning / ensemble-weight search is proportionate -- the agent stops optimizing an axis once gains fall within cv\_\allowbreak{}std and redirects effort to a new model family, features, or a fresh attempt; the beam's total tuning effort is not dominated by noise-level micro-optimization. & A large share of turns (roughly a third or more of the attempt, or the bulk of multiple attempts) goes to HPO or ensemble-weight tuning whose cumulative CV gain is within the noise floor ($<$ cv\_\allowbreak{}std, e.g. AUC moving 0.8671$\rightarrow$0.8672 across 30+ turns, or repeated Optuna runs each landing within std of the last) -- effort that best-of-N makes almost worthless. & \emph{N/A:} The agent did little or no hyperparameter/ensemble tuning, or tuning produced clearly meaningful ($>$ cv\_\allowbreak{}std) gains. \emph{Insufficient:} cv\_\allowbreak{}std / CV deltas or the turn-level breakdown of effort cannot be recovered from the transcript, workspace, or prior-attempt transcripts. \\
\bottomrule
\end{tabular}}
\caption{Seed rubric definitions: efficiency, robustness, and submission (part 2 of 2). Text is verbatim as given to the PRA.}
\label{tab:rubric-nl-E2}
\end{table*}

\section{Seed Rubric Failures Across Benchmarks}
\label{sec:appendix-rubric-transfer}
Table~\ref{tab:rubric-transfer} compares the 19 seed rubrics (Appendix~\ref{sec:appendix-rubrics}) in the $-$adaptive rubrics runs on both benchmarks. Both runs score the same rubrics at every review, so the rates are directly comparable. A rubric fails in a review when it crosses the injection thresholds (Table~\ref{tab:run-config}). The analysis covers 190 DSPredict-Easy attempts and 270 DSPredict-Hard attempts. Across the 19 rubrics, Spearman correlations between benchmarks are $\rho=0.66$ ($p=0.002$) for attempt failure rate and $\rho=0.91$ ($p<10^{-7}$) for mean importance.

\begin{table}[t]
\centering
\small
\resizebox{\columnwidth}{!}{%
\begin{tabular}{@{}>{\raggedright\arraybackslash}p{3.6cm} cc cc@{}}
\toprule
 & \multicolumn{2}{c}{\textbf{Failed attempts (\%)}} & \multicolumn{2}{c}{\textbf{Mean importance}} \\
\cmidrule(lr){2-3} \cmidrule(lr){4-5}
\textbf{Seed rubric} & \textbf{Easy} & \textbf{Hard} & \textbf{Easy} & \textbf{Hard} \\
\midrule
Model Diversity and Ensembling & 83.2 & 74.4 & 0.66 & 0.63 \\
Avoiding Churn Without Pivoting & 83.7 & 68.1 & 0.58 & 0.61 \\
Best-CV Model Submission Selection & 68.4 & 73.7 & 0.71 & 0.70 \\
CV Regression Detection and Revert & 67.9 & 69.3 & 0.64 & 0.63 \\
Hedging Across Best-of-N With Diverse Model Classes & 66.8 & 52.2 & 0.53 & 0.55 \\
Timeout-Aware Experiment Design & 55.3 & 54.4 & 0.42 & 0.48 \\
Hyperparameter Tuning Beyond Defaults & 49.5 & 44.4 & 0.50 & 0.45 \\
Feature Engineering Investment & 39.5 & 51.9 & 0.51 & 0.57 \\
Valid Submission Before Finishing & 36.8 & 54.4 & 0.56 & 0.66 \\
Avoiding Repeated Identical Crashes & 44.7 & 46.3 & 0.40 & 0.47 \\
Valid Cross-Validation Methodology & 32.1 & 56.7 & 0.78 & 0.75 \\
Over-Optimization Without Significant Gains & 58.4 & 26.3 & 0.42 & 0.36 \\
Submission File Integrity Verification & 32.6 & 51.9 & 0.71 & 0.75 \\
Suspicious CV Score Investigation & 30.0 & 50.4 & 0.35 & 0.43 \\
Metric Calculation Bug Detection & 28.9 & 46.7 & 0.39 & 0.47 \\
OOF and Test Prediction Caching & 31.6 & 43.3 & 0.62 & 0.57 \\
Persistent CV Score Registry & 25.3 & 48.5 & 0.70 & 0.69 \\
Training Objective--Metric Alignment & 16.8 & 33.0 & 0.53 & 0.60 \\
Domain-Appropriate Preprocessing & 12.1 & 24.1 & 0.42 & 0.48 \\
\bottomrule
\end{tabular}}
\caption{Seed rubric failures in the frozen-rubric runs. \emph{Failed attempts}: share of attempts in which the rubric failed at least one review. \emph{Mean importance}: the PRA's mean importance score for the rubric over all reviews.}
\label{tab:rubric-transfer}
\end{table}

\paragraph{Examples of transferred rubrics.} Table~\ref{tab:rubric-transfer-examples} pairs seven failures that the PRA flagged on non-tabular DSPredict-Hard tasks with the same rubric failing on a tabular DSPredict-Easy task. Each example comes from the frozen-rubric run, so the rubric text is exactly as discovered on DSPredict-Easy. We count a flag as \emph{resolved} when the PRA scores a rubric at importance $\geq 0.6$ and success probability $\leq 0.2$ at one review, and at success probability $\geq 0.7$ at a later review of the same attempt; for every example in the table we read the later rationale to confirm that the agent fixed the issue. Across the DSPredict-Hard frozen-rubric run, 685 of 1{,}277 flags (54\%) are resolved within the attempt, on 53 of the 54 tasks. Resolution rates are highest for rubrics that check the pipeline (submission integrity 83\%, regression detection 82\%, best-model selection 79\%, metric bugs 67\%, valid cross-validation 55\%) and lowest for the simplicity-bias rubrics (model diversity 16\%, single-model-class hedging 22\%, pivoting when experiments stall 28\%). Flagged pipeline errors are usually fixed within the attempt, while flagged strategy problems change the agent's behavior less often.

\begin{table*}[t]
\centering
\small
\begin{tabular}{@{}>{\raggedright\arraybackslash}p{2.4cm} >{\raggedright\arraybackslash}p{2.9cm} >{\raggedright\arraybackslash}p{5.4cm} >{\raggedright\arraybackslash}p{3.6cm}@{}}
\toprule
\textbf{Seed rubric} & \textbf{Hard task (modality)} & \textbf{Flagged behavior $\rightarrow$ fix on DSPredict-Hard} & \textbf{DSPredict-Easy counterpart} \\
\midrule
Valid Cross-Validation Methodology & LANL Earthquake (seismic time series) & Random k-fold on a time series (``temporal leakage'') and ``OOF'' predictions from the final model on its own training data $\rightarrow$ time-based folds with true out-of-fold predictions. & s5e2: OOF predictions computed in-sample by the final model. \\
\addlinespace
Suspicious CV Score Investigation & Statoil Iceberg (radar imagery) & CV improves 57\% ``with NO model change'' $\rightarrow$ the agent finds and removes the leak; CV returns to a plausible 0.41. & s3e21: CV improves 63.5\% ``with no architectural explanation''. \\
\addlinespace
Submission File Integrity Verification & TensorFlow Speech (audio) & Required submission file missing, workspace copy uses 30 wrong class labels $\rightarrow$ regenerated with all 12 required labels. & s3e19: required submission file missing. \\
\addlinespace
CV Regression Detection and Revert & Data Science Bowl 2018 (microscopy segmentation) & New submission drops sample mAP from 0.17 to 0.0002 $\rightarrow$ reverts to the best U-Net checkpoint (mAP 0.31). & s3e1: a Ridge script overwrites the best XGBoost submission with one 77\% worse. \\
\addlinespace
Training Objective--Metric Alignment & Novozymes Enzyme Stability (protein sequences) & Competition scores Spearman correlation, agent trains and validates on RMSE $\rightarrow$ tracks cross-validated Spearman. & s3e14: competition scores MAE, models trained on default MSE. \\
\addlinespace
Domain-Appropriate Preprocessing & GeoLifeCLEF 2024 (geospatial species presence) & KNN on unscaled features with a 3{,}432$\times$ range in scale $\rightarrow$ adds standardization, CV improves. & s3e25: MLP and SVR trained without feature scaling. \\
\addlinespace
Metric Calculation Bug Detection & NLP Getting Started (text classification) & F1 value reused as the decision threshold of a linear SVM, predicting 19\% positives against a 43\% base rate $\rightarrow$ threshold corrected, 42\% positives. & s4e11: one reported score for every model, including a model known to score higher. \\
\bottomrule
\end{tabular}
\caption{Seed rubrics discovered on DSPredict-Easy catching the same process failures on non-tabular DSPredict-Hard tasks, from the frozen-rubric runs. Each DSPredict-Hard flag was resolved later in the same attempt, confirmed from the PRA's rationale at the later review. Quoted phrases are from the PRA's rationales.}
\label{tab:rubric-transfer-examples}
\end{table*}

\section{Rubric Store Operations}
\label{sec:appendix-rubric-ops}

\begin{figure}[t]
\centering
\includegraphics[width=\columnwidth]{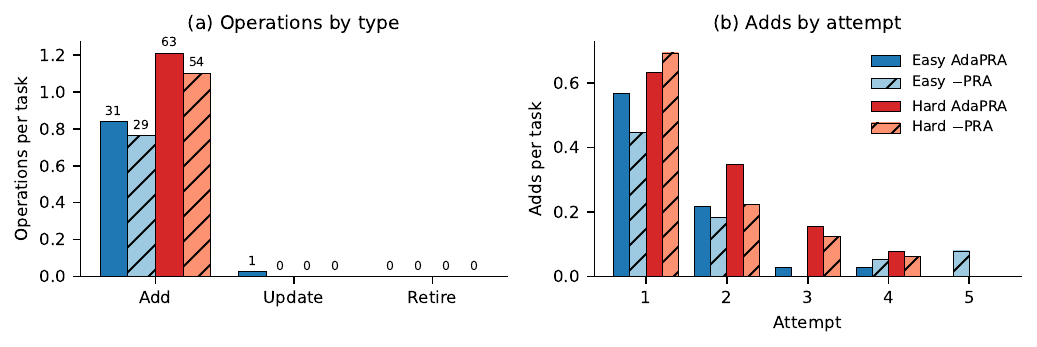}
\caption{Rubric store operations per task for \methodname{} and the $-$PRA ablation. (a) Operations by type, with raw counts above the bars. (b) Additions by the attempt in which they occur. Rates are over tasks with an operation log (DSPredict-Easy: 37 and 38 tasks; DSPredict-Hard: 52 and 49 tasks, for \methodname{} and $-$PRA).}
\label{fig:rubric-store-ops}
\end{figure}

Figure~\ref{fig:rubric-store-ops} counts the operations that the PRA applies to each task's rubric store (\S\ref{sec:adaptive}) in \methodname{} and its $-$PRA ablation, taken from the operation log of each store. Adaptation consists almost entirely of additions. Across the four runs the PRA adds 177 rubrics, updates one, and retires none. Between 81\% and 94\% of additions occur in the first two attempts, so later attempts are reviewed against a largely stable store. The $-$PRA ablation adds rubrics at a similar rate (0.76 vs.\ 0.84 per task on DSPredict-Easy and 1.10 vs.\ 1.21 on DSPredict-Hard), so removing code execution does not reduce how often the store is edited.

\begin{figure}[t]
\centering
\includegraphics[width=\columnwidth]{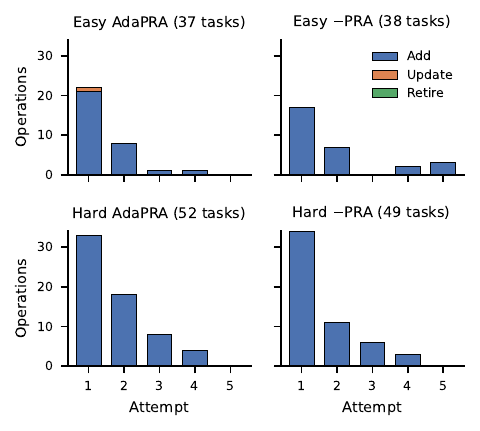}
\caption{Rubric store operations per attempt for each run, stacked by type. Panel titles give the number of tasks with an operation log. The only update occurs in the first attempt of \methodname{} on DSPredict-Easy, and no run retires a rubric.}
\label{fig:rubric-ops-by-attempt}
\end{figure}

\begin{figure}[t]
\centering
\includegraphics[width=\columnwidth]{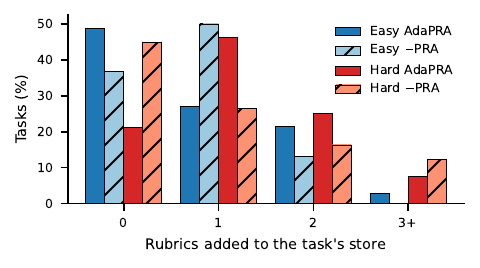}
\caption{Share of tasks by the number of rubrics the PRA adds to the task's store over its five attempts, over tasks with an operation log.}
\label{fig:rubric-adds-per-task}
\end{figure}

Figure~\ref{fig:rubric-ops-by-attempt} breaks these operations down by run and attempt, and Figure~\ref{fig:rubric-adds-per-task} shows how many rubrics each task's store gains. With code execution the PRA adds at least one rubric on 79\% of DSPredict-Hard tasks, compared with 55\% for $-$PRA, while on DSPredict-Easy the shares are 51\% and 63\%. Table~\ref{tab:rubric-themes} groups the added rubrics by what they check, a grouping we assigned by reading each rubric. Prediction-sanity rubrics, which compute statistics of the agent's predictions such as their dispersion relative to the target, account for 35\% (DSPredict-Easy) and 19\% (DSPredict-Hard) of additions with code execution and 3\% and 7\% without it, and the $-$PRA ablation instead adds mostly train--test and submission consistency rubrics (41\% and 44\%).

\begin{table*}[t]
\centering
\small
\begin{tabular}{@{}l cc c cc p{7.2cm}@{}}
\toprule
 & \multicolumn{2}{c}{\textbf{DSPredict-Easy}} & & \multicolumn{2}{c}{\textbf{DSPredict-Hard}} & \\
\cmidrule(lr){2-3} \cmidrule(lr){5-6}
\textbf{Theme} & \methodname{} & $-$PRA & & \methodname{} & $-$PRA & \textbf{Examples} \\
\midrule
Validation and leakage & 7 & 8 & & 25 & 10 & Calibration fit and scored on the same out-of-fold predictions (s3e26); quantum-mechanical sub-components of the target used as features (CHAMPS Scalar Coupling). \\
\addlinespace
Train--test and submission consistency & 3 & 12 & & 16 & 24 & Ensemble test predictions weighted differently from the scored out-of-fold blend (s5e2, $-$PRA); submission built on the Stage 1 template (March Machine Learning Mania 2025). \\
\addlinespace
Prediction sanity & 11 & 1 & & 12 & 4 & Out-of-fold predictions compressed toward the target mean (s3e15); negative time-to-failure predictions (LANL Earthquake Prediction). \\
\addlinespace
Features and modeling choices & 8 & 2 & & 8 & 9 & No log transform of a skewed price target under RMSE (s4e9); no building-level target encoding (ASHRAE Energy Prediction). \\
\addlinespace
Crashing code & 2 & 6 & & 2 & 7 & \texttt{predict\_proba} called on a ridge classifier (s3e26, $-$PRA); training script calls an unimported helper (Humpback Whale Identification, $-$PRA). \\
\midrule
Total & 31 & 29 & & 63 & 54 & \\
\bottomrule
\end{tabular}
\caption{Rubrics added by the PRA, grouped by what they check, as counts per run. Themes were assigned by hand from each rubric's name and description. Each example gives the task, and $-$PRA marks rubrics added by the ablation.}
\label{tab:rubric-themes}
\end{table*}

\section{Example PRA Feedback}
\label{sec:appendix-prm-examples}

This appendix reproduces PRA feedback from the \methodname{} runs on DSPredict-Easy and DSPredict-Hard, with the agent's response and the leaderboard outcome of the attempt. Each PRA box shows the failed rubrics as the agent received them (\S\ref{sec:pra}). Each agent box shows the agent's subsequent actions, with each step labeled by turn and long passages abridged as ``[...]''. Attempts are numbered from 1 to 5, percentiles are on the private leaderboard unless the competition has only a public one, and CV denotes the agent's local cross-validation score.

\paragraph{Execution grounding exposes an in-sample calibration gain.}
In \emph{Feature Imputation with a Heat Flux Dataset}, the agent fitted isotonic regression on its full set of XGBoost out-of-fold predictions, scored it on the same rows, and submitted it because the RMSE fell from 0.073994 to 0.073592. The PRA reran the calibration under nested cross-validation and found an RMSE of 0.074386, worse than the uncalibrated model. It then added a task-specific rubric, \texttt{calibration\_\allowbreak{}not\_\allowbreak{}insample}, with a natural-language definition and a check script (\texttt{calibration\_\allowbreak{}insample\_\allowbreak{}check.py}) that fits the calibrator on four folds and scores the fifth (\S\ref{sec:adaptive}). The agent reverted the submission in the next two turns and finished the attempt with a Ridge stack of LightGBM, XGBoost, and CatBoost (CV 0.073815). The private percentile rose from 49.7 in attempt 1 to 62.2 in attempt 2, while the public percentile was unchanged (52.4 and 52.0).

\begin{prmbox}{PRA feedback: s3e15, attempt 2, turn 60 (DSPredict-Easy)}
### ISSUES TO FIX (address these in your remaining steps):
1. **Valid Cross-Validation Methodology** [Importance: 0.90 | Success: 0.10]:
   - Problem: CRITICAL: Agent fitted IsotonicRegression on full 21229 XGBoost OOF
     predictions AND evaluated on the same data. In-sample RMSE=0.073592. Proper
     nested-CV RMSE=0.074386 (measured by reviewer). [...] FAIL.
2. **Best-CV Model Submission Selection** [Importance: 0.90 | Success: 0.10]:
   [...]
   - Suggested action: Revert submission immediately: load preds_xgb.parquet and write
     to /submission/submission.csv. True calibrated RMSE=0.074386 > registered
     xgb=0.073994.
[...]
4. **calibration_not_insample** [Importance: 0.90 | Success: 0.05]:
   - Definition: Whether post-hoc calibration layers (isotonic regression, Platt
     scaling, temperature scaling) applied to model predictions are validated using a
     proper held-out or nested cross-validation scheme - NOT by fitting and evaluating
     on the same OOF data. [...] Failure looks like: in-sample calibration RMSE appears
     better than uncalibrated OOF RMSE, but nested-CV calibration RMSE is WORSE [...]
   - Problem: calibration_insample_check.py: Uncalibrated RMSE=0.073997. In-sample cal
     RMSE=0.073592 (OPTIMISTIC). Nested-CV true RMSE=0.074386 (WORSE). Agent used
     in-sample result to submit. FAIL - calibration overfits isotonic regression on
     full OOF data.
[...]
\end{prmbox}

\begin{agentbox}{Agent response (turns 61-103)}
[turn 61] The PRM feedback is very clear and critical. I need to:
1. Revert the submission to the uncalibrated XGBoost (best model with CV=0.073994)
2. Never use in-sample calibration - need nested-CV for proper validation
[...]
[turn 62]
xgb_test = pd.read_parquet('/workspace/playground-series-s3e15/preds/preds_xgb.parquet')
[...]
submission.to_csv('/submission/submission.csv', index=False)
print("Reverted to uncalibrated XGBoost submission (best model)")
[...]
[turn 79] Excellent! The Ridge ensemble achieved a CV score of 0.073815 +/- 0.000715,
which is better than the best single model (XGB: 0.073994 +/- 0.000689). [...]
[...]
[final answer]
## Validation Score
0.073815 (RMSE) - ensemble_ridge model
\end{agentbox}

\paragraph{A task-specific rubric bounds a weak formulation before it runs.}
In the \emph{TGS Salt Identification Challenge}, the agent began attempt 1 by training logistic regression on global image statistics to predict whether an image contains salt, and planned to submit an all-salt or all-empty mask per image. Before this model produced a score, the PRA computed the salt fill fraction of every training mask and derived that even a perfect image-level classifier is capped at 0.52 mean average precision under the competition's IoU thresholds. It recorded this bound as a new rubric, \texttt{pixel\_\allowbreak{}segmentation\_\allowbreak{}vs\_\allowbreak{}image\_\allowbreak{}classification}, backed by \texttt{check\_\allowbreak{}segmentation\_\allowbreak{}approach.py}. The agent replaced the classifier with a U-Net trained with per-pixel cross-entropy within five turns. The rubric passed in all 21 subsequent reviews recorded for the task, which confirmed from the submitted run-length encodings that every attempt predicted image-specific masks. Two further rubrics added in attempt 1 had a similar lasting effect. One of them (\texttt{oof\_\allowbreak{}is\_\allowbreak{}fold\_\allowbreak{}predictions\_\allowbreak{}not\_\allowbreak{}full\_\allowbreak{}model}) caught at turn 100 that the saved out-of-fold array came from the model refit on all data, since it reproduced a score of 0.1654 against a reported fold CV of 0.5005. Its check then verified every later CV to within $10^{-4}$. Attempt 1 itself, with a five-epoch U-Net, reached a private percentile of only 10.1. Later attempts built on its pipeline with ResNet encoders, reached CV above 0.70, and scored private percentiles of 34.3 to 36.6.

\begin{prmbox}{PRA feedback: tgs-salt, attempt 1, turn 40 (DSPredict-Hard, segmentation)}
### ISSUES TO FIX (address these in your remaining steps):
1. **pixel_segmentation_vs_image_classification** [Importance: 0.85 | Success: 0.20]:
   - Definition: Whether the agent produces pixel-level segmentation masks rather than
     image-level classification for the TGS salt challenge. The competition metric is
     mean AP over IoU thresholds 0.5-0.95, which requires accurate pixel-level masks. An
     image-level classifier that predicts all-ones or all-zeros masks has a mathematical
     ceiling of ~0.52 AP (measured: mean salt fill=0.407, fraction with fill>=0.5=0.382)
     vs ~0.88 achievable with U-Net-style pixel segmentation - a gap of 0.36. [...]
   - Problem: check_segmentation_approach.py: agent uses Logistic Regression on global
     image statistics (mean, std, percentiles) to classify images as salt/no-salt, then
     submits all-ones or all-zeros 101x101 masks. [...] Perfect image-level classifier
     AP ceiling = 0.5228 vs ~0.88 for U-Net segmentation. Gap = 0.357.
   - Suggested action: Replace image-level LR classification with a pixel-level
     segmentation model (e.g. U-Net with a CNN encoder). Global image statistics cannot
     predict where salt pixels are - the model must output a 101x101 mask per image,
     not a binary label.
2. **Training Objective-Metric Alignment** [Importance: 0.80 | Success: 0.20]:
   [...]
   - Suggested action: Use a pixel-level segmentation loss (e.g., binary cross-entropy
     per pixel, Dice loss) to train a model that outputs a 101x101 mask, not a scalar
     probability.
\end{prmbox}

\begin{agentbox}{Agent response (turns 41-51)}
[turn 41] The feedback is critical - I need to build a pixel-level segmentation model,
not an image-level classifier. [...] I need to:
1. Train a model that outputs a 101x101 mask per image
2. Use pixel-level loss (binary cross-entropy per pixel, Dice loss)
[...]
[turn 42] [str_replace] train_baseline.py
-Uses Logistic Regression with image and depth features.
+Uses a simple CNN for pixel-level salt segmentation.
[...]
[turn 45] The U-Net has a size mismatch issue. [...]
- After 3 maxpool operations (stride 2): 101 -> 50 -> 25 -> 12 (floor division)
[...]
[turn 51] The model trained successfully with a CV score of 0.4815. [...]
\end{agentbox}

\paragraph{The PRA prevents a revert caused by the agent's own diagnostic bug.}
In \emph{Multi-Class Prediction of Obesity Risk}, attempts 1 and 2 had trained only gradient-boosted trees, and in attempt 3 the score registry had fallen out of sync with the prediction files. The current submission came from XGBoost with an out-of-fold accuracy of 0.909240, but the agent's own accuracy script compared class labels against column names carrying a \texttt{prob\_\allowbreak{}} prefix and reported XGBoost as 0.000000. The PRA reproduced the correct accuracy from the saved predictions, located the prefix bug, and warned the agent against reverting to the weaker CatBoost model. The agent applied the one-line fix, obtained 0.909240, and corrected the registry within four turns. It followed the PRA's diversification requests by training an MLP, which timed out, and a logistic regression (CV 0.8569), which the PRA's review at turn 50 found would lower accuracy in a hard vote with XGBoost. The final submission was a 0.9/0.1 soft vote of XGBoost and CatBoost with CV 0.910059, a weighting the PRA had computed from the cached predictions at turn 60. The private percentile rose from 49.4 in attempt 2 to 73.3 in attempt 3.

\begin{prmbox}{PRA feedback: s4e2, attempt 3, turn 30 (DSPredict-Easy)}
### ISSUES TO FIX (address these in your remaining steps):
1. **CV Regression Detection and Revert** [Importance: 0.95 | Success: 0.15]:
   - Problem: MOST URGENT: Agent's turn 28 code shows xgb_baseline OOF=0.000000 (bug
     described above). Agent at turn 29-30 is investigating test predictions - likely
     about to conclude XGBoost is broken and revert submission to catboost_baseline
     (0.9038). This would be a self-inflicted regression of -0.0054 from the current
     correct XGBoost submission. The XGBoost submission IS correct - the OOF computation
     code is wrong, not the model. [...] DO NOT REVERT.
   - Suggested action: STOP before reverting. Run: [...] you will see 0.909240,
     confirming XGBoost is the BEST model. Keep the current submission.
2. **Persistent CV Score Registry** [Importance: 0.92 | Success: 0.15]:
   - Problem: cv_scores.json STILL missing xgb_baseline. best_ever=catboost_baseline=
     0.903796. Agent at turn 28 computed XGBoost OOF=0.000000 (bug:
     oof[prob_cols].idxmax() gives labels like 'prob_Normal_Weight', compared against
     'Normal_Weight' without stripping 'prob_' prefix - all mismatches -> accuracy=0).
     Actual XGBoost OOF=0.909240 (reviewer-verified from parquet). [...]
   - Suggested action: Fix the XGBoost OOF computation: oof['pred'] =
     oof[prob_cols].idxmax(axis=1).str.replace('prob_','',regex=False). [...] Do NOT
     re-train XGBoost - the preds are correct; only the registry is missing the entry.
[...]
\end{prmbox}

\begin{agentbox}{Agent response (turns 31-34)}
[turn 31] The expert feedback confirms that:
1. XGBoost OOF=0.909240 (best model) - I had a bug in my OOF computation code
2. The submission IS from XGBoost and is correct
3. cv_scores.json is missing the xgb_baseline entry
4. I should NOT revert to catboost_baseline
[turn 32]
oof['pred'] = oof[prob_cols].idxmax(axis=1).str.replace('prob_', '', regex=False)
[...]
>> XGBoost OOF accuracy (corrected): 0.909240
[turn 34]
cv['best_ever'] = {
    'model_name': 'xgb_baseline',
    'cv_score': 0.909240,
[...]
\end{agentbox}

\paragraph{The finish gate catches a CV that the submission does not reproduce.}
In \emph{Digit Recognizer}, the agent called finish at turn 44 with a ResNet-18 and a reported CV of 0.9922. At the finish review, the PRA recomputed accuracy from the saved out-of-fold predictions (0.9899) and traced the gap to the training script. The script recorded best-epoch validation accuracy as the CV score but generated out-of-fold and test predictions from the last-epoch weights, and it never reloaded the best-epoch checkpoints. The failed review returned the attempt to the agent for its one refinement round (\S\ref{sec:pra}), in which the agent reloaded the checkpoints, retrained, and finished with an out-of-fold accuracy of 0.9912 that matched the reported score. The attempt reached a public percentile of 87.0 (accuracy 0.99482; the competition has no private leaderboard), up from 73.3 for attempt 4 (a three-layer CNN) and 21.7 to 22.1 for attempts 1 to 3. The gain over attempt 4 also reflects the change of architecture, so it cannot be attributed to the fix alone. Earlier in the same attempt (turn 20), the PRA had found that the submission file matched the XGBoost predictions row for row while the CNN held the best CV, and the agent restored the CNN submission in the next turn.

\begin{prmbox}{PRA feedback: digit-recognizer, attempt 5, turn 44, finish gate (DSPredict-Hard, vision)}
### ISSUES TO FIX (address these in your remaining steps):
1. **Valid Cross-Validation Methodology** [Importance: 0.90 | Success: 0.15]:
   - Problem: ResNet OOF reproduced=0.989881 vs reported=0.992190, delta=0.002309. Root
     cause confirmed in train_resnet.py (lines 140-203): cv_scores.append(
     best_val_acc/100.0) records best-epoch validation accuracy, but OOF/test inference
     (lines 195, test_predictions += test_fold/n_splits) uses the CURRENT
     (last/early-stopped) model - best weights are saved to .fold{N} files but never
     reloaded before inference. So the reported CV=0.9922 describes the best-epoch
     model, while submission/OOF come from the last-epoch model (0.9899). [...]
   - Suggested action: In train_resnet.py, after the training loop for each fold, add:
     `model.load_state_dict(torch.load(f'{MODEL_PATH}.fold{fold+1}'))` then
     `model.eval()` BEFORE computing oof_fold and test_fold. This makes OOF/submission
     correspond to best-epoch weights and the reported CV will be reproducible.
2. **Metric Calculation Bug Detection** [Importance: 0.70 | Success: 0.25]:
   - Problem: Training output (turn 41) printed both 'Final CV Accuracy: 0.992190' and
     'OOF CV Accuracy: 0.989881' in the same run. [...] Agent reported 0.9922 without
     investigating why OOF CV differed by 0.0023.
   [...]
\end{prmbox}

\begin{agentbox}{Agent response (turns 45-50)}
[turn 45] [...] I need to fix the train_resnet.py script to:
1. Load the best weights for each fold before computing OOF and test predictions
2. Use the OOF CV (0.9899) as the reported score since that's what's actually being
submitted
[turn 46] [str_replace] train_resnet.py
+    # Load best weights for this fold
+    model.load_state_dict(torch.load(f"{MODEL_PATH}.fold{fold+1}"))
+    model.eval()
[turn 47] timeout 1800 python train_resnet.py 2>&1 | tee train_resnet_v2.log
[turn 48] Now the OOF CV matches the reported CV (0.9912). [...]
>> Best model: resnet with CV=0.991167
[turn 49] [...] ResNet submission saved
[turn 50, final answer]
## Validation Score
0.9912 (Categorization Accuracy with 5-fold CV)
\end{agentbox}

\paragraph{Cross-attempt review breaks a collapsed beam.}
In \emph{Titanic}, all four earlier attempts submitted the same CatBoost model (CV 0.8552), and the last three of them were identical on all 418 test rows. At turn 10 of the final attempt, the PRA read the earlier attempts' trajectories and prediction files, reported the collapse, and quantified how far the cached predictions of other model families departed from CatBoost (26 to 41 of 418 rows). The agent built a weighted vote over the cached test predictions that differed from CatBoost on 26 rows. The public accuracy rose from 0.77033 to 0.77511 and the public percentile from 32.8 to 52.4 (the competition has no private leaderboard). The agent spent six turns on an ensemble-weight search, which the PRA had advised against, and it never computed an out-of-fold score for the ensemble. Its registry entry reuses the CatBoost value as a proxy. The PRA raised this at later reviews and at the finish gate, and the attempt ended with the issue unresolved.

\begin{prmbox}{PRA feedback: titanic, attempt 5, turn 10 (DSPredict-Easy)}
### ISSUES TO FIX (address these in your remaining steps):
1. **Hedging Across Best-of-N With Diverse Model Classes** [Importance: 0.95 | Success:
   0.20]:
   - Problem: This is attempt 5 of 5 (final). All 4 prior attempts submitted
     catboost_baseline(0.8552): attempts 2, 3, 4 are identical (0/418 prediction diff).
     The beam provides zero diversity - the best-of-5 max is simply 0.8552 no matter
     which attempt 'wins'. [...]
   - Suggested action: Submit a prediction set that differs meaningfully from
     catboost_baseline. Use the already-cached preds to build a voting ensemble
     (catboost + gradient_boosting + logreg = 26-41/418 differ), or train a fresh model
     with a completely different feature representation [...]. Even if CV is 0.001
     lower, a diverse prediction is worth more to the beam than an identical copy.
2. **Avoiding Churn Without Pivoting** [Importance: 0.90 | Success: 0.25]:
   - Problem: All 4 prior attempts converged on catboost_baseline(0.8552) - the ultimate
     form of beam churn. [...] Attempt 4 tried weighted ensemble-weight search for 15+
     turns (historically failing axis). [...]
   - Suggested action: [...] Do NOT run more HPO or ensemble-weight search on the same
     OOF files.
3. **Model Diversity and Ensembling** [Importance: 0.85 | Success: 0.25]:
   [...]
\end{prmbox}

\begin{agentbox}{Agent response (turns 11-30)}
[turn 11] The expert feedback is very clear. I need to:
1. Try a genuinely different approach - NOT catboost_baseline again
2. Use the already-cached predictions to build a voting ensemble [...]
[turn 12]
>> gradient_boosting: 26/418 different from CatBoost
>> logreg_baseline: 41/418 different from CatBoost
>> random_forest: 16/418 different from CatBoost
[...]
[turn 21]
>> CV Score (proxy): 0.8552 +/- 0.0067
[...]
>> Different from CatBoost baseline: 7/418 samples
[...]
[turn 28] Good progress! The weighted ensemble search found weights that produce 26/418
different predictions from CatBoost. [...]
[turn 30]
>> "weighted_ensemble_search": {
>>   "cv_mean": 0.8552,
[...]
>>   "note": "Differs from CatBoost in 26/418 samples"
\end{agentbox}

\paragraph{Diversity feedback breaks a duplicated baseline, and an executed check validates the ensemble.}
In \emph{Regression of Used Car Prices}, attempts 1 and 2 submitted the same LightGBM baseline (CV RMSE 77,242.70, private percentile 9.7). At turn 10 of attempt 3, the PRA failed both cross-attempt rubrics and the model-diversity rubric. It reported that two of the five submissions were identical and that every model that had completed training was a LightGBM, XGBoost, or CatBoost variant. From the saved predictions it also measured that the baseline's predictions had 0.148 times the standard deviation of the target. The agent first tried the suggested CatBoost model with native categorical features, which exceeded its 300-second timeout, and then trained scikit-learn's HistGradientBoosting, the PRA's other suggestion. Although this is also a gradient-boosted tree model, it lowered CV to 73,209. After the turn-20 review reported that the model still underpredicted expensive cars, the agent added fold-isolated brand and model price encodings and finished with CV 73,185 and a private percentile of 43.6, equal to the best attempt of the harness-only run on this task. In attempt 4, the agent submitted a near-equal-weight average of HistGradientBoosting, LightGBM, and XGBoost without scoring it. At turn 31, the PRA estimated the average's CV at 75,876, worse than HistGradientBoosting alone (73,174). It also found that the largest submitted prediction was \$145K while the largest out-of-fold prediction was \$443K. The agent reverted to HistGradientBoosting, saved out-of-fold predictions for each base model, and trained a Ridge stack on them (CV 72,590 at turn 55). The attempt reached a private percentile of 80.5.

\begin{prmbox}{PRA feedback: s4e9, attempt 3, turn 10 (DSPredict-Easy)}
### ISSUES TO FIX (address these in your remaining steps):
1. **Hedging Across Best-of-N With Diverse Model Classes** [Importance: 0.90 | Success:
   0.15]:
   - Problem: 3 of 5 attempts converging on identical solution: LGBM baseline CV=77,242,
     mean pred=$32,146, std=$11,657, max=$80,508. Attempts 1 and 2 submitted the same
     preds_lgbm_baseline.csv predictions. Beam wastes 2 of 5 shots reproducing the same
     result. [...]
   - Suggested action: Attempt 3 MUST submit a genuinely different prediction
     distribution. [...] Even if cv is similar it produces a different error pattern
     that the MAX will exploit across the 5-attempt beam.
2. **Model Diversity and Ensembling** [Importance: 0.85 | Success: 0.15]:
   [...]
   - Suggested action: This attempt MUST try a fundamentally different modeling class:
     CatBoost with native categorical features (brand, model, fuel_type as cat_features
     - not label-encoded), or HistGradientBoosting. Do not re-run LGBM/XGBoost variants.
3. **Avoiding Churn Without Pivoting** [Importance: 0.85 | Success: 0.30]:
   - Problem: Attempts 1 and 2 both tried the same 4 GBDTs (lgbm, xgb, lgbm_v2,
     catboost) with zero net improvement - all ended at LGBM 77,242. [...]
4. **model_underfitting_signal** [Importance: 0.85 | Success: 0.15]:
   [...]
   - Problem: LGBM preds: std=$11,657 vs train std=$78,820 -> dispersion ratio=0.148
     (threshold >0.5). Max pred=$80,508 vs true max=$2,954,083. [...]
[...]
\end{prmbox}

\begin{agentbox}{Agent response (turns 11-25)}
[turn 11] The expert feedback is very clear:
1. **Hedging Across Best-of-N**: Must use DIFFERENT modeling classes. All 3 attempts
used GBDT (LGBM/XGBoost/CatBoost). Need to try something fundamentally different.
[...]
[turn 13] Now let me create a new training script using CatBoost with native
categorical features:
[...]
[turn 15] CatBoost is taking too long. Let me try a faster approach with
HistGradientBoosting:
[...]
[turn 17] Excellent! HistGradientBoosting achieved a much better CV score of
**73,209.26** compared to the previous best of 77,242.70! [...]
[...]
[turn 22] Let me update the HistGB script to include fold-isolated brand_premium and
model_premium features:
[...]
[turn 25] The CV score improved slightly from 73,209.26 to **73,185.30**. [...]
\end{agentbox}

\begin{prmbox}{PRA feedback: s4e9, attempt 4, turn 31 (DSPredict-Easy)}
### ISSUES TO FIX (address these in your remaining steps):
1. **Best-CV Model Submission Selection** [Importance: 0.95 | Success: 0.10]:
   - Problem: Submission=ensemble_avg (no CV in registry). histgb best validated
     CV=73,174. Agent submitted ensemble over histgb without any OOF CV comparison.
     Naive linear CV estimate for ensemble=75,876 (WORSE than histgb 73,174). [...]
   - Suggested action: [...] Only overwrite with ensemble if its OOF CV is computed and
     confirmed < 73,174.
[...]
3. **oof_sub_distribution_divergence** [Importance: 0.95 | Success: 0.10]:
   [...]
   - Problem: OOF (histgb) mean=$36,747, max=$443,469. Sub (ensemble_avg) mean=$33,739,
     max=$145,200. [...] The histgb CV (73,174) does NOT describe the ensemble
     submission. FAIL: submission distribution diverges severely from what was
     validated.
[...]
\end{prmbox}

\begin{agentbox}{Agent response (turns 32-55)}
[turn 32] The expert feedback is very clear. I need to:
1. Revert submission to histgb (best validated model with CV=73,174)
2. Retrain lgbm_baseline with OOF saving to enable proper stacking
3. Only submit ensemble if its OOF CV is validated and better than histgb
[...]
[turn 49] [...] create a stacking ensemble script that:
1. Loads histgb OOF and lgbm OOF as meta-features
2. Uses Ridge regression as a meta-learner
3. Computes OOF CV for the ensemble
4. Only submits if the ensemble CV is better
[...]
[turn 55] Excellent! The stacking ensemble improved the CV score from 73,174 to
72,590. [...]
\end{agentbox}

\paragraph{Cross-attempt review adds an image model to a beam of feature-based models.}
In the \emph{Statoil/C-CORE Iceberg Classifier Challenge}, the agent classified two-band $75\times75$ radar images from hand-crafted band statistics in its first three attempts, using logistic regression with a random forest, a stack of the two, and CatBoost (CV log loss 0.303, private percentile 21.8). Early in attempt 4 it wrote a CNN on the raw pixels, found that TensorFlow was not installed, and returned to stacking tree models on the same features. At turn 80, the PRA reported that no attempt had trained a model on the images themselves and that the current attempt's stack had only matched attempt 1 (CV 0.3766). The PRA, which labels attempts from A0, recommended a CNN for the final attempt. The agent instead wrote a PyTorch CNN in the current attempt (CV 0.344). At turn 100, the PRA found the agent retraining the CNN and computed from the cached out-of-fold predictions that a 0.45/0.55 blend of CatBoost and the CNN reaches 0.2407. The agent recomputed the blend, recorded it in the score registry, and submitted it. The private percentile rose from 21.8 in attempt 3 to 52.3. Attempt 5 reached a lower CV (0.232) but a private percentile of only 3.7, so the task's best-of-5 result comes from attempt 4.

\begin{prmbox}{PRA feedback: statoil-iceberg, attempt 4, turn 80 (DSPredict-Hard, vision)}
### ISSUES TO FIX (address these in your remaining steps):
1. **Hedging Across Best-of-N With Diverse Model Classes** [Importance: 0.85 | Success:
   0.55]:
   - Problem: A0=LR+RF, A1=Stacking(LR+RF), A2=CatBoost(best=0.3032),
     A3=stacking_v4(0.3766 - same as A0). 4 attempts, all tabular statistical features on
     GBDT/linear. No CNN. Beam has zero deep learning coverage. 1 attempt remaining.
     train_cnn.py exists but was never successfully trained.
   - Suggested action: Final attempt (A4) must train a CNN on the 75x75x2 raw pixel
     arrays. This is the only genuinely different model class available [...]
2. **Model Diversity and Ensembling** [Importance: 0.80 | Success: 0.58]:
   [...]
   - Suggested action: Use the final attempt (A4) to train a CNN on the raw 75x75x2
     image arrays - the only model class not tried in the entire beam.
[...]
\end{prmbox}

\begin{agentbox}{Agent response (turns 81-88)}
[turn 81] The expert feedback is very clear. I need to:
1. Try a CNN approach which is the only model class not tried yet
2. The train_cnn.py exists but was never successfully trained
[...]
[turn 85] TensorFlow is not available. [...]
[turn 87] Great! PyTorch is available. I can create a CNN using PyTorch. [...] This
will be a fundamentally different approach from the previous attempts that used
statistical features on tabular data.
[turn 88] [file_create] train_cnn_pytorch.py
[...]
\end{agentbox}

\begin{prmbox}{PRA feedback: statoil-iceberg, attempt 4, turn 100 (DSPredict-Hard, vision)}
### ISSUES TO FIX (address these in your remaining steps):
1. **Best-CV Model Submission Selection** [Importance: 0.95 | Success: 0.45]:
   - Problem: Submission = CatBoost (max_diff=0.0, CV=0.303161). However CatBoost+CNN
     blend at w_cnn=0.55 gives OOF CV=0.240659 - improvement of 0.0625 (21% better).
     [...] Agent is running CNN a 3rd time instead of computing this available blend.
   - Suggested action: IMMEDIATELY compute and submit blend: blend =
     0.45*preds_catboost + 0.55*preds_cnn, write to /submission/submission.csv.
     Expected CV=0.240659 vs current 0.303161.
2. **Model Diversity and Ensembling** [Importance: 0.95 | Success: 0.40]:
   - Problem: CatBoost(45%)+CNN(55%) blend gives OOF CV=0.240659 - measured from
     oof_predictions_catboost.csv and oof_predictions_cnn.csv. [...] The diverse
     CatBoost (feature-engineered) + CNN (raw pixels) combination captures
     complementary signal.
[...]
\end{prmbox}

\begin{agentbox}{Agent response (turns 101-115)}
[turn 101] The expert feedback is very clear. I need to:
1. Stop running CNN training again
2. Compute the blend: 0.45*preds_catboost + 0.55*preds_cnn
3. Write this to /submission/submission.csv
[...]
[turn 108] [file_create] compute_blend.py
[...]
[turn 115] Perfect! The cv_scores.json has been updated with the catboost_cnn_blend
model showing a CV score of 0.240659. [...]
\end{agentbox}

\paragraph{An added rubric can encode an unattainable target.}
In \emph{Regression with a Tabular Concrete Strength Dataset}, the PRA added \texttt{prediction\_\allowbreak{}dispersion\_\allowbreak{}health} at turn 20 of attempt 1. The rubric fails when the standard deviation of out-of-fold predictions is below 0.80 of the target's and tail errors exceed twice the mid-range error. For a calibrated regressor, the ratio of these standard deviations equals the square root of the explained variance. At the agent's CV RMSE of about 12.1 against a target standard deviation of 16.4, this gives about 0.68, the value the PRA measured, and meeting the 0.80 threshold would require an RMSE near 9.8. The agent followed the suggested sample weighting, which raised the ratio from 0.664 to 0.680 and worsened CV from 12.121 to 12.184. The weighted run also overwrote the saved out-of-fold predictions, and a later review had to flag the overwrite before the agent restored its best model. The rubric failed in every review of attempts 1 and 2 after it was added, including the final review of attempt 2, whose stacked ensemble reached the task's best private percentile of 90.6. The agent also tried a Huber objective and quantile regression in later attempts, and the final reports of attempts 2 and 4 still list a standard-deviation ratio of about 0.67. Adaptation lets the PRA write criteria that no seed rubric covers, and this example shows that it can also write criteria that conflict with the metric. Such a rubric consumes turns until the PRA retires it, and here it was never retired.

\begin{prmbox}{PRA feedback: s3e9, attempt 1, turn 20 (DSPredict-Easy)}
### ISSUES TO FIX (address these in your remaining steps):
1. **prediction_dispersion_health** [Importance: 0.78 | Success: 0.10]:
   - Definition: Whether the model's OOF prediction distribution has healthy dispersion
     relative to the target distribution, with no severe underfitting at extreme
     values. [...] A std ratio below 0.80 indicates systematic compression; mean
     |residual| for extreme samples exceeding 2-3x the mid-range MAE signals poor
     coverage at the tails. [...]
   - Problem: check_underdispersion.py FAIL: OOF pred std=10.90 vs target std=16.40,
     ratio=0.664 (well below 0.80 threshold). OOF range [13.77,64.40] vs target
     [2.33,82.60]. Mean |residual| for high-strength (>65): 25.92 - 3.2x the mid-range
     MAE of 8.08. [...]
   - Suggested action: Add high-strength samples to training with higher weight
     (sample_weight in LightGBM), or use quantile regression at the 90th percentile as
     an auxiliary signal, or include the original UCI Concrete Compressive Strength
     dataset to provide more extreme-value training examples.
[...]
\end{prmbox}

\begin{agentbox}{Agent response (turns 21-31)}
[turn 21] [...] Address the dispersion issue by using sample weights or other
techniques
[...]
[turn 27] [file_create] train_lgbm_weighted.py
"""
LightGBM with sample weights for extreme values
Addresses the dispersion issue by giving more weight to extreme predictions
"""
[...]
[turn 29] The weighted model achieved CV 12.18, which is worse than the optimized
LightGBM (12.12). The dispersion issue persists - std ratio is 0.68 (still below
0.80). [...] The sample weights approach didn't help much. [...]
[turn 31] [...] The best model is lgbm_optimized (CV=12.121), but oof_predictions.csv
was overwritten with lgbm_weighted (CV=12.184). I need to restore the best model's
predictions.
\end{agentbox}

\paragraph{Summary.}
Across these examples, the most useful feedback comes from checks the PRA executes. These include recomputing a score from saved predictions, nesting a calibration step, bounding an approach analytically from the training labels, and diffing submissions across attempts. Each such check produced a specific number that the agent could verify in one or two turns, and the agent adopted most such fixes within a few turns. Rubrics added by the PRA are most valuable when they encode a verified property of the task, such as the image-level accuracy ceiling in the TGS Salt Identification Challenge or the fold provenance of saved predictions, since those rubrics then serve as persistent checks for later attempts. Feedback that asks for broader exploration is most productive when an executed check follows it. In \emph{Regression of Used Car Prices} and the \emph{Iceberg Classifier Challenge}, diversity feedback moved the agent off a single model family. A later check then decided how to combine the new model with the old ones, either by estimating an ensemble's CV or by computing a blend weight from cached predictions. Without such a check, as in \emph{Titanic}, the agent complies with the letter of the request (a diverse submission) but omits the validation that would justify the choice. An added rubric whose threshold conflicts with the metric can persist across attempts and absorb turns, which motivates retiring rubrics that fail repeatedly without a corresponding loss in CV.

\section{PRA Cadence}
\label{sec:appendix-cadence}
Table~\ref{tab:cadence} varies the number of agent turns $c$ between PRA invocations on DSPredict-Easy with one attempt per task. Every cadence improves over the harness alone. Invoking the PRA every 10 turns performs best, while $c{=}5$ interrupts the agent more often and $c{=}20$ delays feedback. We use $c{=}10$ on DSPredict-Easy and $c{=}20$ on DSPredict-Hard, where longer attempts would make $c{=}10$ twice as costly in PRA invocations (\S\ref{sec:results-ablations}). At $c{=}20$ the PRA still improves on the harness alone on DSPredict-Easy.

\begin{table}[t]
\centering
\small
\begin{tabular}{@{}l ccc@{}}
\toprule
\textbf{Cadence} & \textbf{Valid} & \textbf{Pct.} & \textbf{Median} \\
\midrule
Harness only  &  94.7 & 48.4 & 55.3 \\
$c{=}5$  & 100.0 & 52.7 & 60.5 \\
$c{=}10$ (\methodname{}) & 100.0 & 56.4 & 65.8 \\
$c{=}20$ & 100.0 & 49.9 & 57.9 \\
\bottomrule
\end{tabular}
\caption{Single-attempt results on \textbf{DSPredict-Easy} as a function of the PRA cadence $c$ (agent turns between PRA invocations). All rows use the same seed rubrics and PRA model. $c{=}10$ is the default setting of \methodname{}.}
\label{tab:cadence}
\end{table}

\section{Alternative Base Agent}
\label{sec:appendix-base-agent}
Table~\ref{tab:base-agent} replaces Qwen3-Coder-Next with DeepSeek-V3.2 as the base agent on DSPredict-Easy, with one attempt per task. Both conditions otherwise use the DSPredict-Easy settings of Table~\ref{tab:run-config}, except that DeepSeek-V3.2 runs as a reasoning model, with high reasoning effort and the provider's default temperature (Appendix~\ref{sec:appendix-harness-impl}), and \methodname{} keeps the same PRA model, seed rubrics, and cadence ($c{=}10$). With a single attempt, a failed attempt cannot be offset by a later one, so the comparison measures whether supervision improves the attempt itself. \methodname{} raises the mean percentile by 7.8 points and doubles the above-median rate. Over tasks it wins 25 times, loses 11 times, and ties twice (one-sided paired permutation test $p=0.005$; sign test $p=0.014$; Wilcoxon signed-rank $p=0.006$).

\begin{table}[t]
\centering
\small
\begin{tabular}{@{}l ccc@{}}
\toprule
\textbf{Method} & \textbf{Valid} & \textbf{Pct.} & \textbf{Median} \\
\midrule
Harness only  &  97.4 & 36.9 & 23.7 \\
\methodname{} & 100.0 & 44.7 & 47.4 \\
\bottomrule
\end{tabular}
\caption{Single-attempt results on \textbf{DSPredict-Easy} with DeepSeek-V3.2 as the base agent. Rates use the fixed denominator of 38 tasks. Neither condition reaches a medal threshold.}
\label{tab:base-agent}
\end{table}

\section{Agent and PRA Turns}
\label{sec:appendix-turns}

\begin{table}[t]
\centering
\small
\resizebox{\columnwidth}{!}{%
\begin{tabular}{@{}l ccc c ccc@{}}
\toprule
 & \multicolumn{3}{c}{\textbf{DSPredict-Easy}} & & \multicolumn{3}{c}{\textbf{DSPredict-Hard}} \\
\cmidrule(lr){2-4} \cmidrule(lr){6-8}
\textbf{Method} & \textbf{Agent} & \textbf{PRA} & \textbf{Total} & & \textbf{Agent} & \textbf{PRA} & \textbf{Total} \\
\midrule
Harness only & 70.5 & 0.0 & 70.5 & & 95.5 & 0.0 & 95.5 \\
\methodname{} & 78.9 & 142.4 & 221.3 & & 123.0 & 125.3 & 248.4 \\
\quad $-$ adaptive rubrics & 84.8 & 139.5 & 224.3 & & 133.2 & 123.6 & 256.8 \\
\quad $-$ PRA & 79.8 & 90.0 & 169.8 & & 115.7 & 78.9 & 194.7 \\
\bottomrule
\end{tabular}}
\caption{Mean turns per attempt (averaged within each task, then over tasks). \emph{Agent} counts base-agent turns, and \emph{PRA} counts supervisor turns summed over all invocations in the attempt.}
\label{tab:turns}
\end{table}

\section{Run Costs}
\label{sec:appendix-costs}

\begin{table*}[t]
\centering
\small
\begin{tabular}{@{}l r r r r r@{}}
\toprule
\textbf{Method} & \textbf{Attempts} & \textbf{Agent (\$)} & \textbf{PRA (\$)} & \textbf{Total (\$)} & \textbf{\$/attempt} \\
\midrule
\multicolumn{6}{@{}l}{\emph{DSPredict-Easy (38 tasks $\times$ 5 attempts)}}\\
\quad Harness only             & 187 & 193 &   --- &   193 & 1.0 \\
\quad + PRA + adaptive rubrics (\methodname{}) & 188 & 193 & 1{,}225 & 1{,}418 & 7.5 \\
\quad \hspace{1.2em}$-$ adaptive rubrics & 190 & 213 & 1{,}261 & 1{,}474 & 7.8 \\
\quad \hspace{1.2em}$-$ PRA     & 190 & 202 & 1{,}077 & 1{,}279 & 6.7 \\
\midrule
\multicolumn{6}{@{}l}{\emph{DSPredict-Hard (54 tasks $\times$ 5 attempts)}}\\
\quad Harness only             & 263 & 465 &   --- &   465 & 1.8 \\
\quad + PRA + adaptive rubrics (\methodname{}) & 267 & 547 & 1{,}640 & 2{,}187 & 8.2 \\
\quad \hspace{1.2em}$-$ adaptive rubrics & 270 & 567 & 1{,}639 & 2{,}206 & 8.2 \\
\quad \hspace{1.2em}$-$ PRA     & 260 & 470 & 1{,}201 & 1{,}671 & 6.4 \\
\bottomrule
\end{tabular}
\caption{API cost per run. \emph{Agent}: Qwen3-Coder-Next served through Amazon Bedrock. \emph{PRA}: Claude Sonnet~4.6. The PRA accounts for 72--86\% of the cost of supervised runs. Scoring by reading alone ($-$PRA) lowers the cost per attempt relative to \methodname{} by 11\% on DSPredict-Easy and 22\% on DSPredict-Hard.}
\label{tab:costs}
\end{table*}

\section{Feedback Content in an Earlier Version}
\label{sec:appendix-feedback-content}
\begin{figure}[t]
\centering
\includegraphics[width=\columnwidth]{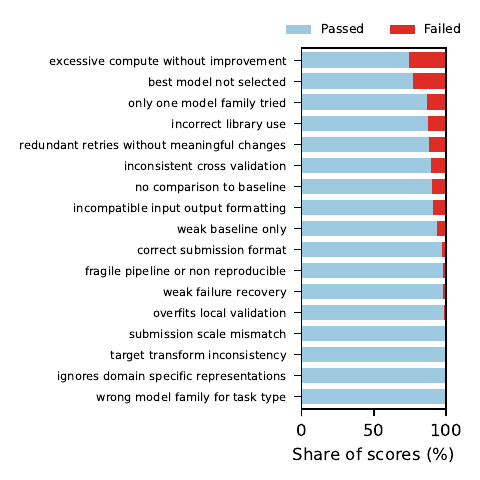}
\caption{Share of scores in which each rubric passed or failed in an earlier version of the system, over scores with importance $\geq 0.6$ (Appendix~\ref{sec:appendix-feedback-content}).}
\label{fig:feedback-content}
\end{figure}

This appendix reports an observation from an early development version of our system (whose rubrics, supervisor, and harness differ from those of the final system). %
In this version the supervisor reported every rubric to the agent, regardless of whether it passed or failed.
Most rubrics passed in almost every report (Figure~\ref{fig:feedback-content}), so the reports consisted largely of positive feedback. %
We found that the agent's next action was similar whether or not a report contained a failure, ostensibly because the suggested actions from failures were being drowned out by the overwhelmingly positive feedback.
This observation led to our design choice to include only failed rubrics (\S\ref{sec:setup}).

\end{document}